\documentclass[10pt,twocolumn,letterpaper]{article}
\usepackage{booktabs,makecell,multirow}
\usepackage[normalem]{ulem} % for \uline
\usepackage[pagenumbers]{cvpr} % To force page numbers, e.g. for an arXiv version

\usepackage[utf8]{inputenc} % allow utf-8 input
\usepackage[T1]{fontenc}    % use 8-bit T1 fonts
\usepackage{url}            % simple URL typesetting
\usepackage{booktabs}       % professional-quality tables
\usepackage{amsfonts}       % blackboard math symbols
\usepackage{ulem}           % for \uline (underlining) command
\usepackage{nicefrac}       % compact symbols for 1/2, etc.
\usepackage{microtype}      % microtypography
\usepackage{xcolor}         % colors
\usepackage{amsmath}
\usepackage{makecell}
\usepackage{algorithm}
\usepackage{algpseudocode}
\usepackage{enumitem}
\usepackage{wrapfig} 
\usepackage{multirow}
\usepackage{graphicx}
\usepackage{siunitx}
\usepackage{tabularx}
\usepackage{listings}
\usepackage{fvextra} 
\usepackage{tcolorbox}
\usepackage{caption}
\usepackage{morefloats}
\tcbuselibrary{breakable}

\usepackage{array}
\usepackage{lscape}

\definecolor{deepgreen}{HTML}{006400} % Dark green
\definecolor{mypurple}{HTML}{800080}  % Purple

\setlist[enumerate]{leftmargin=1.5em}
\setlist[itemize]{leftmargin=1.5em}
\tcbset{
  prettytxt/.style={
    breakable,
    parbox=false,
    colback=gray!6,
    colframe=gray!70,
    boxrule=1pt,
    left=.7em,right=.7em,top=.7em,bottom=.7em,
    arc=.7em,           % 圆角
    fonttitle=\bfseries
  }
}

\newcommand{\method}[0]{\textsc{OmniAlpha}}

\definecolor{cvprblue}{rgb}{0.21,0.49,0.74}
\usepackage[pagebackref,breaklinks,colorlinks,allcolors=cvprblue]{hyperref}

\title{\method{}: Aligning Transparency-Aware Generation via \\  Multi-Task Unified Reinforcement Learning}

\author{%
  \textbf{Hao Yu}$^{1*}$ \quad
  \textbf{Jinglin Wang}$^{12*}$  \quad
  \textbf{Jiabo Zhan}$^{1*}$ \quad
  \textbf{Rui Chen}$^{3}$  \quad
  \textbf{Zile Wang}$^{1}$ \\
  \textbf{Huaisong Zhang}$^{1}$ \quad
  \textbf{Hongyu Li}$^{3}$ \quad
  \textbf{Xinrui Chen}$^{1}$ \quad
  \textbf{Yongxian Wei}$^{1}$ \quad
  \textbf{Chun Yuan}$^{1\dagger}$  
  \\[.5em]
  $^{1}$Tsinghua University \quad
  $^{2}$Beijing University of Posts and Telecommunications \quad
  $^{3}$Beihang University 
  \\[.2em]
  \texttt{yuh24@mails.tsinghua.edu.cn, yuanc@sz.tsinghua.edu.cn} \\[.5em]
  \texttt{https://github.com/Longin-Yu/OmniAlpha}
  \vspace{-2em}
}

\begin{document}

\twocolumn[{%
\renewcommand\twocolumn[1][]{#1}%
\maketitle
\begin{center}
    \centering
    \captionsetup{type=figure}
    % \vspace{-1em}
    \includegraphics[width=1.0\linewidth]{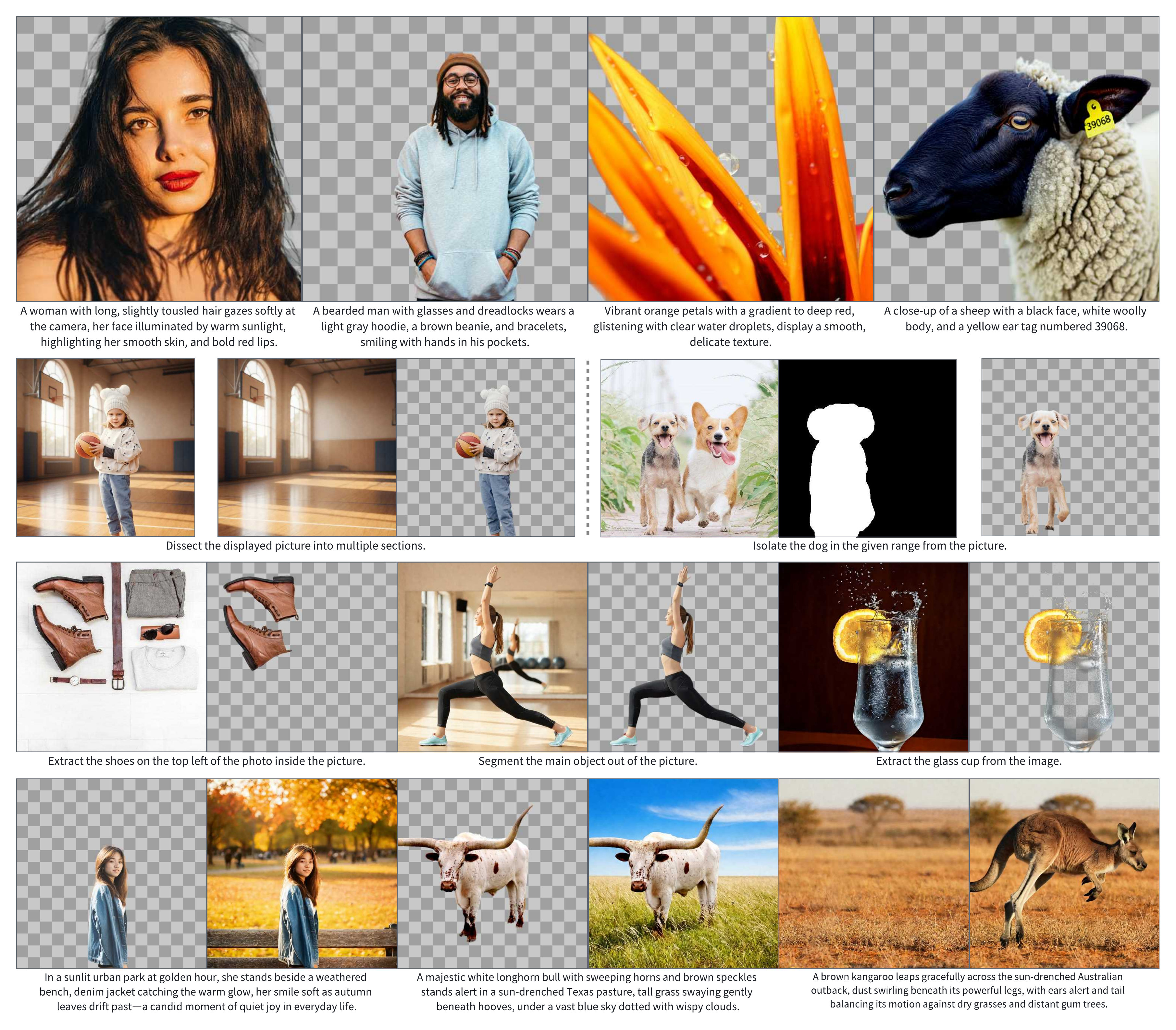}
    \captionof{figure}{
Demonstrating \method{}'s versatility across a range of RGBA tasks. Our unified model handles: text-to-image generation (Row 1); layer decomposition and object removal (Row 2); referring matting and automatic matting (Row 3).
}
    \label{fig:example}
\end{center}
}]

\maketitle
\newcommand\blfootnote[1]{%
  \begingroup
  \renewcommand\thefootnote{}\footnote{#1}%
  \addtocounter{footnote}{-1}%
  \endgroup
}
\blfootnote{$^*$Equal contribution. $^\dagger$Corresponding author. This work was done when Jinglin was a research intern at Tsinghua University.}

\begin{abstract}
Transparency-aware generation requires modeling not only RGB appearance but also alpha-based opacity and cross-layer composition, which are essential for tasks such as image matting, object removal, layer decomposition, and multi-layer content creation. 
However, existing RGBA-related methods remain largely fragmented, with separate pipelines designed for individual tasks.
While a unified model is desirable, supervised fine-tuning alone is insufficient, as localized regression objectives cannot directly optimize the compositional fidelity, alpha-boundary precision, and structural consistency required for high-quality RGBA generation. 
To address this, we propose \method{}, a unified multi-task reinforcement learning framework for transparency-aware generation and manipulation. 
\method{} combines an end-to-end alpha-aware VAE and a sequence-to-sequence Diffusion Transformer, with a bi-directional layer axis in positional encoding to jointly model multiple RGBA inputs and outputs within a single forward pass.
Built on a multi-task SFT cold start, it further performs GRPO-style post-training with layer-aware rewards defined on decoded RGBA outputs, enabling direct optimization of cross-layer coherence and fine transparency details.
Experiments across five categories of transparency-aware tasks show that \method{} consistently outperforms its unified SFT baseline and achieves strong performance against specialized expert models, including a 9.07\% relative reduction in RGB L1 on layer decomposition and 74\%/68\% improvements over conventional matting tools on SAD/Grad for automatic matting.
% For reproducibility, our implementation is available at \url{https://anonymous.4open.science/r/OmniAlpha}.
\end{abstract}

\section{Introduction}
\label{sec:intro}

Diffusion-based generative models~\citep{ho2020denoisingdiffusionprobabilisticmodels, rombach2022highresolutionimagesynthesislatent, peebles2023scalablediffusionmodelstransformers} have become the dominant paradigm for high-fidelity image synthesis. Powered by scalable latent autoencoding and large-scale text-to-image pre-training, modern systems such as Stable Diffusion~\citep{rombach2022highresolutionimagesynthesislatent}, FLUX~\citep{labs2025flux1kontextflowmatching}, and Qwen-Image~\citep{wu2025qwenimagetechnicalreport} have established strong foundations for RGB image generation.

However, many real-world visual creation workflows require representations beyond opaque RGB pixels. In applications such as visual effects, graphic design, image editing, and multi-layer content creation, transparency is a first-class necessity rather than an optional attribute. The RGBA format explicitly models per-pixel opacity through an additional alpha channel, making it essential for preserving fine structures such as hair and fur, representing semi-transparent materials such as glass, smoke, and water, and supporting flexible layer composition.
 
To address these needs, prior work has largely developed specialized models for individual RGBA-related tasks.
Image matting techniques~\citep{yao2023vitmatteboostingimagematting, hu2024diffusionnaturalimagematting} 
focus on accurate alpha estimation;
layer decomposition pipelines~\citep{yang2024generativeimagelayerdecomposition} 
separate an image into its constituent foreground and background layers; 
object removal systems~\citep{winter2024objectdropbootstrappingcounterfactualsphotorealistic} 
remove specified objects and reconstruct occluded content;
and text-to-image workflows like LayerDiffuse~\citep{zhang2024transparentimagelayerdiffusion} 
explore text-conditioned RGBA synthesis.
Despite their effectiveness, these approaches are inherently fragmented.
Each task is addressed by a dedicated pipeline, which is not only inefficient but also fails to capture the shared layer-aware structure underlying diverse RGBA workflows.

This fragmentation motivates a unified foundation model for transparency-aware generation and manipulation, analogous to the recent trend towards unified modeling in the RGB domain
~\citep{xiao2024omnigenunifiedimagegeneration, fang2024pumaempoweringunifiedmllm, li2025visualclozeuniversalimagegeneration}.
To achieve this unification, scaling up Supervised Fine-Tuning (SFT) across diverse tasks serves as the intuitive next step. 
Yet, pushing this strategy to its limits exposes a fundamental bottleneck: while effective for instilling basic multi-task capabilities, its localized regression objectives do not directly optimize the properties that matter most in transparency-aware generation, such as cross-layer consistency, alpha-boundary precision, and compositional faithfulness.
Consequently, an SFT-only model often plateaus, struggling to capture the complex cross-layer physics and precise boundary fidelity essential for seamless transparency-aware generation.

This limitation suggests moving from purely SFT to outcome-oriented post-training. Reinforcement learning (RL) has recently shown strong promise for aligning RGB diffusion models, with methods such as DanceGRPO~\citep{xue2025dancegrpounleashinggrpovisual} and Flow-GRPO~\citep{liu2025flowgrpotrainingflowmatching} demonstrating the value of optimizing generation quality at the trajectory level.
However, this emerging RL paradigm remains strictly confined to the RGB space. Fundamentally geared toward opaque pixels, their alignment mechanisms lack the structural awareness needed to process the alpha channel or evaluate multi-layer compositions. This renders existing RL approaches incapable of addressing the complex, layer-based workflows inherent to the RGBA domain, leaving transparency-aware generation largely unexplored.

To bridge this gap, we propose \method{}, a unified multi-task reinforcement learning framework designed to align transparency-aware generation.
Our framework combines an end-to-end alpha-aware VAE with a sequence-to-sequence Diffusion Transformer (DiT), and uses a bi-directional layer coordinate in positional embeddings to jointly process multiple RGBA inputs and outputs within a single forward pass. 
We first use multi-task SFT as a cold-start stage to establish broad task competence, and then perform
Group Relative Policy Optimization (GRPO)-style post-training, driven by rewards tailored specifically to the decoded RGBA layer structures. By harnessing RL to orchestrate the unified generative process, \method{} explicitly optimizes 
structural fidelity, compositional consistency, and fine alpha details that are difficult to capture through SFT alone.

Our experiments show that the RL alignment in \method{} unlocks significant gains over standard supervised training, yielding a 9.07\% relative improvement in RGB L1 for layer decomposition over its SFT counterpart. Furthermore, by unifying fragmented task-specific solutions into a single aligned policy, \method{} eclipses conventional auto-matting baselines by reducing SAD and Grad by 74\% and 68\%, respectively. These results highlight the strong generalizability of our approach, delivering robust, state-of-the-art performance across diverse datasets.

\vspace{.5em}
\noindent
In summary, our contributions are as follows:
\begin{itemize}
\item We propose \method{}, the first reinforcement learning framework for unified multi-task RGBA generation. By shifting the paradigm from localized supervised fine-tuning to holistic RL alignment, we enable a single policy to master diverse transparency-aware workflows.
\item We introduce structurally-grounded rewards for multi-layer compositional alignment, establishing a post-training paradigm that successfully transcends the performance ceiling of SFT.
\item Experiments show that \method{} consistently achieves state-of-the-art results across multiple benchmarks, outperforming both its multi-task SFT baseline and specialized expert models.
\end{itemize}

\section{Related Work}
\label{sec:related}

% \noindent Test Paragraph  Test Paragraph  Test Paragraph  Test Paragraph  Test Paragraph  Test Paragraph  Test Paragraph  Test Paragraph  Test Paragraph  Test Paragraph  Test Paragraph  Test Paragraph  Test Paragraph  Test Paragraph  Test Paragraph  Test Paragraph  Test Paragraph  Test Paragraph  Test Paragraph  Test Paragraph  Test Paragraph  Test Paragraph  Test Paragraph  Test Paragraph  Test Paragraph  Test Paragraph  Test Paragraph  Test Paragraph  Test Paragraph  Test Paragraph  Test Paragraph  Test Paragraph  Test Paragraph  Test Paragraph  Test Paragraph  Test Paragraph  Test Paragraph  Test Paragraph  Test Paragraph  Test Paragraph  Test Paragraph  Test Paragraph  

\noindent\textbf{Transparency-aware Image Generation.}
Large diffusion models, such as Stable Diffusion~\citep{rombach2022highresolutionimagesynthesislatent}, have achieved high-quality image synthesis for T2I tasks, but their outputs are limited to single-layer RGB images. Recent advances have extended these models to generate RGBA content directly, often using diffusion transformer ~\citep{peebles2023scalablediffusionmodelstransformers} architectures. Numerous studies have demonstrated the feasibility of multi-layer RGBA generation (PSDiffusion~\citep{huang2025psdiffusionharmonizedmultilayerimage}, LayerFusion~\citep{dalva2024layerfusionharmonizedmultilayertexttoimage}, LayerDiffuse~\citep{zhang2024transparentimagelayerdiffusion}, ART~\citep{pu2025artanonymousregiontransformer}, DreamLayer~\citep{huang2025dreamlayersimultaneousmultilayergeneration}) and end-to-end generation  (AlphaVAE~\citep{wang2025alphavae0}, Alfie~\citep{quattrini2024alfiedemocratisingrgbaimage}) using diffusion models. However, these works focus mainly on T2I generation, with limited exploration of more complex visual generation tasks.

\noindent\textbf{Task-Specific and Multi-Task Models.} 
Some methods have extended the RGB/RGBA generation capability of diffusion models, with representative approaches emerging for specific tasks. For instance, dedicated models have emerged for object removal (ObjectDrop~\citep{winter2024objectdropbootstrappingcounterfactualsphotorealistic}, ObjectClear~\citep{zhao2026preciseobjecteffectremoval}, PowerPaint~\citep{zhuang2024taskworthwordlearning}, DesignEdit~\citep{jia2024designeditmultilayeredlatentdecomposition}), image matting (MAM~\citep{li2023matting}, Matte Anything~\citep{yao2024matteanythinginteractivenatural}, TeachDiffusionMatting~\citep{xiang2025teaching}, ViTMatte~\citep{yao2023vitmatteboostingimagematting}, DiffMatte~\citep{hu2024diffusionnaturalimagematting}, DRIP~\citep{Li2024DRIPUD}), and generative layer decomposition (LayerDecomp~\citep{yang2024generativeimagelayerdecomposition}). While powerful, these approaches are inherently siloed, lacking the flexibility to generalize across multiple generation scenarios. The pursuit of such generalization has led to several unified multi-task models (OmniGen~\citep{xiao2024omnigenunifiedimagegeneration}, OmniGen2~\citep{wu2025omnigen2explorationadvancedmultimodal}, VisualCloze~\citep{li2025visualclozeuniversalimagegeneration}, DreamOmni~\citep{xia2025dreamomniunifiedimagegeneration}, PUMA~\citep{fang2024pumaempoweringunifiedmllm}), but these frameworks are confined to RGB generation. Consequently, a framework that jointly addresses multi-task RGBA generation remains a relatively underexplored area.

\noindent\textbf{Reinforcement Learning for Diffusion Post-training}
Generative diffusion models have increasingly adopted post-training to better align their outputs with target objectives.
Early efforts (DPOK~\citep{fan2023dpokreinforcementlearningfinetuning}, DDPO~\citep{black2024trainingdiffusionmodelsreinforcement}) formulated denoising as a sequential decision problem and optimized sampling with PPO-style methods~\citep{schulman2017proximalpolicyoptimizationalgorithms}.
This line of work was later developed under the GRPO paradigm~\citep{shao2024deepseekmathpushinglimitsmathematical} (DanceGRPO~\citep{xue2025dancegrpounleashinggrpovisual}, Flow-GRPO~\citep{liu2025flowgrpotrainingflowmatching}, MixGRPO~\citep{li2026mixgrpounlockingflowbasedgrpo}), with subsequent efforts further improving its efficiency and applicability.
However, confined to RGB-centric rewards, these paradigms lack the structural priors necessary for alpha-channel optimization.

\begin{figure*}[t]
    \centering
    \includegraphics[width=.9\linewidth]{figs/arch_update.pdf} 
    \vspace{-1em}
    \caption{Overview of the \method{} Diffusion Transformer architecture and the Cold-Start Stage. Conditioned on a task instruction and $n$ RGBA images, the model simultaneously denoises $m$ target images. To process multiple layers concurrently, we extend positional embeddings with a bi-directional $z$-axis.}
    \label{fig:arch}   
\end{figure*}

\section{Methodology}
\label{sec:method}

\newcommand{\imgspace}{\mathbb{R}^{H\times W\times 4}}
\newcommand{\rgbspace}{\mathbb{R}^{H\times W\times 3}}
\newcommand{\rgbaspace}{\mathbb{R}^{H\times W\times 4}}
\newcommand{\latentspace}{\mathcal{Z}}
\newcommand{\vaee}{\mathcal{E}}
\newcommand{\vaed}{\mathcal{D}}
\newcommand{\vaeeref}{{\vaee_{\text{ref}}}}
\newcommand{\vaedref}{{\vaed_{\text{ref}}}}
\newcommand{\posint}{\mathbb{N}^{+}}

In this section, we introduce our proposed \method{}. Our architecture is built upon the latent diffusion paradigm and comprises two core components. The first is an end-to-end, transparency-aware VAE, which is efficiently initialized from a pre-trained RGB autoencoder using an opaque initialization strategy. The second is the denoising backbone, a diffusion transformer. To support our sequence-to-sequence task formulation, we extend the standard 3D rotary positional encoding with a bi-directional layer coordinate (z-axis). This structural adaptation enables the DiT to differentiate and process multiple input and output images concurrently within a single forward pass.

\subsection{Formulation and Unified Architecture}

\noindent\textbf{Task Formulation.}
We formulate the multi-layer visual manipulation task as a conditional sequence-to-sequence generation problem. Given a text instruction $T$ and $n$ input RGBA images $x_1, \cdots, x_n \in \imgspace$, the model $\mathcal{M}$ predicts $m$ target RGBA images $\hat{y}_1, \cdots, \hat{y}_m \in \imgspace$:
\begin{equation}
      \hat{y}_1,\cdots,\hat{y}_m = \mathcal{M}(x_1,\cdots,x_n;m,T)
\end{equation}
Operating in a lower-dimensional latent space $\latentspace$, this generative process is parameterized by a velocity predictor $v_\theta(z_t, t, T, \{x_k\}_{k=1}^{n})$ within the flow-matching framework, which estimates the velocity field of the latent variable $z_t$ at timestep $t$. 
In practice, we instantiate this unified paradigm across five categories of transparency-aware tasks: text-to-image generation, object removal, automatic matting, referring matting, and layer decomposition.

\noindent\textbf{End-to-End Transparency-aware VAE.}
To map the pixel space $\imgspace$ to the latent space $\latentspace$, we build the transparency-aware autoencoder $\langle\vaee,\vaed\rangle$ following AlphaVAE. It is initialized from a pre-trained RGB VAE via an opaque initialization strategy: the first layer of $\vaee$ zeros out the alpha channel weights to focus initially on RGB features, while the final layer of $\vaed$ is deterministically initialized to predict a fully opaque alpha channel ($\alpha=1.0$). Detailed formulations are provided in Appendix~\ref{app:arch:vae}.

\noindent\textbf{Sequence-to-Sequence Diffusion Transformer.}
The denoising backbone $v_\theta$ is a Multimodal Diffusion Transformer. Following a dual-stream paradigm~\citep{wu2025qwenimagetechnicalreport}, the $n$ input images are simultaneously processed through two pathways: fed into a Vision-Language Model alongside instruction $T$ for semantic context, and compressed by $\vaee$ into spatial conditions $\vaee(x_k)$. To concurrently manage these $n$ inputs and $m$ noisy targets, we extend 2D rotary positional embeddings with a bi-directional layer coordinate (z-axis). As illustrated in Figure~\ref{fig:arch}, this z-axis spatially differentiates the sequence by assigning distinct layer indices to the input latents, target latents, and language embeddings (details in Appendix~\ref{app:arch:rope}).
% The denoising backbone $v_\theta$ is a Multimodal Diffusion Transformer that operationalizes our sequence-to-sequence formulation. 
% Adopting a dual-stream conditioning paradigm~\citep{wu2025qwenimagetechnicalreport}, we use a Vision-Language Model for the semantic instruction $T$, and the VAE encoder $\vaee$ to independently compress the $n$ input images into spatial conditions $z_{x,k} = \vaee(x_k)$. 
% To concurrently process the $n$ inputs and $m$ noisy targets within a single forward pass, we extend 2D rotary positional embeddings with a bi-directional layer coordinate (z-axis), as shown in Figure~\ref{fig:arch}. 
% This structural adaptation allows the transformer to spatially differentiate multimodal inputs and multiple outputs by assigning distinct layer indices to the input latents, target latents, and language embeddings. Detailed formulations are provided in Appendix~\ref{app:arch:rope}.

\subsection{Supervised Cold-Start}

\noindent\textbf{Transparency-Aware Alignment of VAE.}
Before training the diffusion backbone, we align our autoencoder to the 4-channel RGBA space. 
% Utilizing the complete curated multi-layer dataset, we fine-tune the opaque-initialized VAE $\langle\vaee, \vaed\rangle$. 
Following AlphaVAE, the training objective incorporates reconstruction, perceptual, two KL divergence terms, and adversarial losses:
\begin{equation}
\begin{split}
    \mathcal{L}(\vaee,\vaed) 
    & = \lambda_{rec} \mathcal{L}_{rec}(\vaee,\vaed)
    + \lambda_{perc} \mathcal{L}_{perc}(\vaee,\vaed) \\
    & + \lambda_{kl} \mathcal{L}_{kl}(\vaee;\vaeeref)
    + \lambda_{ref} \mathcal{L}_{kl}(\vaee;\mathcal{N}(\mathbf{0};\mathbf{I})) \\
    & + \lambda_{GAN} \mathcal{L}_{GAN}(\{\vaee,\vaed\};\mathcal{P}_d)
\end{split}
\end{equation}

\noindent\textbf{Multi-Task Joint Fine-tuning.}
With the VAE aligned and the diverse transparency-aware tasks unified under a single sequence-to-sequence paradigm, we train the DiT using the Flow Matching objective~\citep{DBLP:conf/iclr/LipmanCBNL23, esser2024scaling}. Let $\mathcal{T}$ denote the set of all tasks and $\mathcal{D}_\tau$ the data distribution for task $\tau \in \mathcal{T}$. For a sampled instance $(c, Y) \sim \mathcal{D}_\tau$, where $c=(T,\{x_k\}_{k=1}^{n})$ represents the multimodal condition and $Y = (y_1, \dots, y_m)$ is the sequence of $m$ target RGBA images, we encode the targets into a latent sequence $Z_0 = (z_{0,1}, \dots, z_{0,m})$ via $z_{0,k} = \vaee(y_k)$. Given a standard Gaussian noise sequence $Z_1 = (z_{1,1}, \dots, z_{1,m})$ where $z_{1,k} \sim \mathcal{N}(\mathbf{0}, \mathbf{I})$, we define the intermediate state at timestep $t \sim \mathcal{U}[0, 1]$ as $Z_t = (1-t) Z_0 + t Z_1$. The model $v_\theta$ is trained to predict the marginal velocity field. To normalize the loss across variable output lengths $m$, we average the $\ell_2$ errors:
\begin{equation}
\resizebox{.43\textwidth}{!}{
$
    \mathcal{L}_{\mathrm{SFT}} = \mathbb{E}_{\tau \sim \mathcal{T}, (c, Y) \sim \mathcal{D}_\tau, t, Z_1} \left[ \frac{1}{m} \sum_{k=1}^{m} \left\| v_{\theta}^{(k)}(Z_t, t, c) - (z_{1,k} - z_{0,k}) \right\|^2_2 \right],
$
}
\end{equation}
where $v_{\theta}^{(k)}$ denotes the velocity prediction corresponding to the $k$-th target latent. 

\subsection{Layer-aware Alignment via RL}
While SFT provides a foundational generative policy, its density-based regression objective often fails to capture spatially sparse yet visually critical structures, such as fine-grained hair boundaries. To address this, we introduce a post-training stage that shifts from localized velocity regression to direct output-space supervision.

% \noindent\textbf{Formulating RGBA Denoising as an MDP.}
% We formulate the reverse denoising process for unified RGBA generation as a Markov Decision Process (MDP), where each episode corresponds to a complete denoising rollout from an initial noisy latent sequence to the final decoded layered RGBA outputs.

% At timestep $t$, the state is given by $s_t \triangleq (c, t, Z_t)$, where $c$ denotes the conditioning context of a training instance and $Z_t$ is the noisy latent sequence of the target RGBA layers at timestep $t$. The action is defined as the next reverse-diffusion state, $a_t \triangleq Z_{t-1}$, corresponding to a denoising step from timestep $t$ to $t-1$. The policy is therefore given by
% \begin{equation}
% \pi_{\theta}(a_t \mid s_t) \triangleq p_{\theta}(Z_{t-1} \mid Z_t, c),
% \end{equation}
% which is parameterized by the unified RGBA denoising network. We adopt a terminal reward formulation:
% \begin{equation}
% R(s_t, a_t)=
% \begin{cases}
% r(\hat{Y}; c), & t=0,\\
% 0, & \text{otherwise},
% \end{cases}
% \end{equation}
% where $\hat{Y}=\mathcal{D}(Z_0)$ denotes the final decoded layered RGBA outputs.

\medskip
\noindent\textbf{Formulating RGBA Generation as an MDP.}
Our flow-matching framework generates samples by integrating the velocity field backwards from $t=1$ to $t=0$. We formulate this generative process as a Markov Decision Process (MDP), where each episode corresponds to a complete ordinary differential equation (ODE) rollout from an initial noise sequence $Z_1 \sim \mathcal{N}(\mathbf{0}, \mathbf{I})$ to the final target latent.
Specifically, we discretize the integration trajectory into $N$ steps, denoted by $1 = t_0 > t_1 > \dots > t_N = 0$. At step $i \in \{0, \dots, N-1\}$, the state is defined as $s_i \triangleq (c, t_i, Z_{t_i})$, comprising the multimodal condition $c$, the current integration time $t_i$, and the intermediate latent sequence $Z_{t_i}$. The action $a_i$ is derived from a numerical ODE step guided by the predicted velocity $v_\theta(Z_{t_i}, t_i, c)$. The transition policy is thus formalized as:
\begin{equation}
    \pi_{\theta}(a_i \mid s_i) \triangleq p_{\theta}(Z_{t_{i+1}} \mid Z_{t_i}, c),
\end{equation}
which is parameterized by the unified DiT backbone. We adopt an outcome reward formulation, assigning non-zero feedback exclusively at the end of the rollout:
\begin{equation}
    R(s_i, a_i) =
    \begin{cases}
        r\Bigl(\vaed(Z_{t_N}); c\Bigr), & i = N-1,\\
        0, & \text{otherwise}.
    \end{cases}
\end{equation}
% where $$ represents the sequence of final decoded RGBA images mapped back to the pixel space.

% \medskip
\noindent\textbf{Policy Optimization Algorithm.}
Motivated by DeepSeekMath~\citep{shao2024deepseekmathpushinglimitsmathematical} and DanceGRPO, for each condition $c$, we sample a group of $G$ rollouts $\{Z_{t_N}^{(g)}\}_{g=1}^{G}$ from the current policy $\pi_{\theta_{\text{old}}}$. We optimize the policy model $\pi_\theta$ by maximizing the following group-relative objective function:
\begin{equation}
    \mathcal{J}(\theta) = \mathbb{E}_{\{Z^{(g)}_{t_N}\}_{g=1}^G} %_{\substack{\{Z^{(g)}\}_{g=1}^G \sim \pi_{\theta_{\text{old}}}\\ a_i^{(g)} \sim \pi_{\theta_{\text{old}}}(\cdot \mid s_i^{(g)})}}
    \left[ \frac{1}{G\cdot N} \sum_{g=1}^{G} \sum_{i = 0}^{N-1} 
    % \min\left( \rho_i^{(g)} A_g,\, \text{clip}(\rho_i^{(g)}, 1-\epsilon, 1+\epsilon)A_g \right) 
    \mathcal{A}(A_g,\rho_i^{(g)};\epsilon)
    \right],
\end{equation}
where $\mathcal{A}(A,\rho;\epsilon)=\min\left( \rho A,\, \text{clip}(\rho, 1-\epsilon, 1+\epsilon)A \right) $.
The term $\rho_i^{(g)} = \frac{\pi_{\theta}(a_i^{(g)} \mid s_i^{(g)})}{\pi_{\theta_{\text{old}}}(a_i^{(g)} \mid s_i^{(g)})}$ denotes the importance sampling ratio, enabling off-policy optimization from the sampled trajectories.
To align the model without a value network, we compute the advantage $A_g$ by normalizing the outcome rewards $r_g = r\bigl(\vaed(Z_{t_N}^{(g)}); c\bigr)$ within each group:
\begin{equation}
    A_g = \frac{r_g - \mathrm{mean}(\{r_k\}_{k=1}^{G})}{\mathrm{std}(\{r_k\}_{k=1}^{G}) + \varepsilon}.
\end{equation}
By emphasizing relative quality, this formulation allows \method{} to effectively prioritize superior compositional structures and refine sparse visual details across diverse RGBA tasks.

\begin{figure}
    \centering
    \includegraphics[width=1\linewidth]{figs/rl_update.pdf}
    \vspace{-2.4em}
    \caption{Demonstration of Layer-Aware Reward Shaping.}
    \label{fig:rl}
\end{figure}

\medskip\noindent\textbf{Task-Agnostic Layer-Aware Rewards.} 
To preserve the structural generality of \method{}, we eschew task-specific reward engineering in favor of four universal objectives defined on the decoded RGBA sequence $\hat{Y}=(\hat y_1,\dots,\hat y_m)$, as shown in Figure~\ref{fig:rl}. Let $Y=(y_1,\dots,y_m)$ denote the ground-truth sequence, and $B(f_1,\cdots,f_k;b)$ represent the alpha-blending operator that composites an ordered sequence of foreground $f_1, \cdots, f_k$ over a background $b$. To evaluate transparency across diverse contexts, we introduce a background set $\mathcal{B} = \{\mathbf{0}, \mathbf{1}\}$, consisting of constant pure black and pure white images. For a given task $\tau$, let $\mathcal{K}=\{1,\dots,m\}$, and $\mathcal{K}_{\mathrm{fg}}, \mathcal{K}_{\mathrm{bg}} \subseteq \mathcal{K}$ denote the index sets of foreground and background layers, respectively.
The reward suite is formulated as follows:

\begin{itemize}
    \item \textbf{Layer Fidelity ($r_{\mathrm{layer}}$):} 
    Measures RGB similarity $\phi \in \{\mathrm{SSIM}, \mathrm{PSNR}\}$ across all valid layers:
    \begin{equation}\resizebox{.4\textwidth}{!}{$
    r_{\mathrm{layer}} =
    \frac{1}{|\mathcal{K}|} 
        \sum\limits_{k \in \mathcal{K}}
        \mathop{\mathbb{E}}\limits_{\phi,\mathbf{b}\in\mathcal{B}}\left[
        \frac{
            \phi\!\left(B(\hat{y}_{k};\mathbf{b}),\, B(y_{k};\mathbf{b})\right)
        }{
            \max_{\mathcal{I}_1,\mathcal{I}_2}\phi\!\left(\mathcal{I}_1,\mathcal{I}_2\right)
        }
        \right]
        .
    $}
    \end{equation}

    \item \textbf{Composition Fidelity ($r_{\mathrm{comp}}$):} Evaluates the globally composited prediction against full-image ground truth: 
    \begin{equation}\resizebox{.4\textwidth}{!}{$
    r_{\mathrm{comp}}
    = 
        % \frac{1}{|\mathcal{B}|} \sum_{\mathbf{b}\in\mathcal{B}} 
        \mathop{\mathbb{E}}\limits_{\phi,\mathbf{b}\in\mathcal{B}}\left[
        \frac{
            \phi\!\left(B(\hat y_1, \cdots, \hat y_m;\mathbf{b}),\, B(y_1, \cdots, y_m;\mathbf{b})\right)
        }{
            \max_{\mathcal{I}_1,\mathcal{I}_2}\phi\!\left(\mathcal{I}_1,\mathcal{I}_2\right)
        }
        \right].
    % \phi\!\left(B(\mathrm{Comp}(\hat{Y});\mathbf{b}),\, B(I^\star;\mathbf{b})\right).
    $}\end{equation}

    \item \textbf{Background Structure ($r_{\mathrm{bg}}$):} Penalizes structural deviations in background layers using LPIPS~\citep{zhang2018perceptual}:
    \begin{equation}\resizebox{.38\textwidth}{!}{$
    r_{\mathrm{bg}} = \frac{1}{(|\mathcal{K}_{\mathrm{bg}}| + \varepsilon)}
    \sum\limits_{k \in \mathcal{K}_{\mathrm{bg}}} 
    \mathop{\mathbb{E}}\limits_{\mathbf{b}\in\mathcal{B}} 
    \Bigl[ 1 - \mathrm{LPIPS}\!\left(B(\hat{y}_{k};\mathbf{b}),\, B(y_{k};\mathbf{b})\right) \Bigr].
    $}\end{equation}
    
    \item \textbf{Foreground Boundary ($r_{\mathrm{fg}}$):} 
    Enforces alpha boundary precision and recall for foreground layers:
    \begin{equation}\resizebox{.38\textwidth}{!}{$
    r_{\mathrm{fg}} = 
    \frac{1}{|\mathcal{K}_{\mathrm{fg}}| + \varepsilon} 
    \sum\limits_{k \in \mathcal{K}_{\mathrm{fg}}} 
    \left( 
        1 - \frac{
            \left\|\mathbf{m}_{\mathrm{bdry}}^{(k)} \odot (\hat{\alpha}_{k}-\alpha_{k})\right\|_1
        }{
            \left\|\mathbf{m}_{\mathrm{bdry}}^{(k)}\right\|_1 + \varepsilon
        } 
    \right),
    $}\end{equation}
    where $\hat{\alpha}_{k},\alpha_{k}$ represents the alpha channel of $\hat y_{k}, y_k$ respectively, 
    and $\mathbf{m}_{\mathrm{bdry}}^{(k)}$ denotes the soft boundary mask 
    constructed from both blurred alpha-band masks given the transparency threshold $\tau=[\tau_{\mathrm{low}},\tau_{\mathrm{high}}]$:
    % derived from the union of blurred boundary-band masks computed from both alpha maps:
    % [\tau_{\mathrm{low}},\tau_{\mathrm{high}}]
    \begin{equation}\resizebox{.38\textwidth}{!}{$
    \begin{split}
        \mathbf{m}_{\mathrm{bdry}}^{(k)}
        = \mathrm{clip}\Bigl(
        \mathrm{Blur}\!\left(
        \mathbb{I}\!\left[\hat{\alpha}_{k}\in\tau\right]
        \right)
        +
        \mathrm{Blur}\!\left(
        \mathbb{I}\!\left[\alpha_{k}\in\tau\right]
        \right),0,1
        \Bigr).
    \end{split}
    $}\end{equation}
\end{itemize}

\begin{table}[h]
\centering
\caption{Mapping of tasks to layer-aware reward components. $T$ denotes the text prompt, $x$ and $y$ denote the input and output layers respectively. $\mathcal{K}_{\mathrm{fg}}$ and $\mathcal{K}_{\mathrm{bg}}$ denote the indices of foreground and background layers in the output sequence.}
\label{tab:reward_mapping}
\setlength{\tabcolsep}{3pt} 
\resizebox{\columnwidth}{!}{%
\begin{tabular}{lcccccccc}
\toprule
Task & Input & Output & $\mathcal{K}_{\mathrm{fg}}$ & $\mathcal{K}_{\mathrm{bg}}$ & $\chi_{\mathrm{layer}}$ & $\chi_{\mathrm{comp}}$ & $\chi_{\mathrm{bg}}$ & $\chi_{\mathrm{fg}}$ \\
\midrule
Text-to-Image Generation & $T$ & $y_1$ & $\{1\}$ & $\emptyset$ & 0 & 0 & 0 & 0 \\
Object Removal & $x_1, x_2$ & $y_1$ & $\emptyset$ & $\{1\}$ & 1 & 0 & 1 & 0 \\
Referring Matting & $T, x_1$ & $y_1$ & $\{1\}$ & $\emptyset$ & 1 & 0 & 0 & 1 \\
Automatic Matting & $x_1$ & $y_1$ & $\{1\}$ & $\emptyset$ & 1 & 0 & 0 & 1 \\
Layer Decomposition & $x_1$ & $y_1, y_2$ & $\{2\}$ & $\{1\}$ & 1 & 1 & 1 & 1 \\
\bottomrule
\end{tabular}
}
\end{table}

\medskip\noindent\textbf{Adaptive Reward Aggregation.}
To handle variable output structures across diverse RGBA tasks, we introduce a task-specific indicator vector $\boldsymbol{\chi}_\tau = [\chi_{\mathrm{layer}}, \chi_{\mathrm{comp}}, \chi_{\mathrm{bg}}, \chi_{\mathrm{fg}}]^\top \in \{0, 1\}^4$, where $\chi_j=1$ if the $j$-th reward component is applicable to task $\tau$. As detailed in Table~\ref{tab:reward_mapping}, \method{} dynamically activates these reward terms based on the specific multi-layer requirements of each supervised task (e.g., setting $\chi_{\mathrm{bg}}=1$ only if $\mathcal{K}_{\mathrm{bg}} \neq \emptyset$). In particular, since text-to-image generation lacks objective, pixel-aligned ground-truth references, we set $\boldsymbol{\chi}_{\mathrm{T2I}} = \mathbf{0}$, effectively bypassing the RL phase for this task. 
For each sampled trajectory, we independently apply Z-score normalization to all active reward terms to align their scales, yielding normalized scores $\tilde{r}_j$. The final trajectory-level outcome reward is then computed as a normalized weighted sum:
\begin{equation}
r = \frac{\sum_{j \in \mathcal{R}} \lambda_{j} \chi_{j} \tilde{r}_{j}}{\sum_{j \in \mathcal{R}} \lambda_{j} \chi_{j} + \varepsilon},
\end{equation}
where $\mathcal{R} = \{\mathrm{layer, comp, bg, fg}\}$ and $\lambda_j$ are the corresponding weights. In our implementation, we weight all enabled components equally, setting $\lambda_{\mathrm{layer}} = \lambda_{\mathrm{comp}} = \lambda_{\mathrm{bg}} = \lambda_{\mathrm{fg}} = 1.0$.

\begin{table*}[t]
  \centering
  \caption{Object removal results on RORD-Val, OBER-Test, and OBER-Wild. }
  % \vspace{.4em}
  \label{tab:object_remove}
  \renewcommand{\arraystretch}{1.2}
  \resizebox{.85\linewidth}{!}{
  \begin{tabular}{lccccccccc}
    \toprule
    \multirow{2.5}{*}{\textbf{Method}} &
    \multicolumn{4}{c}{\textbf{RORD-Val}} &
    \multicolumn{4}{c}{\textbf{OBER-Test}} &
    \multicolumn{1}{c}{\textbf{OBER-Wild}} \\
    \cmidrule(lr){2-5} \cmidrule(lr){6-9} \cmidrule(lr){10-10}
    & PSNR $\uparrow$ & PSNR-BG $\uparrow$ & LPIPS $\downarrow$ & CLIP $\downarrow$
    & PSNR $\uparrow$ & PSNR-BG $\uparrow$ & LPIPS $\downarrow$ & CLIP $\downarrow$
    & ReMOVE $\uparrow$ \\
    \midrule
    SDXL-INP~\citep{podell2023sdxlimprovinglatentdiffusion}        & 20.230 & 24.830 & 0.204 & 0.087 & 22.420 & 25.770 & 0.143 & 0.077 & 0.804 \\
    PowerPaint~\citep{zhuang2024taskworthwordlearning}      & 21.460 & 24.620 & 0.180 & 0.065 & 22.760 & 24.670 & 0.154 & 0.073 & 0.854 \\
    Attentive Eraser~\citep{sun2025attentiveeraserunleashingdiffusion}& 22.170 & 24.590 & 0.188 & 0.064 & 25.700 & 27.080 & 0.120 & 0.044 & 0.901 \\
    RORem~\citep{li2025roremtrainingrobustobject}           & 22.490 & 24.100 & 0.294 & 0.063 & 24.510 & 25.280 & 0.129 & 0.046 & 0.903 \\
    \midrule
    ObjectClear~\citep{zhao2026preciseobjecteffectremoval}     & \textbf{26.400} & \textbf{30.750} & \textbf{0.080} & \textbf{0.035} & \textbf{32.980} & \textbf{35.140} & \textbf{0.035} & \textbf{0.010} & \textbf{0.917} \\
    \midrule
    \textbf{Ours (SFT)}      & 26.214 & 29.005 & \uline{0.141} & \uline{0.036} & 30.736 & 32.315 & 0.098 & 0.019 & 0.914 \\
    \textbf{Ours (RL)}       & \uline{26.315} & \uline{29.367} & 0.147 & 0.046 & \uline{31.389} & \uline{32.903} & \uline{0.095} & \uline{0.016} & \uline{0.915}  \\
    \bottomrule
  \end{tabular}
  }
\end{table*}

\section{Experiments}
\label{sec:exp}

\subsection{Setup}

\begin{figure}[h]
    \centering
    \includegraphics[width=0.95\linewidth]{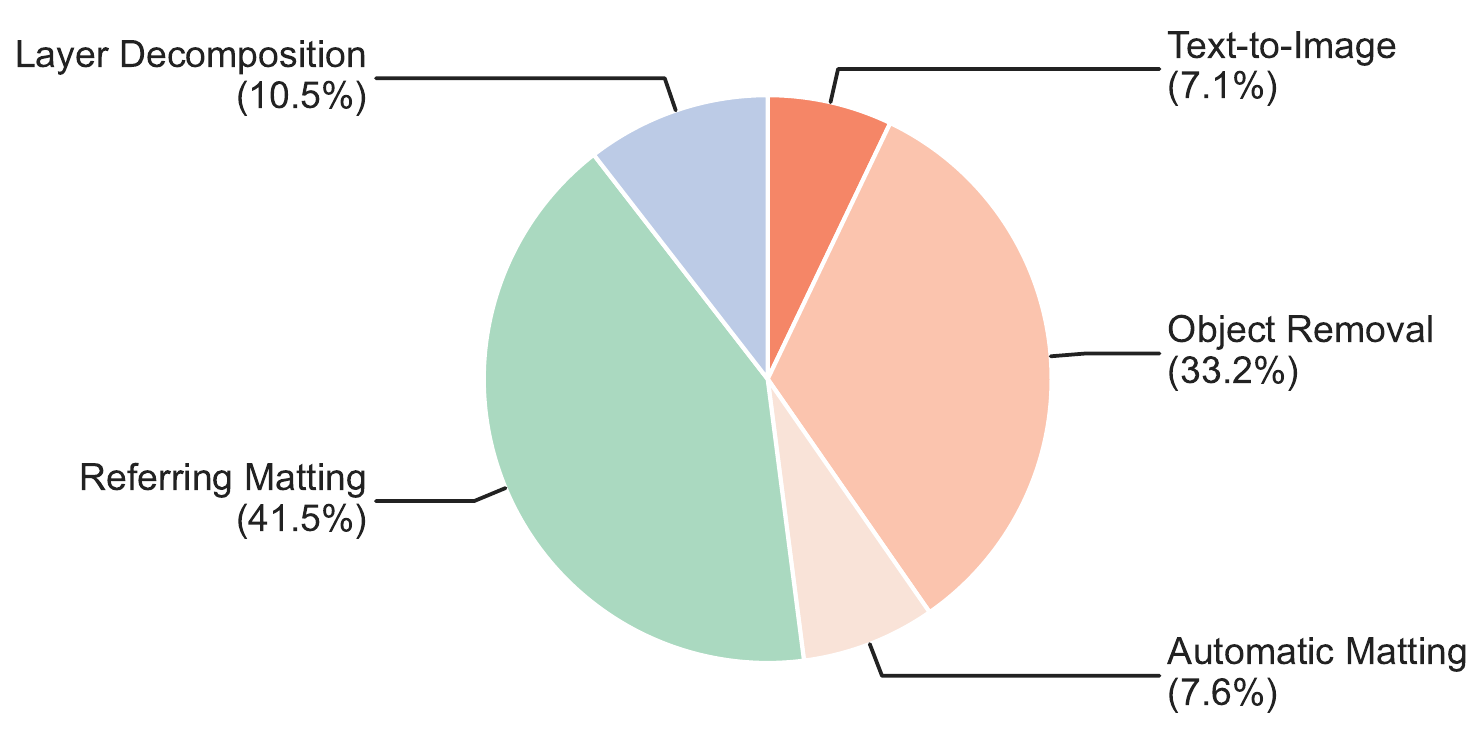}
    \vspace{-1.6em}
    \caption{Distribution of the training data across the five tasks in OmniAlpha. The numbers of training samples for Text-to-Image, Object Removal, Automatic Matting, Referring Matting, and Layer Decomposition are 8,124, 37,994, 8,710, 47,500, and 12,000, respectively.}
    \label{fig:task_distribution_pie}
\end{figure}

\noindent\textbf{Preparation of Training Data.}
To support both supervised fine-tuning and reinforcement learning, we construct a unified layer-aware training corpus covering the included tasks.
For text-to-image generation, we follow AlphaVAE to construct 8,124 high-quality RGBA images using the same data preparation pipeline. Specifically, the data are derived from the same ten public image matting datasets used in AlphaVAE and converted into four-channel RGBA format by combining each foreground image with its corresponding alpha matte.
For object removal, we use the training split of OBER-Dataset, which is introduced in ObjectClear~\citep{zhao2026preciseobjecteffectremoval}. 
For automatic matting, we adopt the data construction protocol of AIM~\citep{li2021deepautomaticnaturalimage} and construct a training set of 8,710 images. The data are derived from Composition-1k~\citep{xu2017deepimagematting}, HAtt~\citep{Qiao_2020_CVPR}, and AM-2k~\citep{li2021bridgingcompositerealendtoend}, together with BG-20k~\citep{li2021bridgingcompositerealendtoend} backgrounds under the composition strategy of AIM. 
For referring matting, we adopt RefMatte~\citep{li2023referringimagematting}, which contains 47,500 images and 118,749 expression-region pairs with high-quality alpha mattes.
For layer decomposition, we use PrismLayersReal and PrismLayersPlus~\citep{chen2025prismlayersopendatahighquality}.
Among the approximately 100,000 cases provided by the two datasets as shown in Figure~\ref{fig:task_distribution_pie}, we uniformly sample 12,000 cases for our training.

\noindent\textbf{Training Implementation.}
We adopt Qwen-Image-Edit~\citep{wu2025qwenimagetechnicalreport} as our base model and employ a three-stage training paradigm.
We first adapt the pre-trained RGB VAE from Qwen-Image-Edit into an alpha-aware (RGBA) VAE by fine-tuning it on our RGBA image dataset for 32k steps. This stage uses a global batch size of 16 and the AdamW optimizer~\citep{loshchilov2017decoupled} ($\beta_1{=}0.9, \beta_2{=}0.999$). We use a base learning rate of $1.5\times10^{-5}$ with a 5\% linear warmup, followed by a cosine decay schedule.
With the fine-tuned RGBA VAE weights frozen, we train the DiT backbone, also initialized from Qwen-Image-Edit, for 15k cold-start steps and 150 RL steps. This stage uses a batch size of 64 with the AdamW optimizer ($\beta_1{=}0.9, \beta_2{=}0.999$). For parameter-efficient fine-tuning, we apply LoRA~\citep{hu2021loralowrankadaptationlarge} with a rank of 256 to all attention weights and MLP layers, using a constant learning rate of $5\times10^{-5}$.
All models are trained on 64 NVIDIA H20 GPUs.

\subsection{Main Experimental Results}
\begin{figure*}[h]
    \centering
    \includegraphics[width=0.93\linewidth]{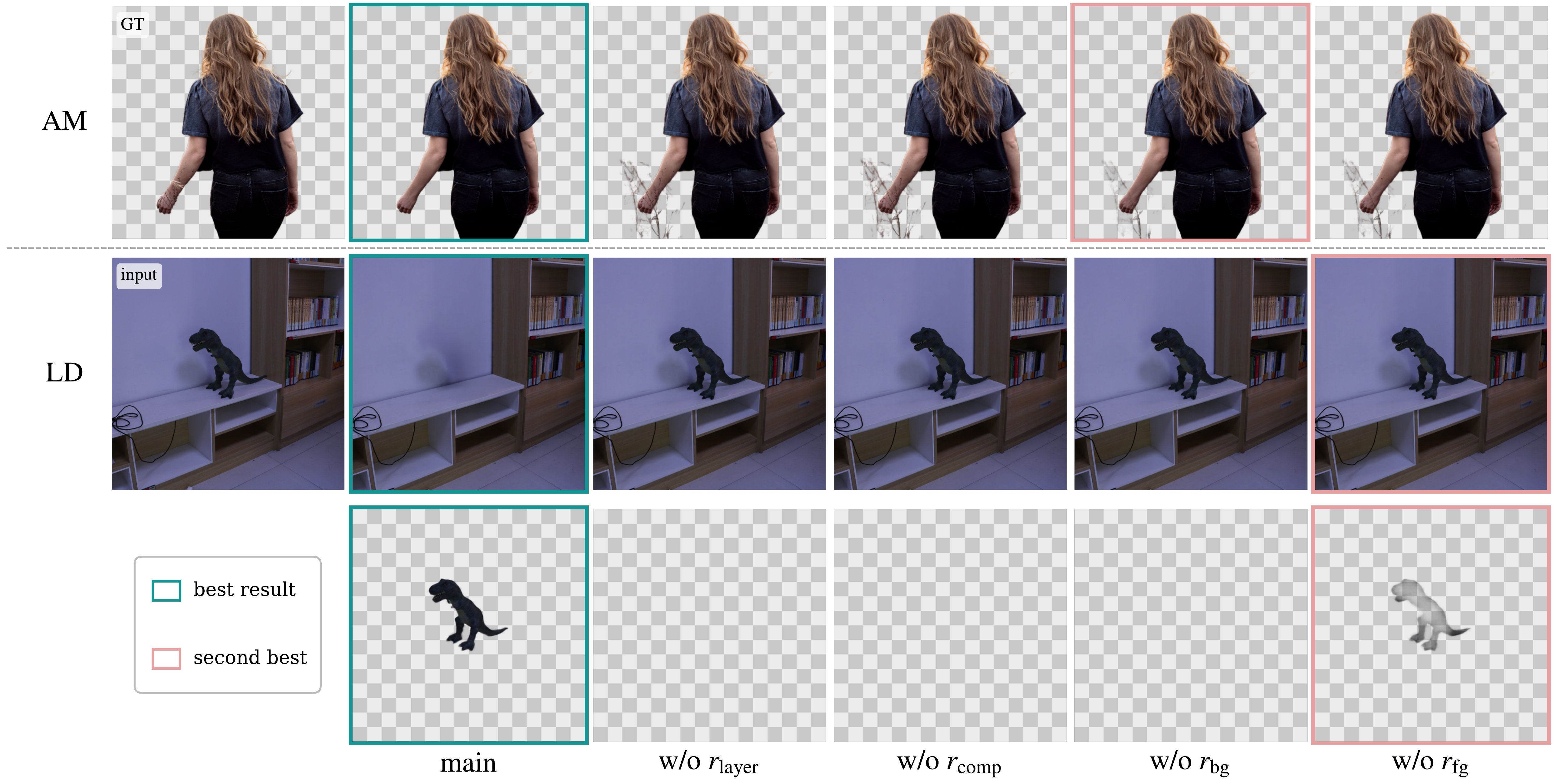}
    \vspace{-1.2em}
    \caption{Qualitative ablation cases on Automatic Matting (AM) and Layer Decomposition (LD). The legend entries ''best result'' and ''second best'' indicate the variants with the highest and second-highest metric values, respectively.  Removing individual reward terms leads to visible performance degradation on both tasks.}
    \label{fig:ablation_case}
\end{figure*}

\method{} is evaluated on five transparency-aware tasks, including text-to-image generation, referring matting, object removal, automatic matting, and layer decomposition, using both public benchmarks and our curated test split. Across these heterogeneous settings, the model delivers state-of-the-art or competitive performance against strong task-specific baselines, with especially clear advantages on transparency-sensitive tasks such as referring  and automatic matting, while remaining robust on synthesis, removal, and decomposition. Overall, the results indicate that a shared layer-aware RGBA representation transfers effectively across tasks within a single unified architecture, and that RL further strengthens the model on tasks requiring stronger structural alignment and perceptual fidelity.

\begin{table}[h]
  \centering
  \caption{Text-to-image results on AIM-500.}
  % \vspace{.4em}
  \footnotesize
  \label{tab:t2i_aim500}
  \renewcommand{\arraystretch}{1.2}
  \resizebox{.45\textwidth}{!}{
  \begin{tabular}{lcccc}
    \toprule
    \multirow{2.5}{*}{\textbf{Method}} &
    \multicolumn{4}{c}{\textbf{AIM-500}} \\
    \cmidrule(lr){2-5}
    & FID $\downarrow$ & CLIP-Score $\uparrow$ & LAION-AES $\uparrow$ & ImageReward $\uparrow$ \\
    \midrule
    GT & - & 0.837 & 5.234 & 0.611 \\
    \midrule
    LayerDiffuse~\citep{zhang2024transparentimagelayerdiffusion} & 78.499 & 0.836 & \uline{5.405} & 0.853 \\
    AlphaVAE~\citep{wang2025alphavae0} & 72.080 & 0.842 & \textbf{5.453} & 0.931 \\
    \midrule
    \textbf{Ours (SFT)} & \textbf{67.891} & \textbf{0.852} & 5.317 & \uline{1.050} \\
    \textbf{Ours (RL)} & \uline{68.829} & \uline{0.850} & 5.323 & \textbf{1.060} \\
    \bottomrule
  \end{tabular}
  }
\end{table}

\medskip\noindent\textbf{Text-to-Image.}
Table~\ref{tab:t2i_aim500} reports transparent text-to-image generation results on AIM-500~\citep{li2021deepautomaticnaturalimage}. Compared with specialized baselines such as LayerDiffuse~\citep{zhang2024transparentimagelayerdiffusion} and AlphaVAE~\citep{wang2025alphavae0}, our unified \method{} remains highly competitive. The SFT variant achieves the best FID (67.891) and CLIP-Score~\citep{hessel2022clipscorereferencefreeevaluationmetric} (0.852), while the RL variant, despite unseen during RL post-training, still achieves the best ImageReward~\citep{xu2023imagereward} (1.060). This shows that \method{} generalizes well without task-specific architectures or alignment tax.

\medskip\noindent\textbf{Referring Matting.}
Referring matting is evaluated on RefMatte-RW100~\citep{li2023referringimagematting} against specialized baselines such as CLIPMat~\citep{li2023referringimagematting}. As shown in Table~\ref{tab:text_guided_matting}, both variants of \method{} outperform all prior methods by a large margin. In particular, RL further improves upon SFT and achieves the best results on all three metrics, with 14.768 SAD, 0.022 MSE, and 0.026 MAD. These results demonstrate that \method{} can accurately localize the referred target and recover high-quality alpha mattes directly from language.

\begin{table}[h]
  \centering
  \caption{Referring matting results on RefMatte-RW100.}
  % \vspace{.4em}
  \small
  \label{tab:text_guided_matting}
  \renewcommand{\arraystretch}{1.2}
  \begin{tabular}{l >{\centering\arraybackslash}p{1.6cm} >{\centering\arraybackslash}p{1.6cm} >{\centering\arraybackslash}p{1.6cm}}
    \toprule
    \multirow{2.5}{*}{\textbf{Method}} &
    \multicolumn{3}{c}{\textbf{RefMatte-RW100}} \\
    \cmidrule(lr){2-4}
    & SAD $\downarrow$ & MSE $\downarrow$ & MAD $\downarrow$ \\
    \midrule
    MDETR~\citep{kamath2021mdetrmodulateddetection}   & 131.580 & 0.068 & 0.075 \\
    CLIPSeg~\citep{luddecke2022imagesegmentationusingtext} & 211.860 & 0.118 & 0.122 \\
    MAM~\citep{li2023matting}     & 120.100 & 0.066 & 0.068 \\
    CLIPMat~\citep{li2023referringimagematting} & 85.830  & 0.047 & 0.050 \\
    \midrule
    \textbf{Ours (SFT)} & \underline{15.029} & \underline{0.027} & \underline{0.030} \\
    \textbf{Ours (RL)}  & \textbf{14.768} & \textbf{0.022} & \textbf{0.026} \\
    \bottomrule
  \end{tabular}
\end{table}

\medskip\noindent\textbf{Object Removal.}
Object removal is evaluated on RORD-Val, OBER-Test, and OBER-Wild against representative specialized baselines. Table~\ref{tab:object_remove} shows that both our SFT and RL models consistently outperform a wide range of established methods, while RL further improves over SFT on the more challenging OBER-Test benchmark and on the out-of-domain OBER-Wild set. Although the highly specialized ObjectClear model achieves the best overall performance, our unified approach remains highly competitive across benchmarks. On OBER-Wild, SFT and RL achieve 0.914 and 0.915 ReMOVE, respectively, remaining close to the task-specific ObjectClear (0.917), which indicates strong generalization under distribution shift.

\medskip\noindent\textbf{Automatic Matting.}
Automatic matting is evaluated on AIM-500~\citep{li2021deepautomaticnaturalimage} in the mask-free setting, where the model directly identifies the visible subject and predicts its alpha matte without trimap or mask guidance. Table~\ref{tab:auto_matting} shows that both variants outperform all prior methods by a large margin. RL achieves the best overall performance, while SFT obtains the second best score. Compared with the strongest baseline, SMat~\citep{ye2024unifying}, \method{} significantly improves foreground localization and alpha reconstruction.

\begin{table}[h]
  \centering
  \caption{Automatic matting results on AIM-500. }
  % \vspace{.4em}
  \label{tab:auto_matting}
  \renewcommand{\arraystretch}{1.2}
  \resizebox{.45\textwidth}{!}{
  \begin{tabular}{lccccc}
    \toprule
    \multirow{2.5}{*}{\textbf{Method}} &
    \multicolumn{5}{c}{\textbf{AIM-500}} \\
    \cmidrule(lr){2-6}
    & SAD $\downarrow$ & MSE $\downarrow$ & MAD $\downarrow$ & Grad $\downarrow$ & Conn $\downarrow$ \\
    \midrule
    GFM~\citep{li2021bridgingcompositerealendtoend}  & 52.660 & 0.0213 & 0.0313 & 46.110 & 52.690 \\
    AIM~\citep{li2021deepautomaticnaturalimage}  & 48.090 & 0.0183 & 0.0285 & 47.580 & 21.740 \\
    SMat~\citep{ye2024unifying} & 34.300 & 0.0129 & 0.0203 & 31.490 & 13.980 \\
    \midrule
    \textbf{Ours (SFT)} & \uline{9.245} & \uline{0.0109} & \uline{0.0185} & \uline{10.677} & \uline{9.071} \\
    \textbf{Ours (RL)}  & \textbf{9.089} & \textbf{0.0105} & \textbf{0.0182} & \textbf{10.099} & \textbf{8.930} \\
    \bottomrule
  \end{tabular}
  }
\end{table}

\medskip\noindent\textbf{Layer Decomposition.}
Layer Decomposition is evaluated on OBER-Test-decompose, derived from OBER-Test~\cite{zhao2026preciseobjecteffectremoval}, where each composed image is decomposed into editable foreground and background layers with explicit transparency. Table~\ref{tab:decompose_ober} shows that RL achieves the best Alpha soft IoU (0.7069), while SFT also attains a strong result of 0.6831, outperforming both LayerD~\citep{suzuki2025layerddecomposingrastergraphic} and Qwen-Image-Layered~\citep{yin2025qwenimagelayeredinherenteditabilitylayer} on this metric. In terms of RGB L1, LayerD remains the strongest, and Qwen-Image-Layered also outperforms our model. Nevertheless, RL substantially improves over SFT, indicating that RL post-training improves reconstruction quality while further strengthening alpha-aware decomposition.

\begin{table}[h]
  \centering
  \caption{Decomposition results on OBER-Test-decompose.}
  % \vspace{.4em}
  \label{tab:decompose_ober}
  \small
  \renewcommand{\arraystretch}{1.2}
  \resizebox{.95\linewidth}{!}{
  \begin{tabular}{l >{\centering\arraybackslash}p{2.0cm} >{\centering\arraybackslash}p{2.2cm}}
    \toprule
    \multirow{2.5}{*}{\textbf{Method}} &
    \multicolumn{2}{c}{\textbf{OBER-Test-decompose}} \\
    \cmidrule(lr){2-3}
    & RGB L1 $\downarrow$ & Alpha soft IoU $\uparrow$ \\
    \midrule
    LayerD~\citep{suzuki2025layerddecomposingrastergraphic} & \textbf{0.0173} & 0.5645 \\
    Qwen-Image-Layered~\citep{yin2025qwenimagelayeredinherenteditabilitylayer} & \underline{0.0977} & 0.6589 \\
    \midrule
    \textbf{Ours (SFT)} & 0.1135 & \underline{0.6831} \\
    \textbf{Ours (RL)}  & 0.1032 & \textbf{0.7069} \\
    \bottomrule
  \end{tabular}
  }
\end{table}

\subsection{Ablation Study}

% To better understand the role of each component in OmniAlpha, we conduct ablation studies on the major design choices, with the results shown in Figure~\ref{fig:ablation_case}.

To better understand the role of each component in OmniAlpha, we perform reward ablations on all RL objectives. 
% The results show that removing any reward term consistently degrades performance, validating the effectiveness of the full reward design.

\begin{table}[h]
  \centering
  \caption{Ablation study on text-to-image generation.}
  \label{tab:ablation_aim500}
  \small
  \renewcommand{\arraystretch}{0.9}
  \setlength{\tabcolsep}{2.5pt}
  \resizebox{\linewidth}{!}{
  \begin{tabular}{>{\raggedright\arraybackslash}m{2.2cm}cccc}
    \toprule
    \multirow{2}{*}{\textbf{Method}} &
    \multicolumn{4}{c}{\textbf{AIM-500}} \\
    \cmidrule(lr){2-5}
    & FID $\downarrow$ & CLIP-Score $\uparrow$ & LAION-AES $\uparrow$ & ImageReward $\uparrow$ \\
    \midrule
    OmniAlpha RL
    & \textbf{68.829}
    & \textbf{0.850}
    & \smash[b]{\underline{5.323}}
    & \textbf{1.060} \\
    [-0.17em]
    \midrule
    
    \raisebox{1.5ex}{w/o $r_{\text{layer}}$}
    & \shortstack[c]{\strut 69.663\\[-0.45em]\scriptsize{\color{mypurple}(+0.834)}}
    & \shortstack[c]{\strut 0.846\\[-0.45em]\scriptsize{\color{mypurple}(-0.004)}}
    & \shortstack[c]{\strut 5.315\\[-0.45em]\scriptsize{\color{mypurple}(-0.008)}}
    & \shortstack[c]{\strut 1.056\\[-0.45em]\scriptsize{\color{mypurple}(-0.004)}} \\[-.4em]

    \raisebox{1.5ex}{w/o $r_{\text{fg}}$}
    & \shortstack[c]{\strut \smash[b]{\underline{68.904}}\\[-0.45em]\scriptsize{\color{mypurple}(+0.075)}}
    & \shortstack[c]{\strut 0.848\\[-0.45em]\scriptsize{\color{mypurple}(-0.002)}}
    & \shortstack[c]{\strut \textbf{5.328}\\[-0.45em]\scriptsize{\color{deepgreen}(+0.005)}}
    & \shortstack[c]{\strut \smash[b]{\underline{1.059}}\\[-0.45em]\scriptsize{\color{mypurple}(-0.001)}} \\[-.4em]

    \raisebox{1.5ex}{w/o $r_{\text{bg}}$}
    & \shortstack[c]{\strut 69.558\\[-0.45em]\scriptsize{\color{mypurple}(+0.729)}}
    & \shortstack[c]{\strut 0.848\\[-0.45em]\scriptsize{\color{mypurple}(-0.002)}}
    & \shortstack[c]{\strut 5.313\\[-0.45em]\scriptsize{\color{mypurple}(-0.010)}}
    & \shortstack[c]{\strut 1.054\\[-0.45em]\scriptsize{\color{mypurple}(-0.006)}} \\[-.4em]

    \raisebox{1.5ex}{w/o $r_{\text{comp}}$}
    & \shortstack[c]{\strut 69.627\\[-0.45em]\scriptsize{\color{mypurple}(+0.798)}}
    & \shortstack[c]{\strut \smash[b]{\underline{0.849}}\\[-0.45em]\scriptsize{\color{mypurple}(-0.001)}}
    & \shortstack[c]{\strut 5.317\\[-0.45em]\scriptsize{\color{mypurple}(-0.006)}}
    & \shortstack[c]{\strut 1.058\\[-0.45em]\scriptsize{\color{mypurple}(-0.002)}} \\[-.4em]
    \bottomrule
  \end{tabular}
  }
\end{table}

\medskip\noindent\textbf{Text-to-Image Generation.} Table~\ref{tab:ablation_aim500} shows that the full RL model performs best overall on AIM-500~\citep{li2021deepautomaticnaturalimage}, achieving the best FID, CLIP-Score~\citep{hessel2022clipscorereferencefreeevaluationmetric}, and ImageReward~\citep{xu2023imagereward}. Removing any reward term degrades performance, confirming the effectiveness of the full reward design. In particular, $\mathrm{r}_{\text{layer}}$ is the most critical, while the degradation caused by removing $\mathrm{r}_{\text{bg}}$ and $\mathrm{r}_{\text{comp}}$ highlights the importance of boundary-background interaction and compositional realism in transparent image generation.

\medskip\noindent\textbf{Referring Matting.}
Table~\ref{tab:ablation_refmatte} shows that the full RL model achieves the best performance on RefMatte-RW100 across all metrics, confirming the effectiveness of the full reward design for text-guided alpha prediction. In particular, $\mathrm{r}_{\text{layer}}$ and $\mathrm{r}_{\text{fg}}$ are the most critical, as their removal causes the largest drops, while the degradation caused by removing $\mathrm{r}_{\text{bg}}$ is also substantial. Although removing $\mathrm{r}_{\text{comp}}$ has a relatively smaller effect, it remains consistently worse than the full model.

\begin{table}[h]
  \centering
  \caption{Ablation study on RefMatte-RW100.}
  \label{tab:ablation_refmatte}
  \small
  \renewcommand{\arraystretch}{0.85}
  \setlength{\tabcolsep}{2.5pt}
  \resizebox{.9\linewidth}{!}{
  \begin{tabular}{>{\raggedright\arraybackslash}m{2.35cm}
                  >{\centering\arraybackslash}p{1.25cm}
                  >{\centering\arraybackslash}p{1.25cm}
                  >{\centering\arraybackslash}p{1.25cm}}
    \toprule
    \multirow{2}{*}{\textbf{Method}} &
    \multicolumn{3}{c}{\textbf{RefMatte-RW100}} \\
    \cmidrule(lr){2-4}
    & SAD $\downarrow$ & MSE $\downarrow$ & MAD $\downarrow$ \\
    \midrule
    OmniAlpha RL
    & \textbf{14.768}
    & \textbf{0.0217}
    & \textbf{0.0257} \\
    [-0.17em]
    \midrule

    \raisebox{1.5ex}{w/o $r_{\text{layer}}$}
    & \shortstack[c]{\strut 16.251\\[-0.45em]\scriptsize{\color{mypurple}(+1.483)}}
    & \shortstack[c]{\strut 0.0307\\[-0.45em]\scriptsize{\color{mypurple}(+0.0090)}}
    & \shortstack[c]{\strut 0.0326\\[-0.45em]\scriptsize{\color{mypurple}(+0.0069)}} \\[-.4em]

    \raisebox{1.5ex}{w/o $r_{\text{fg}}$}
    & \shortstack[c]{\strut 16.273\\[-0.45em]\scriptsize{\color{mypurple}(+1.505)}}
    & \shortstack[c]{\strut 0.0305\\[-0.45em]\scriptsize{\color{mypurple}(+0.0088)}}
    & \shortstack[c]{\strut 0.0327\\[-0.45em]\scriptsize{\color{mypurple}(+0.0070)}} \\[-.4em]

    \raisebox{1.5ex}{w/o $r_{\text{bg}}$}
    & \shortstack[c]{\strut 16.268\\[-0.45em]\scriptsize{\color{mypurple}(+1.500)}}
    & \shortstack[c]{\strut 0.0307\\[-0.45em]\scriptsize{\color{mypurple}(+0.0090)}}
    & \shortstack[c]{\strut 0.0327\\[-0.45em]\scriptsize{\color{mypurple}(+0.0070)}} \\[-.4em]

    \raisebox{1.5ex}{w/o $r_{\text{comp}}$}
    & \shortstack[c]{\strut \smash[b]{\underline{16.048}}\\[-0.45em]\scriptsize{\color{mypurple}(+1.280)}}
    & \shortstack[c]{\strut \smash[b]{\underline{0.0303}}\\[-0.45em]\scriptsize{\color{mypurple}(+0.0086)}}
    & \shortstack[c]{\strut \smash[b]{\underline{0.0322}}\\[-0.45em]\scriptsize{\color{mypurple}(+0.0065)}} \\[-.4em]
    \bottomrule
  \end{tabular}
  }
\end{table}

\medskip\noindent\textbf{Object Removal.}
Table~\ref{tab:ablation_object_remove} shows that the full RL model performs best overall on object removal, confirming the effectiveness of the full reward design. Removing any reward term degrades performance, with $\mathrm{r}_{\text{layer}}$ and $\mathrm{r}_{\text{bg}}$ being the most critical. Removing $\mathrm{r}_{\text{fg}}$ also causes consistent degradation, whereas removing $\mathrm{r}_{\text{comp}}$ has a relatively smaller effect but is still worse than the full model on the paired benchmarks.

\begin{table*}[htb]
  \centering
  \caption{Ablation study on object removal.}
  \label{tab:ablation_object_remove}
  \normalsize
  \renewcommand{\arraystretch}{0.9}
  \setlength{\tabcolsep}{2.8pt}
  \resizebox{.9\linewidth}{!}{
  \begin{tabular}{>{\raggedright\arraybackslash}m{2.2cm}ccccccccc}
    \toprule
    \multirow{2}{*}{\textbf{Method}} &
    \multicolumn{4}{c}{\textbf{RORD-Val}} &
    \multicolumn{4}{c}{\textbf{OBER-Test}} &
    \multicolumn{1}{c}{\textbf{OBER-Wild}} \\
    \cmidrule(lr){2-5} \cmidrule(lr){6-9} \cmidrule(lr){10-10}
    & PSNR $\uparrow$ & PSNR-BG $\uparrow$ & LPIPS $\downarrow$ & CLIP $\downarrow$
    & PSNR $\uparrow$ & PSNR-BG $\uparrow$ & LPIPS $\downarrow$ & CLIP $\downarrow$
    & ReMOVE $\uparrow$ \\
    \midrule
    OmniAlpha RL
    & \textbf{26.315}
    & \textbf{29.367}
    & \textbf{0.1467}
    & \textbf{0.0460}
    & \textbf{31.389}
    & \textbf{32.903}
    & \textbf{0.0946}
    & \textbf{0.0157}
    & \textbf{0.91481} \\
    [-0.17em]
    \midrule

    \raisebox{1.5ex}{w/o $r_{\text{layer}}$}
    & \shortstack[c]{\strut 26.048\\[-0.45em]\scriptsize{\color{mypurple}(-0.267)}}
    & \shortstack[c]{\strut 29.142\\[-0.45em]\scriptsize{\color{mypurple}(-0.225)}}
    & \shortstack[c]{\strut \smash[b]{\underline{0.1469}}\\[-0.45em]\scriptsize{\color{mypurple}(+0.0002)}}
    & \shortstack[c]{\strut 0.0466\\[-0.45em]\scriptsize{\color{mypurple}(+0.0006)}}
    & \shortstack[c]{\strut 31.193\\[-0.45em]\scriptsize{\color{mypurple}(-0.196)}}
    & \shortstack[c]{\strut 32.719\\[-0.45em]\scriptsize{\color{mypurple}(-0.184)}}
    & \shortstack[c]{\strut \smash[b]{\underline{0.0948}}\\[-0.45em]\scriptsize{\color{mypurple}(+0.0002)}}
    & \shortstack[c]{\strut 0.0160\\[-0.45em]\scriptsize{\color{mypurple}(+0.0003)}}
    & \shortstack[c]{\strut 0.91377\\[-0.45em]\scriptsize{\color{mypurple}(-0.00104)}} \\[-.4em]

    \raisebox{1.5ex}{w/o $r_{\text{fg}}$}
    & \shortstack[c]{\strut 26.032\\[-0.45em]\scriptsize{\color{mypurple}(-0.283)}}
    & \shortstack[c]{\strut 29.120\\[-0.45em]\scriptsize{\color{mypurple}(-0.247)}}
    & \shortstack[c]{\strut 0.1471\\[-0.45em]\scriptsize{\color{mypurple}(+0.0004)}}
    & \shortstack[c]{\strut 0.0468\\[-0.45em]\scriptsize{\color{mypurple}(+0.0008)}}
    & \shortstack[c]{\strut 31.243\\[-0.45em]\scriptsize{\color{mypurple}(-0.146)}}
    & \shortstack[c]{\strut 32.778\\[-0.45em]\scriptsize{\color{mypurple}(-0.125)}}
    & \shortstack[c]{\strut 0.0950\\[-0.45em]\scriptsize{\color{mypurple}(+0.0004)}}
    & \shortstack[c]{\strut 0.0160\\[-0.45em]\scriptsize{\color{mypurple}(+0.0003)}}
    & \shortstack[c]{\strut 0.91372\\[-0.45em]\scriptsize{\color{mypurple}(-0.00109)}} \\[-.4em]

    \raisebox{1.5ex}{w/o $r_{\text{bg}}$}
    & \shortstack[c]{\strut 26.020\\[-0.45em]\scriptsize{\color{mypurple}(-0.295)}}
    & \shortstack[c]{\strut 29.106\\[-0.45em]\scriptsize{\color{mypurple}(-0.261)}}
    & \shortstack[c]{\strut 0.1477\\[-0.45em]\scriptsize{\color{mypurple}(+0.0010)}}
    & \shortstack[c]{\strut 0.0464\\[-0.45em]\scriptsize{\color{mypurple}(+0.0004)}}
    & \shortstack[c]{\strut 31.263\\[-0.45em]\scriptsize{\color{mypurple}(-0.126)}}
    & \shortstack[c]{\strut 32.804\\[-0.45em]\scriptsize{\color{mypurple}(-0.099)}}
    & \shortstack[c]{\strut 0.0953\\[-0.45em]\scriptsize{\color{mypurple}(+0.0007)}}
    & \shortstack[c]{\strut \smash[b]{\underline{0.0158}}\\[-0.45em]\scriptsize{\color{mypurple}(+0.0001)}}
    & \shortstack[c]{\strut 0.91376\\[-0.45em]\scriptsize{\color{mypurple}(-0.00105)}} \\[-.4em]

    \raisebox{1.5ex}{w/o $r_{\text{comp}}$}
    & \shortstack[c]{\strut \smash[b]{\underline{26.058}}\\[-0.45em]\scriptsize{\color{mypurple}(-0.257)}}
    & \shortstack[c]{\strut \smash[b]{\underline{29.142}}\\[-0.45em]\scriptsize{\color{mypurple}(-0.225)}}
    & \shortstack[c]{\strut 0.1473\\[-0.45em]\scriptsize{\color{mypurple}(+0.0006)}}
    & \shortstack[c]{\strut \smash[b]{\underline{0.0462}}\\[-0.45em]\scriptsize{\color{mypurple}(+0.0002)}}
    & \shortstack[c]{\strut \smash[b]{\underline{31.269}}\\[-0.45em]\scriptsize{\color{mypurple}(-0.120)}}
    & \shortstack[c]{\strut \smash[b]{\underline{32.805}}\\[-0.45em]\scriptsize{\color{mypurple}(-0.098)}}
    & \shortstack[c]{\strut 0.0952\\[-0.45em]\scriptsize{\color{mypurple}(+0.0006)}}
    & \shortstack[c]{\strut 0.0159\\[-0.45em]\scriptsize{\color{mypurple}(+0.0002)}}
    & \shortstack[c]{\strut \smash[b]{\underline{0.91390}}\\[-0.45em]\scriptsize{\color{mypurple}(-0.00091)}} \\[-.4em]
    \bottomrule
  \end{tabular}
  }
  % \vspace{.35em}
\end{table*}

\begin{table}[h]
  \centering
  % \vspace{.5em}
  \caption{Ablation study on automatic matting.}
  \label{tab:ablation_auto_matting}
  \small
  \renewcommand{\arraystretch}{0.9}
  \setlength{\tabcolsep}{4pt}
  \resizebox{.45\textwidth}{!}{
  \begin{tabular}{>{\raggedright\arraybackslash}m{2.2cm}ccccc}
    \toprule
    \multirow{2}{*}{\textbf{Method}} &
    \multicolumn{5}{c}{\textbf{AIM-500}} \\
    \cmidrule(lr){2-6}
    & SAD $\downarrow$ & MSE $\downarrow$ & MAD $\downarrow$ & Grad $\downarrow$ & Conn $\downarrow$ \\
    \midrule
    OmniAlpha RL
    & \textbf{9.089}
    & \textbf{0.0105}
    & \textbf{0.0182}
    & \textbf{10.099}
    & \textbf{8.930} \\
    [-0.17em]
    \midrule

    \raisebox{1.5ex}{w/o $r_{\text{layer}}$}
    & \shortstack[c]{\strut \smash[b]{\underline{9.114}}\\[-0.45em]\scriptsize{\color{mypurple}(+0.0250)}}
    & \shortstack[c]{\strut \smash[b]{\underline{0.01063}}\\[-0.45em]\scriptsize{\color{mypurple}(+0.00013)}}
    & \shortstack[c]{\strut \smash[b]{\underline{0.0183}}\\[-0.45em]\scriptsize{\color{mypurple}(+0.00005)}}
    & \shortstack[c]{\strut 10.396\\[-0.45em]\scriptsize{\color{mypurple}(+0.2977)}}
    & \shortstack[c]{\strut \smash[b]{\underline{8.957}}\\[-0.45em]\scriptsize{\color{mypurple}(+0.0267)}} \\[-.4em]

    \raisebox{1.5ex}{w/o $r_{\text{fg}}$}
    & \shortstack[c]{\strut 9.187\\[-0.45em]\scriptsize{\color{mypurple}(+0.0984)}}
    & \shortstack[c]{\strut 0.01064\\[-0.45em]\scriptsize{\color{mypurple}(+0.00014)}}
    & \shortstack[c]{\strut 0.0184\\[-0.45em]\scriptsize{\color{mypurple}(+0.00020)}}
    & \shortstack[c]{\strut \smash[b]{\underline{10.323}}\\[-0.45em]\scriptsize{\color{mypurple}(+0.2239)}}
    & \shortstack[c]{\strut 9.020\\[-0.45em]\scriptsize{\color{mypurple}(+0.0901)}} \\[-.4em]

    \raisebox{1.5ex}{w/o $r_{\text{bg}}$}
    & \shortstack[c]{\strut 9.152\\[-0.45em]\scriptsize{\color{mypurple}(+0.0634)}}
    & \shortstack[c]{\strut 0.01067\\[-0.45em]\scriptsize{\color{mypurple}(+0.00017)}}
    & \shortstack[c]{\strut 0.0183\\[-0.45em]\scriptsize{\color{mypurple}(+0.00013)}}
    & \shortstack[c]{\strut 10.412\\[-0.45em]\scriptsize{\color{mypurple}(+0.3127)}}
    & \shortstack[c]{\strut 8.998\\[-0.45em]\scriptsize{\color{mypurple}(+0.0680)}} \\[-.4em]

    \raisebox{1.5ex}{w/o $r_{\text{comp}}$}
    & \shortstack[c]{\strut 9.180\\[-0.45em]\scriptsize{\color{mypurple}(+0.0913)}}
    & \shortstack[c]{\strut 0.01071\\[-0.45em]\scriptsize{\color{mypurple}(+0.00020)}}
    & \shortstack[c]{\strut 0.0184\\[-0.45em]\scriptsize{\color{mypurple}(+0.00018)}}
    & \shortstack[c]{\strut 10.399\\[-0.45em]\scriptsize{\color{mypurple}(+0.3004)}}
    & \shortstack[c]{\strut 9.028\\[-0.45em]\scriptsize{\color{mypurple}(+0.0976)}} \\[-.4em]
    \bottomrule
  \end{tabular}
  }
\end{table}

\medskip\noindent\textbf{Automatic Matting.}
Table~\ref{tab:ablation_auto_matting} shows that the full \method{} RL model achieves the best performance on AIM-500 across all metrics, confirming the effectiveness of the full reward design. Removing any reward term degrades performance. In particular, removing $\mathrm{r}_{\text{fg}}$, $\mathrm{r}_{\text{bg}}$, or $\mathrm{r}_{\text{comp}}$ causes more noticeable drops on SAD, Grad, and Conn, while removing $\mathrm{r}_{\text{layer}}$ leads to the smallest degradation but remains consistently worse than the full model. Figure~\ref{fig:ablation_case} provides qualitative examples that are consistent with these quantitative results. These results indicate that all reward terms contribute to accurate alpha prediction.

\medskip\noindent\textbf{Layer Decomposition.}
Table~\ref{tab:ablation_decompose} shows that the full \method{} RL model performs best on OBER-Test-Decompose for both RGB L1 and Alpha soft IoU. Removing any reward term degrades performance, indicating that all components are beneficial. In particular, $\mathrm{r}_{\text{layer}}$ is the most critical, while $\mathrm{r}_{\text{fg}}$ and $\mathrm{r}_{\text{comp}}$ also have clear impact. Figure~\ref{fig:ablation_case} provides qualitative examples consistent with these trends.

\begin{table}[h]
  \centering
  \vspace{.5em}
  \caption{Ablation study on layer decomposition.}
  \label{tab:ablation_decompose}
  \small
  \renewcommand{\arraystretch}{0.9}
  \setlength{\tabcolsep}{4pt}
  \begin{tabular}{>{\raggedright\arraybackslash}m{2.2cm} >{\centering\arraybackslash}p{2.1cm} >{\centering\arraybackslash}p{2.6cm}}
    \toprule
    \multirow{2}{*}{\textbf{Method}} &
    \multicolumn{2}{c}{\textbf{OBER-Test-Decompose}} \\
    \cmidrule(lr){2-3}
    & RGB L1 $\downarrow$ & Alpha soft IoU $\uparrow$ \\
    \midrule
    OmniAlpha RL 
    & \textbf{0.1032}
    & \textbf{0.7069} \\
    [-0.17em]
    \midrule

    \raisebox{1.5ex}{w/o $r_{\text{layer}}$}
    & \shortstack[c]{\strut 0.1096\\[-0.45em]\scriptsize{\color{mypurple}(+0.0065)}}
    & \shortstack[c]{\strut 0.6918\\[-0.45em]\scriptsize{\color{mypurple}(-0.0151)}} \\[-.4em]

    \raisebox{1.5ex}{w/o $r_{\text{fg}}$}
    & \shortstack[c]{\strut 0.1069\\[-0.45em]\scriptsize{\color{mypurple}(+0.0037)}}
    & \shortstack[c]{\strut 0.6927\\[-0.45em]\scriptsize{\color{mypurple}(-0.0141)}} \\[-.4em]

    \raisebox{1.5ex}{w/o $r_{\text{bg}}$}
    & \shortstack[c]{\strut \smash[b]{\underline{0.1065}}\\[-0.45em]\scriptsize{\color{mypurple}(+0.0034)}}
    & \shortstack[c]{\strut \smash[b]{\underline{0.6965}}\\[-0.45em]\scriptsize{\color{mypurple}(-0.0103)}} \\[-.4em]

    \raisebox{1.5ex}{w/o $r_{\text{comp}}$}
    & \shortstack[c]{\strut 0.1078\\[-0.45em]\scriptsize{\color{mypurple}(+0.0046)}}
    & \shortstack[c]{\strut 0.6943\\[-0.45em]\scriptsize{\color{mypurple}(-0.0126)}} \\[-.4em]
    \bottomrule
  \end{tabular}
\end{table}

\medskip
Ablation results reveal that the four reward terms are complementary and jointly necessary for robust multi-task RGBA alignment. Although their importance varies across tasks, removing any single term consistently degrades performance on at least one benchmark or metric, while the full reward design yields the strongest overall objective. In particular, $\mathrm{r}_{\text{layer}}$ is essential for structural consistency, $\mathrm{r}_{\text{fg}}$ is especially beneficial for foreground-sensitive tasks such as text-guided matting, $\mathrm{r}_{\text{bg}}$ is crucial for background recovery in object removal, and $\mathrm{r}_{\text{comp}}$ further improves cross-layer compositional fidelity. 
\section{Conclusion}
\label{sec:conclusion}

In this paper, we introduced \method{}, a unified multi-task reinforcement learning framework for transparency-aware generation and manipulation. By combining an alpha-aware VAE, a sequence-to-sequence Diffusion Transformer, and GRPO-style post-training with layer-aware rewards, \method{} enables a single model to handle diverse RGBA workflows in a unified manner. Experiments across five categories of transparency-aware tasks show that unified RL alignment consistently improves over multi-task supervised fine-tuning and achieves strong performance against specialized expert systems.
These results highlight the promise of moving from task-specific RGBA pipelines to general-purpose transparency-aware foundation models. We hope this work motivates future research on reinforcement learning for layer-aware generation, compositional reasoning, and more flexible multi-layer visual creation.

{
    \small
    \bibliographystyle{ieeenat_fullname}
    \bibliography{main}
}

\appendix
\clearpage
\setcounter{page}{1}

\section{Details of Model Architecture}
\label{app:arch}

\subsection{Opaque Initialization of VAE}
\label{app:arch:vae}

Formally, let $(W_0^\vaeeref, b_0^\vaeeref)$ be the parameters of the first convolutional layer of $\vaeeref$, with $W_0^\vaeeref \in \mathbb{R}^{k\times k \times 3 \times D_c}$ and $b_0^\vaeeref \in \mathbb{R}^{D_c}$. Let $(W_L^\vaedref, b_L^\vaedref)$ be the parameters of the final convolutional layer of $\vaedref$, with $W_L^\vaedref \in \mathbb{R}^{k'\times k' \times D_f \times 3}$ and $b_L^\vaedref \in \mathbb{R}^{3}$.
The corresponding layers of the new 4-channel VAE, $(W_0^\vaee, b_0^\vaee)$ and $(W_L^\vaed, b_L^\vaed)$, are initialized as specified in Equation \ref{eq:vae_init}:
\begin{equation}
\label{eq:vae_init}
\begin{cases}
    W_0^\vaee[:,:,1:3,:] = W_0^\vaeeref & \text{(Copy RGB weights)} \\[.2em]
    W_0^\vaee[:,:,4,:] = \mathbf{0} & \text{(0-init alpha weights)} \\[.2em]
    b_0^\vaee = b_0^\vaeeref & \text{(Copy bias)} \\[.2em]
    W_L^\vaed[:,:,:,1:3] = W_L^\vaedref & \text{(Copy RGB weights)} \\[.2em]
    W_L^\vaed[:,:,:,4] = \mathbf{0} & \text{(0-init alpha weights)} \\[.2em]
    b_L^\vaed[1:3] = b_L^\vaedref & \text{(Copy RGB biases)} \\[.2em]
    b_L^\vaed[4] = \textbf{1} & \text{(Set as opaque)}
\end{cases}
\end{equation}
% \begin{equation}
% \label{eq:vaed_init}
% \begin{cases}
% \end{cases}
% \end{equation}
where $W_0^\vaee$ and $W_L^\vaed$ have new shapes $\mathbb{R}^{k\times k \times 4 \times D_c}$ and $\mathbb{R}^{k'\times k' \times D_f \times 4}$, respectively. The bias $b_L^\vaed$ is a 4-dimensional vector.
This opaque initialization provides a stable starting point for fine-tuning.

\subsection{RoPE with Bidirectional Z-axis}
\label{app:arch:rope}

Inspired by~\cite{wu2025qwenimagetechnicalreport}, to unify multimodal inputs and multiple outputs into a single sequence, we equip a bi-directionally extendable layer axis to the standard 2D Rotary Position Embedding (RoPE)~\citep{su2024roformer}. This z-axis differentiates components by assigning unique indices. The $n$ input latents occupy non-negative indices from $z=0$ to $n-1$. The $m$ target latents are assigned negative indices from $z=-1$ to $-m$. The contextual embeddings output by the language encoder are assigned distinct positive indices $z \ge n$, spatially separating them from the input image latents. This coordinate formulation effectively transforms the multi-image generation task into a unified sequence-to-sequence denoising paradigm.

The implementation leverages the inherent translation invariance property of Rotary Positional Embeddings. Fundamentally, the dot product of two RoPE-encoded feature vectors, $\mathbf{q}$ and $\mathbf{k}$, at positions $\mathbf{p}_i = (x_i, y_i, z_i)$ and $\mathbf{p}_j = (x_j, y_j, z_j)$, depends solely on their relative spatial and layer distances:
\begin{equation}
    \langle \text{R}(\mathbf{q}, \mathbf{p}_i), \text{R}(\mathbf{k}, \mathbf{p}_j) \rangle = g(\mathbf{q}, \mathbf{k}, \mathbf{p}_i - \mathbf{p}_j).
\end{equation}
Consequently, shifting the layer index $z$ by a constant scalar $S$ for all tokens preserves the relative positional relationships and, by extension, the attention scores. This property can be formally expressed as:
\begin{equation}
\begin{split}
    &\text{R}(\mathbf{h}, x, y, z) \cdot \text{R}(\mathbf{h}', x', y', z')^{-1} \\
    &=\text{R}(\mathbf{h}, x, y, z + S) \cdot \text{R}(\mathbf{h}', x', y', z' + S)^{-1},
\end{split}
\end{equation}
where $\mathbf{h}$ represents the hidden states.

In our sequence-to-sequence formulation, target images are assigned negative layer indices $z \in \{-m, \dots, -1\}$, where $m$ is the number of generated target frames. However, the pre-trained Qwen-Image-Edit backbone typically utilizes pre-computed frequency tables defined over a non-negative domain. To align our bi-directional indexing strategy with the pre-defined frequency configurations of the base model, we implement a global index shift operation.

Let $z_{\text{raw}}$ denote the logical layer index defined in Section~\ref{sec:method}. Specifically:
\begin{equation}
    z_{\text{raw}} \in 
    \begin{cases} 
        \{-m, \dots, -1\} & \text{for target latents}, \\
        \{0, \dots, n-1\} & \text{for input latents}, \\
        \{n, \dots\} & \text{for VLM condition tokens}.
    \end{cases}
\end{equation}
To ensure all indices map to the valid domain of the pre-trained frequency encodings, we define the implementation index $z_{\text{impl}}$ as:
\begin{equation}
    z_{\text{impl}} = z_{\text{raw}} + S_{\text{offset}}, \quad \text{where } S_{\text{offset}} \ge m.
\end{equation}
In practice, we set $S_{\text{offset}} = m$. This transformation allows \method{} to concurrently process multiple inputs and outputs while seamlessly utilizing the optimized RoPE functions and frequency computations inherited from Qwen-Image-Edit.

\section{More Results}
\label{app:results}

\begin{figure*}[htbp]
    \centering
    \includegraphics[width=1\linewidth]{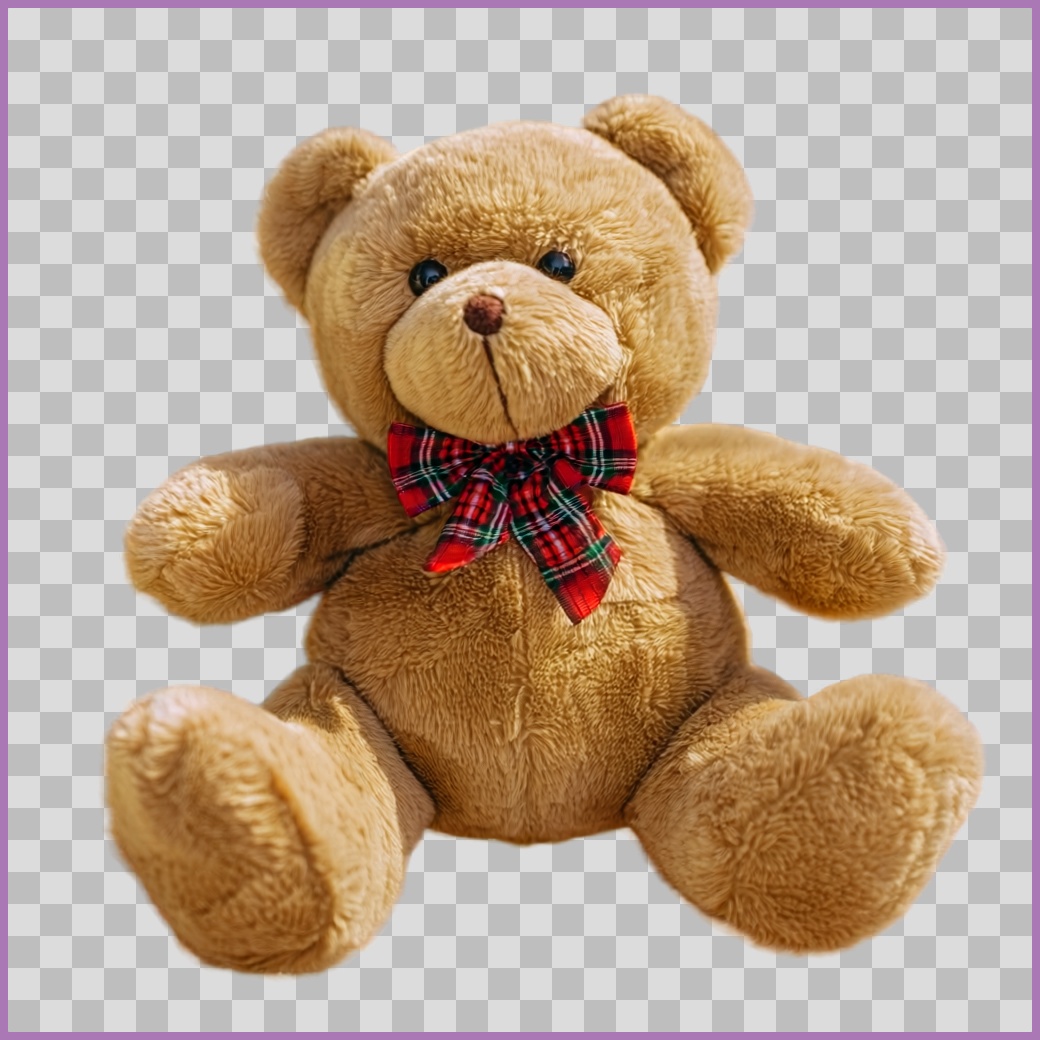}
    \caption{A light brown plush teddy bear with a red plaid bow tie sits upright with arms slightly spread and legs apart.}
    \label{fig:app:t2i:start}
\end{figure*}

\begin{figure*}[htbp]
    \centering
    \includegraphics[width=1\linewidth]{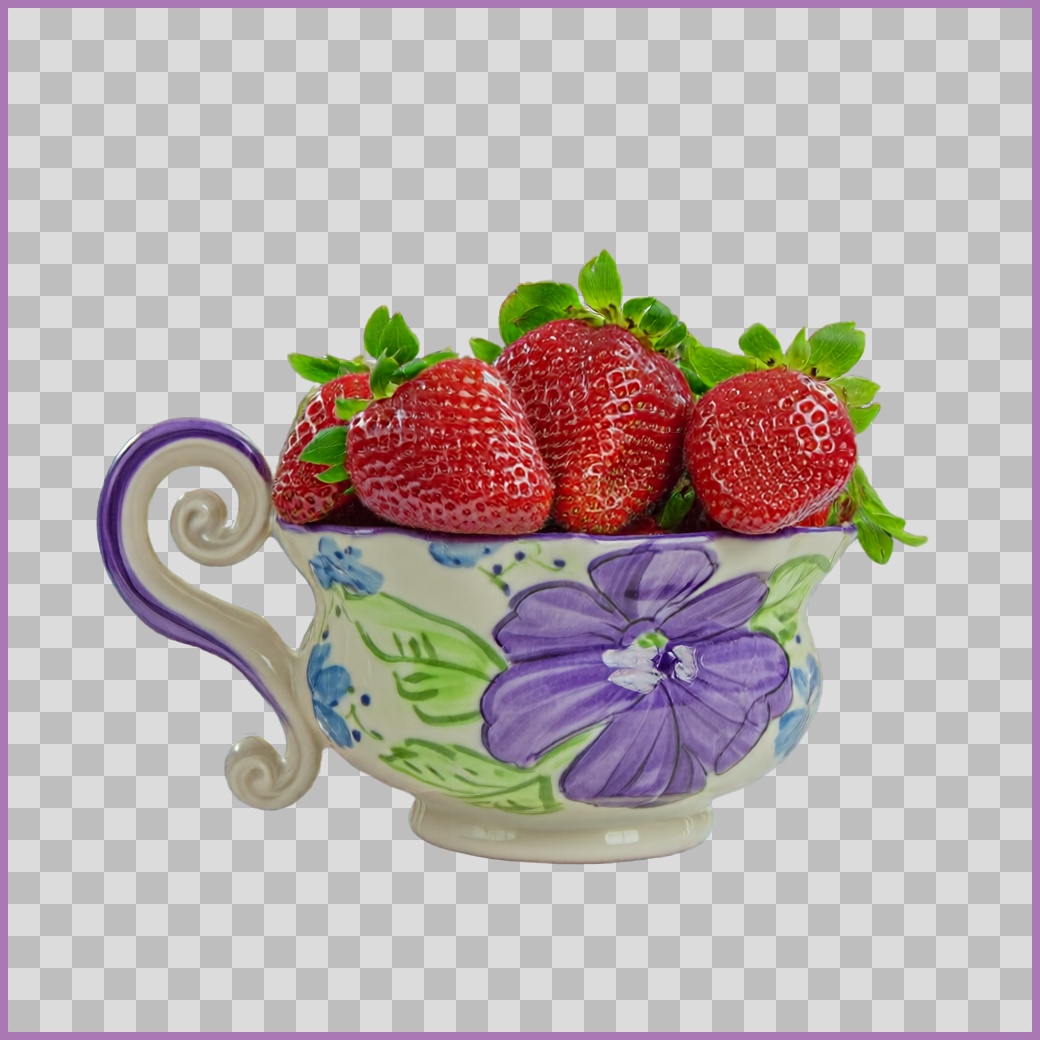}
    \caption{A decorative ceramic teacup with a swirling handle, adorned with floral patterns in purple, green, and blue, overflowing with fresh red strawberries topped with green leaves.}
    \label{fig:app:t2i:o_f7c5c922}
\end{figure*}

\begin{figure*}[htbp]
    \centering
    \includegraphics[width=1\linewidth]{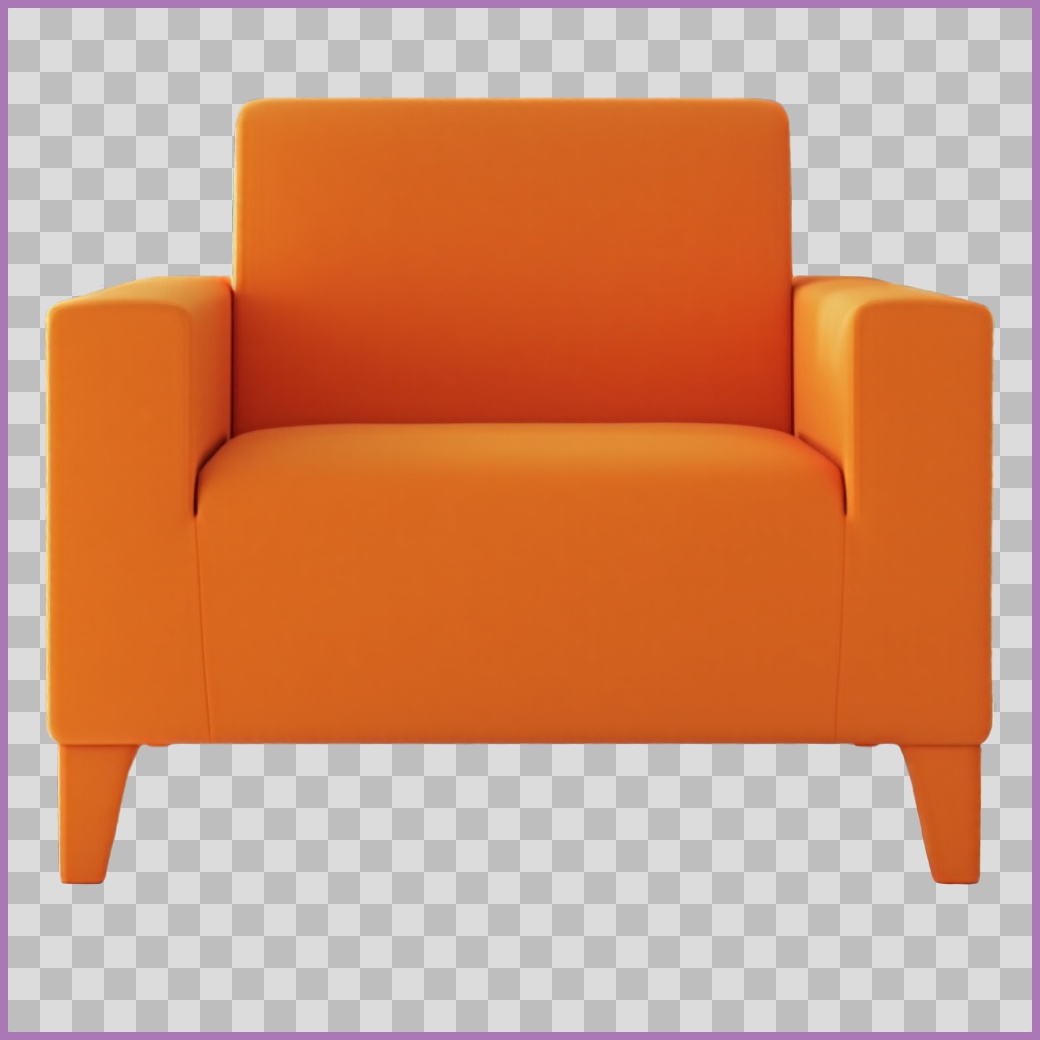}
    \caption{A bright orange modern armchair with a solid blocky design and four angled legs.}
    \label{fig:app:t2i:o_fe6f4047}
\end{figure*}

\begin{figure*}[htbp]
    \centering
    \includegraphics[width=1\linewidth]{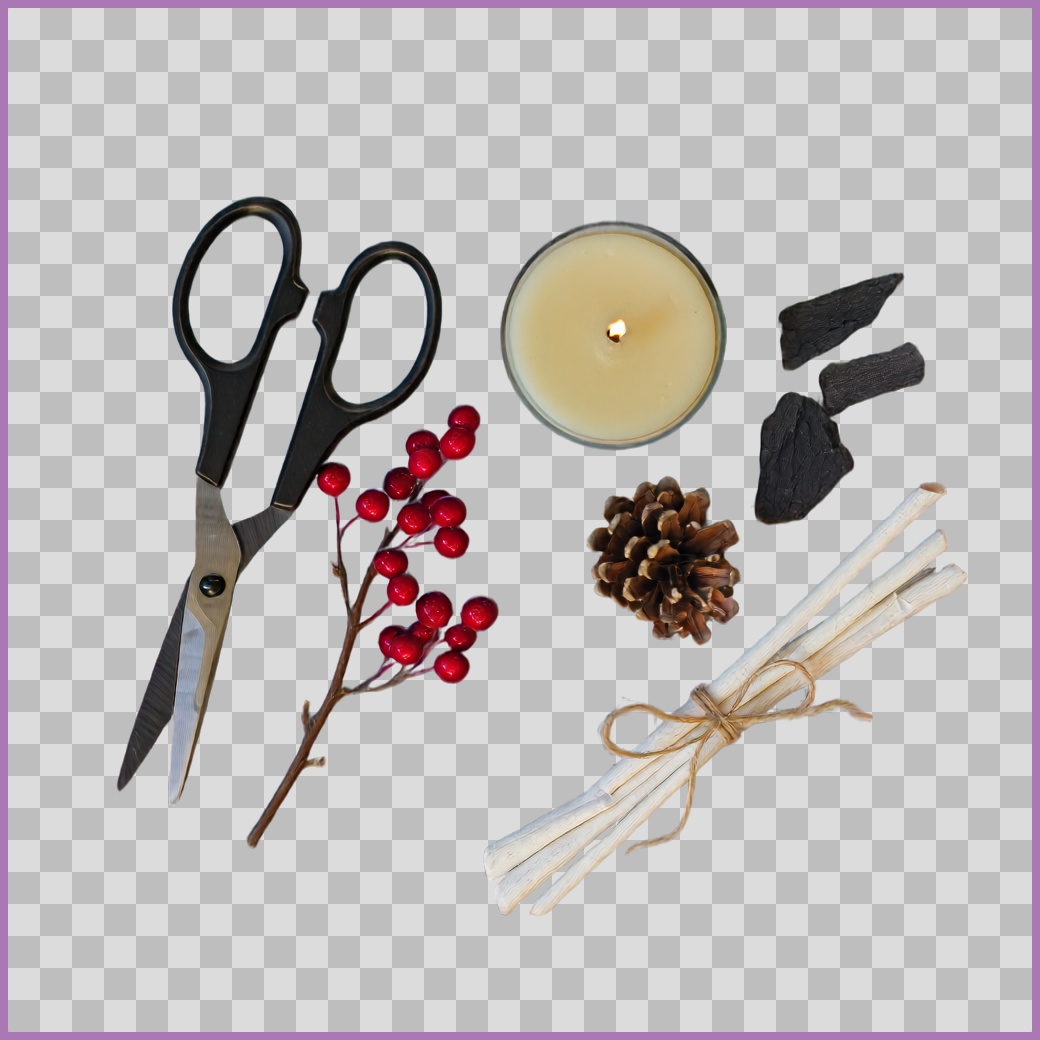}
    \caption{A decorative arrangement featuring black scissors, a cream candle, red berries on a twig, a pinecone, charcoal pieces, and white sticks bound together with twine.}
    \label{fig:app:t2i:o_0b376f13}
\end{figure*}

\begin{figure*}[htbp]
    \centering
    \includegraphics[width=1\linewidth]{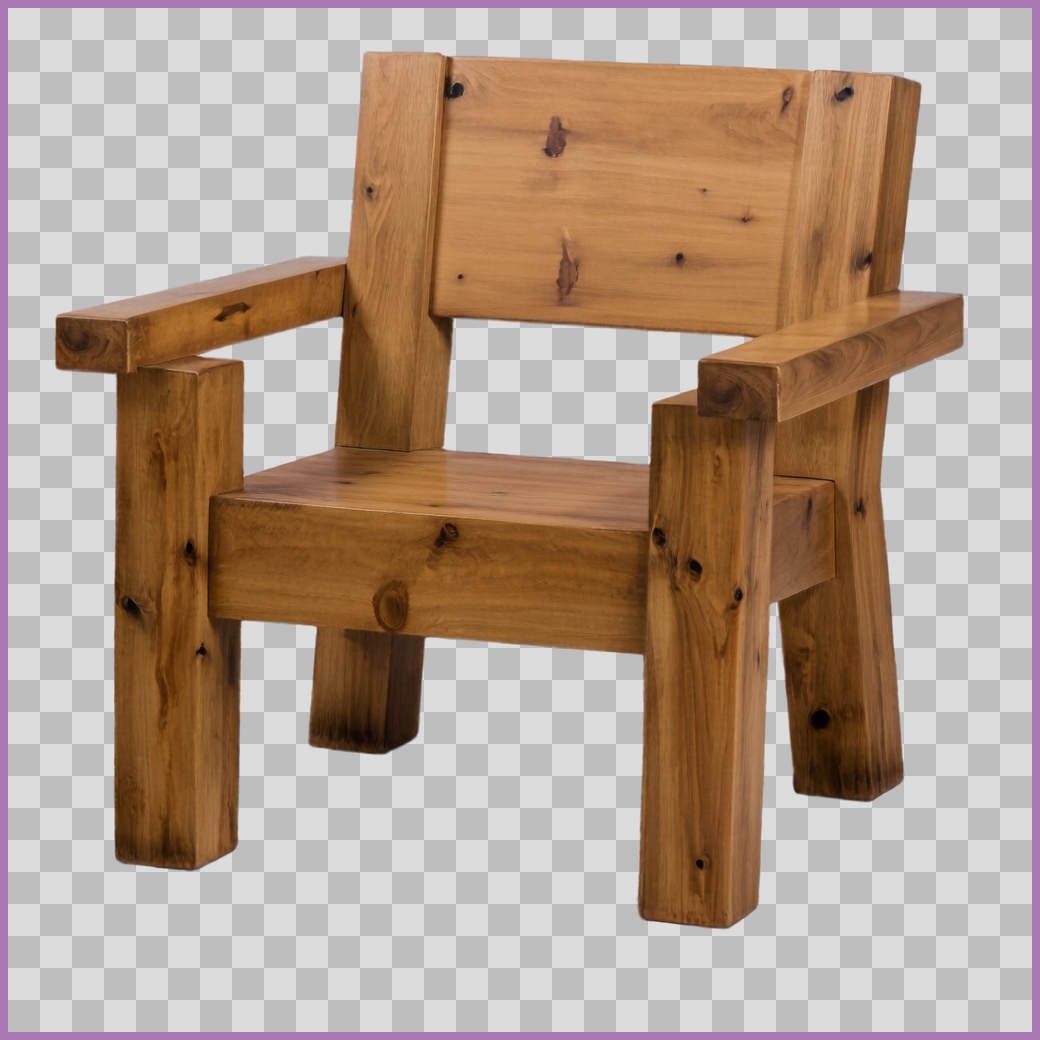}
    \caption{A rustic wooden armchair with a simple rectangular backrest, thick armrests, and sturdy legs showing natural wood grain and knots.}
    \label{fig:app:t2i:o_053490ad}
\end{figure*}

\begin{figure*}[htbp]
    \centering
    \includegraphics[width=1\linewidth]{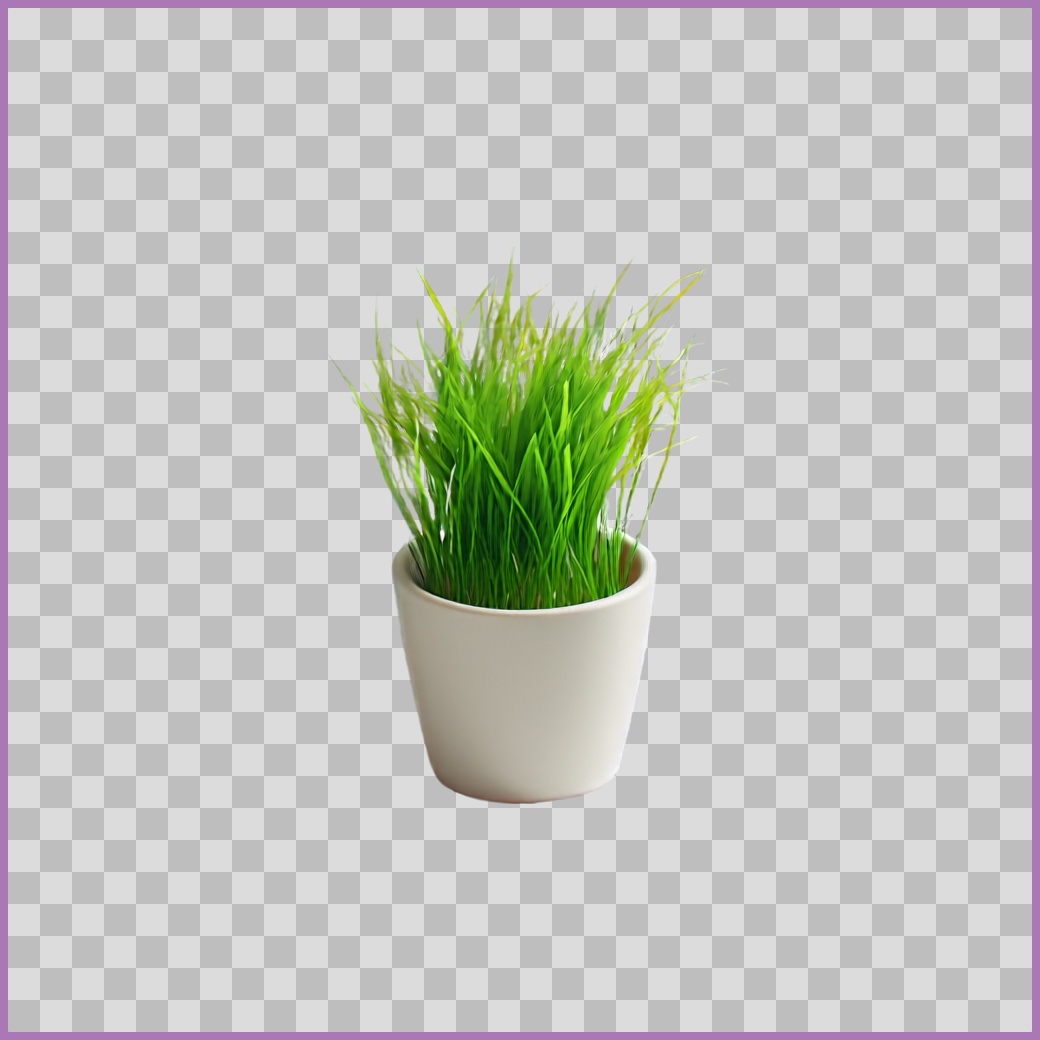}
    \caption{A small potted plant with vibrant green grass-like leaves in a simple white ceramic pot.}
    \label{fig:app:t2i:o_e0ab2760}
\end{figure*}

\begin{figure*}[htbp]
    \centering
    \includegraphics[width=1\linewidth]{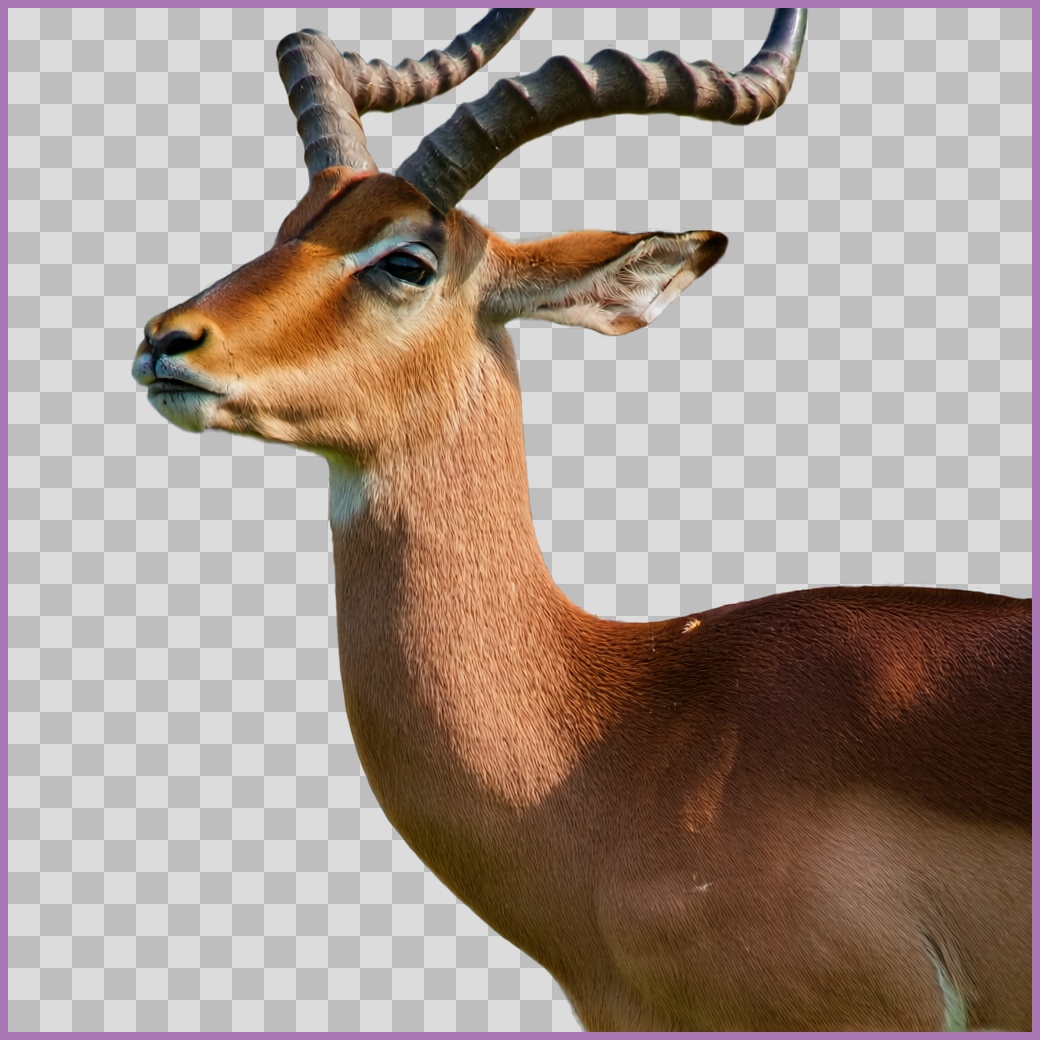}
    \caption{A close-up of an antelope's head and neck, showing its brown fur, large ears, and spiraled horns.}
    \label{fig:app:t2i:o_e78005d6}
\end{figure*}

\begin{figure*}[htbp]
    \centering
    \includegraphics[width=1\linewidth]{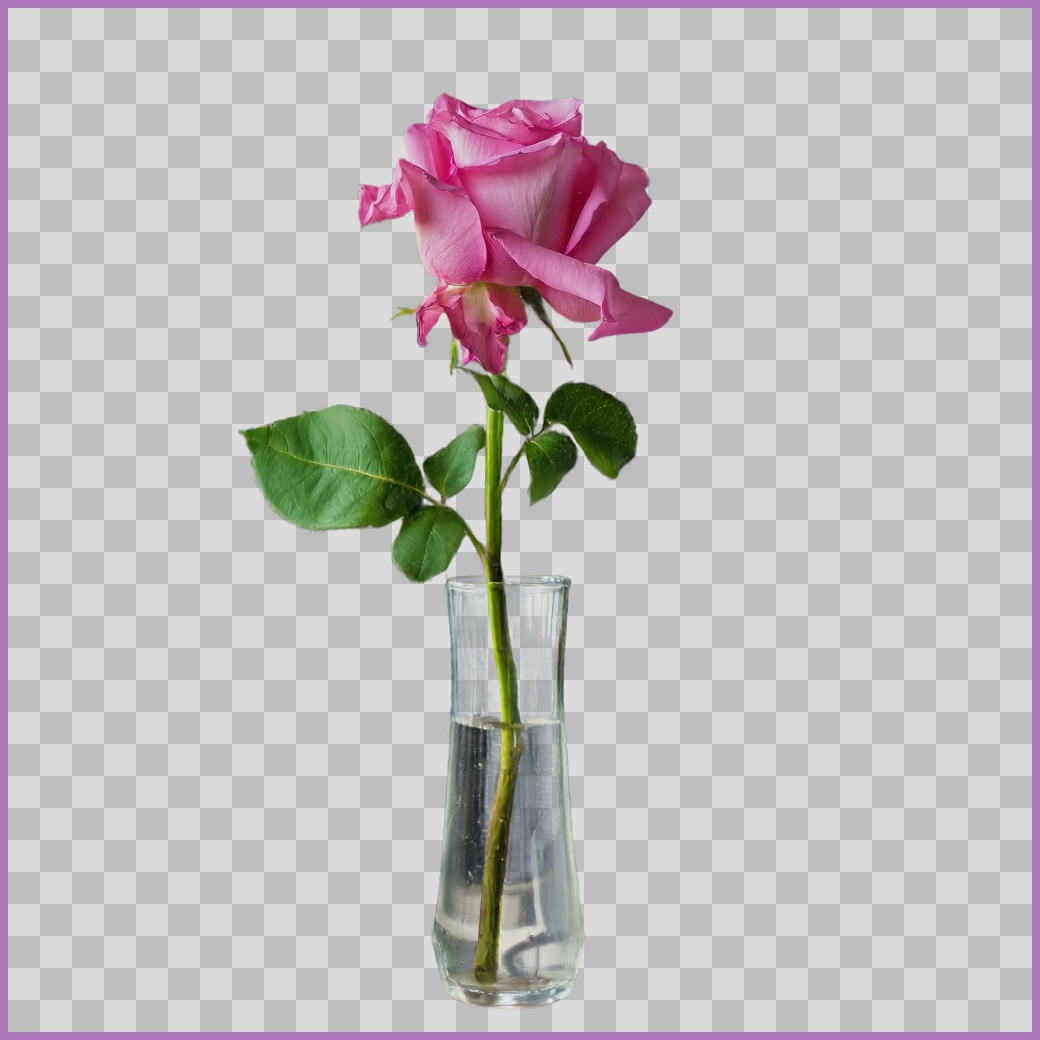}
    \caption{A single pink rose with slightly wilted petals and green leaves, standing upright in a clear glass vase filled with water.}
    \label{fig:app:t2i:o_1f05f13c}
\end{figure*}

\begin{figure*}[htbp]
    \centering
    \includegraphics[width=1\linewidth]{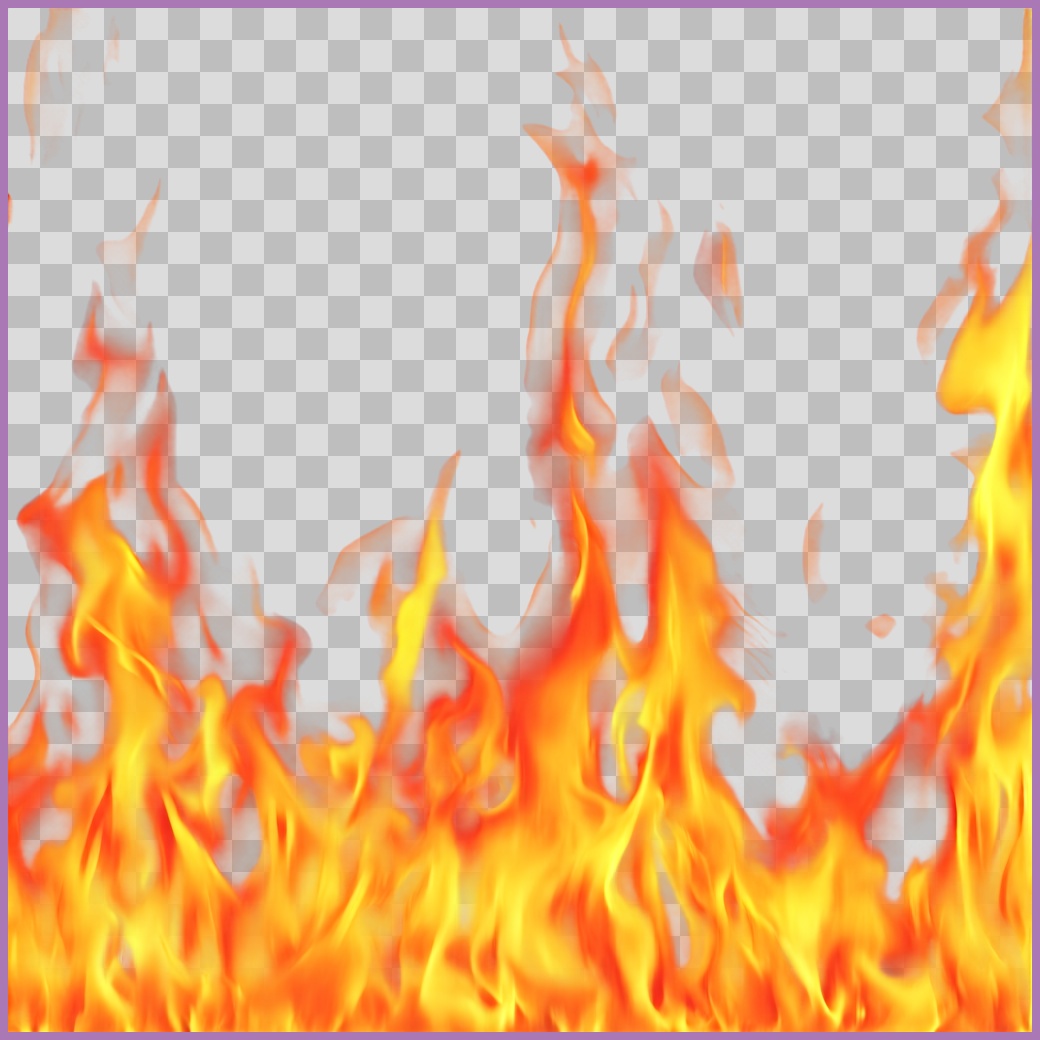}
    \caption{Vivid orange and yellow flames rising upward with dynamic, flowing shapes.}
    \label{fig:app:t2i:o_3c0b2617}
\end{figure*}

\begin{figure*}[htbp]
    \centering
    \includegraphics[width=1\linewidth]{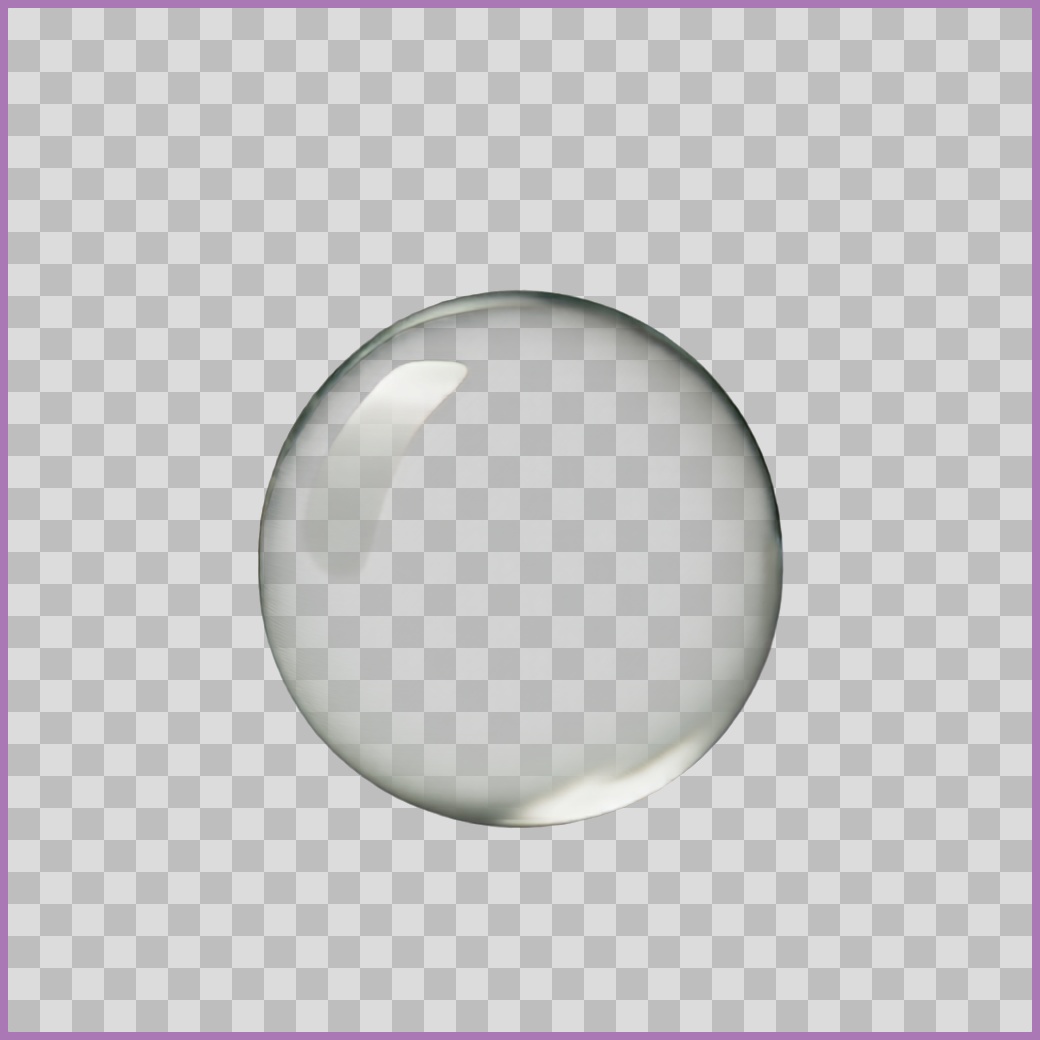}
    \caption{A clear, round water droplet with a smooth, glossy surface.}
    \label{fig:app:t2i:o_b71d875b}
\end{figure*}

\begin{figure*}[htbp]
    \centering
    \includegraphics[width=1\linewidth]{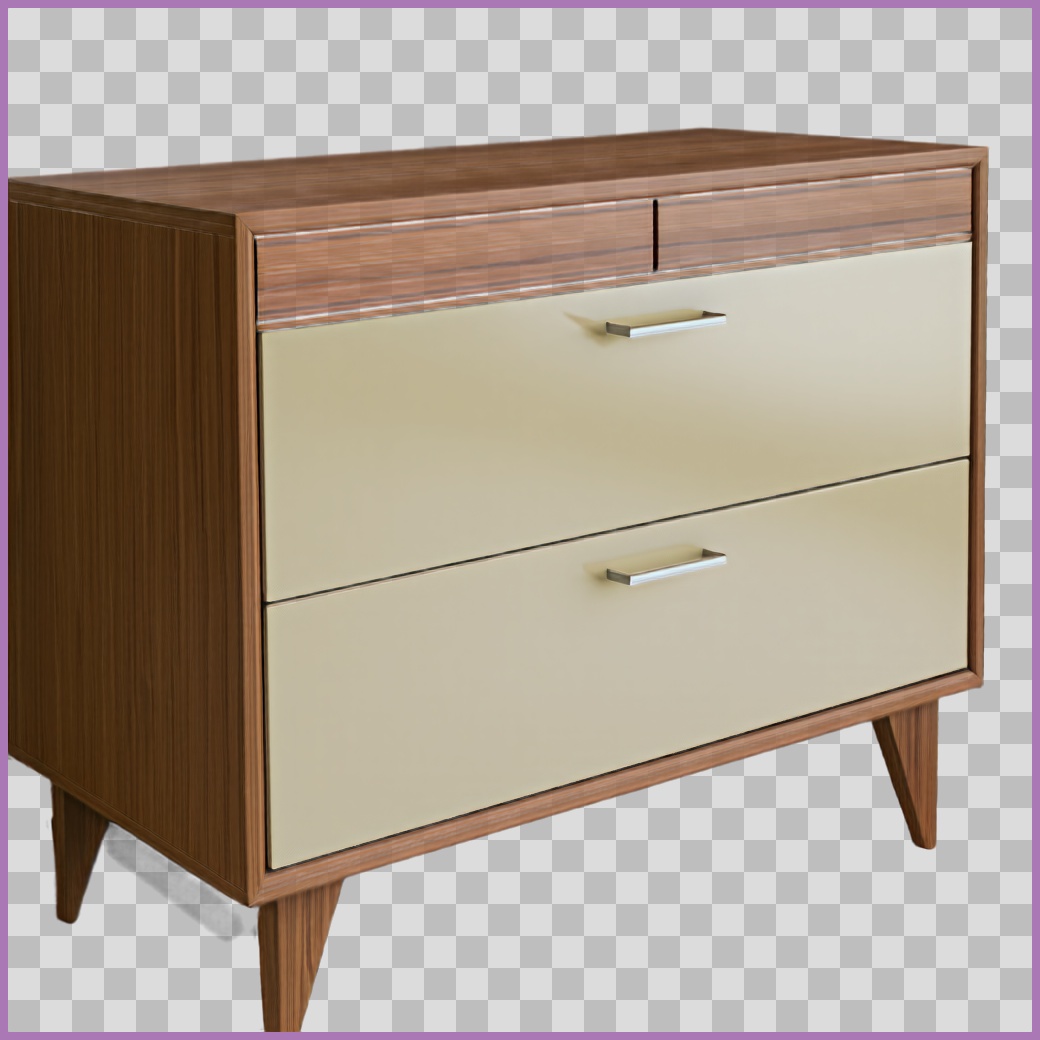}
    \caption{A modern wooden cabinet with light-colored drawers and minimalist handles.}
    \label{fig:app:t2i:o_da41b554}
\end{figure*}

\begin{figure*}[htbp]
    \centering
    \includegraphics[width=1\linewidth]{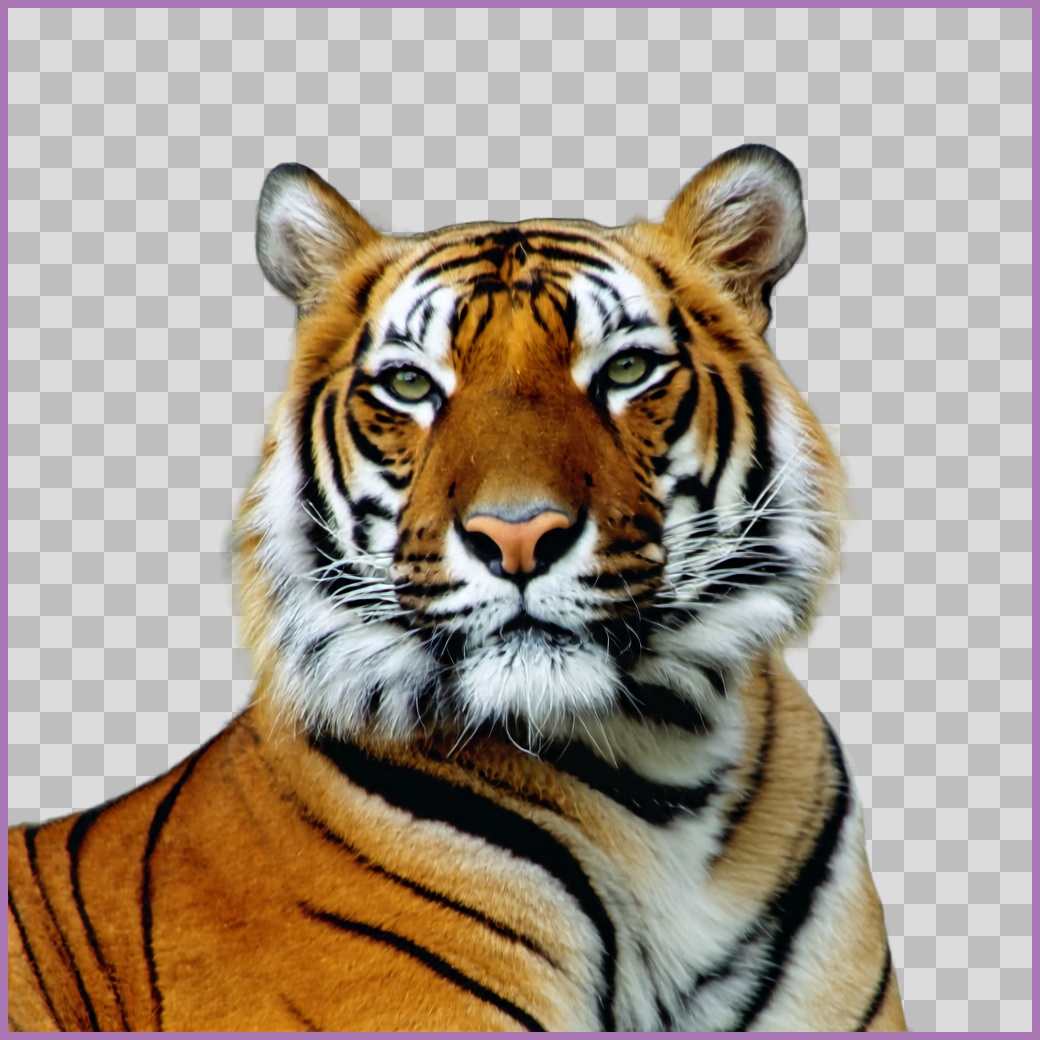}
    \caption{A tiger with orange fur and black stripes stares directly forward, its head and upper torso visible.}
    \label{fig:app:t2i:o_c15a96a6}
\end{figure*}

\begin{figure*}[htbp]
    \centering
    \includegraphics[width=1\linewidth]{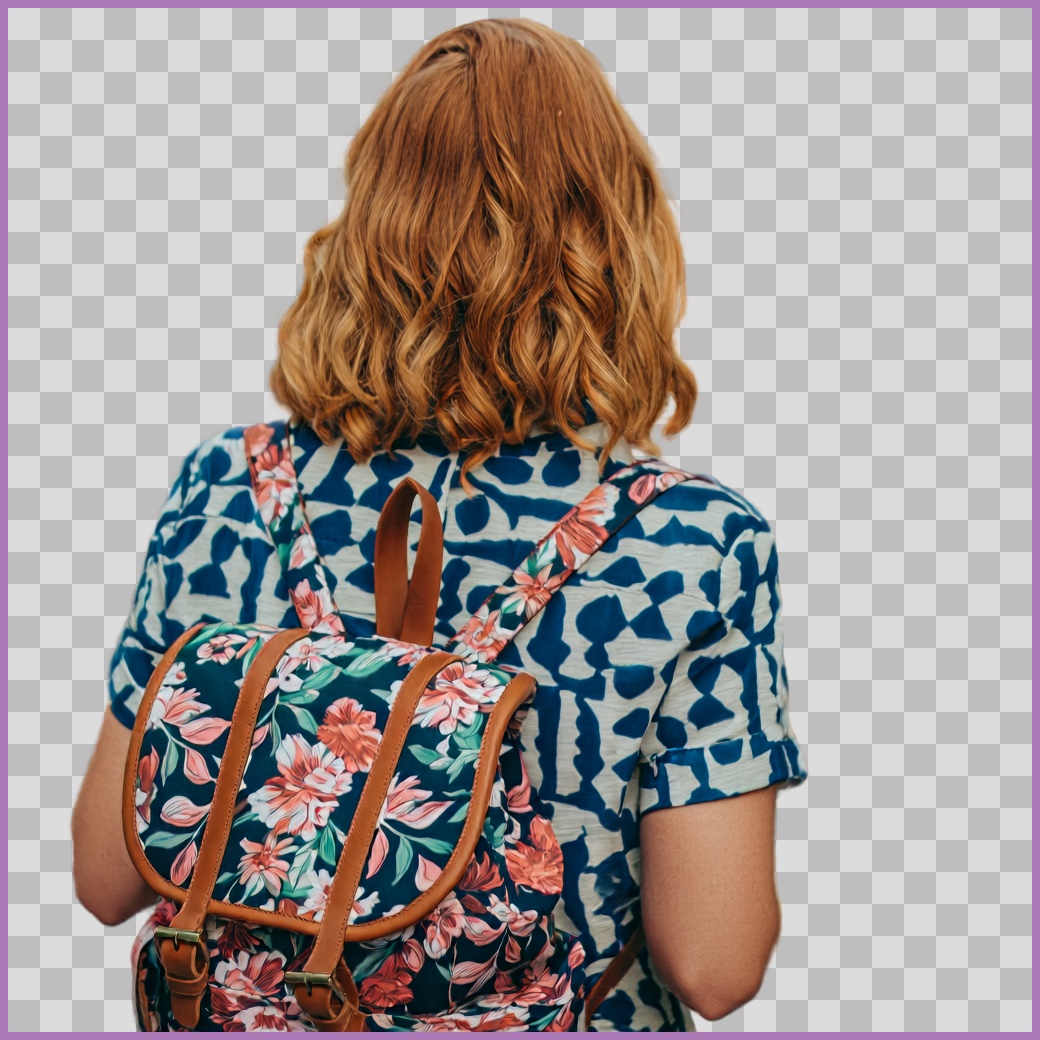}
    \caption{A person with shoulder-length wavy reddish-blonde hair, wearing a blue and white patterned shirt, carries a floral-patterned backpack with brown leather straps.}
    \label{fig:app:t2i:o_c01ea2f5}
\end{figure*}

\begin{figure*}[htbp]
    \centering
    \includegraphics[width=1\linewidth]{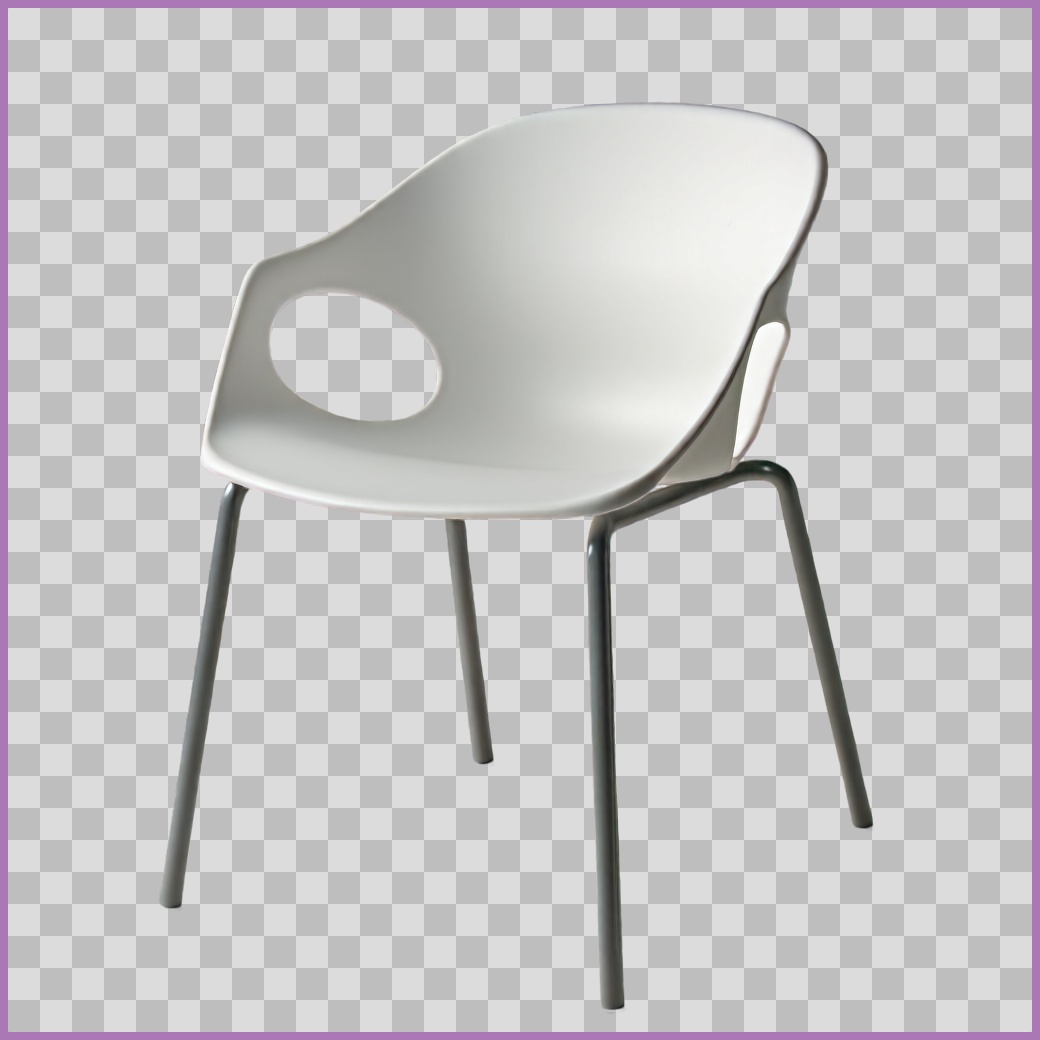}
    \caption{A modern white plastic chair with gray metal legs and cut-out sides.}
    \label{fig:app:t2i:end}
\end{figure*}

\begin{figure*}[htbp]
    \centering
    \includegraphics[width=1\linewidth]{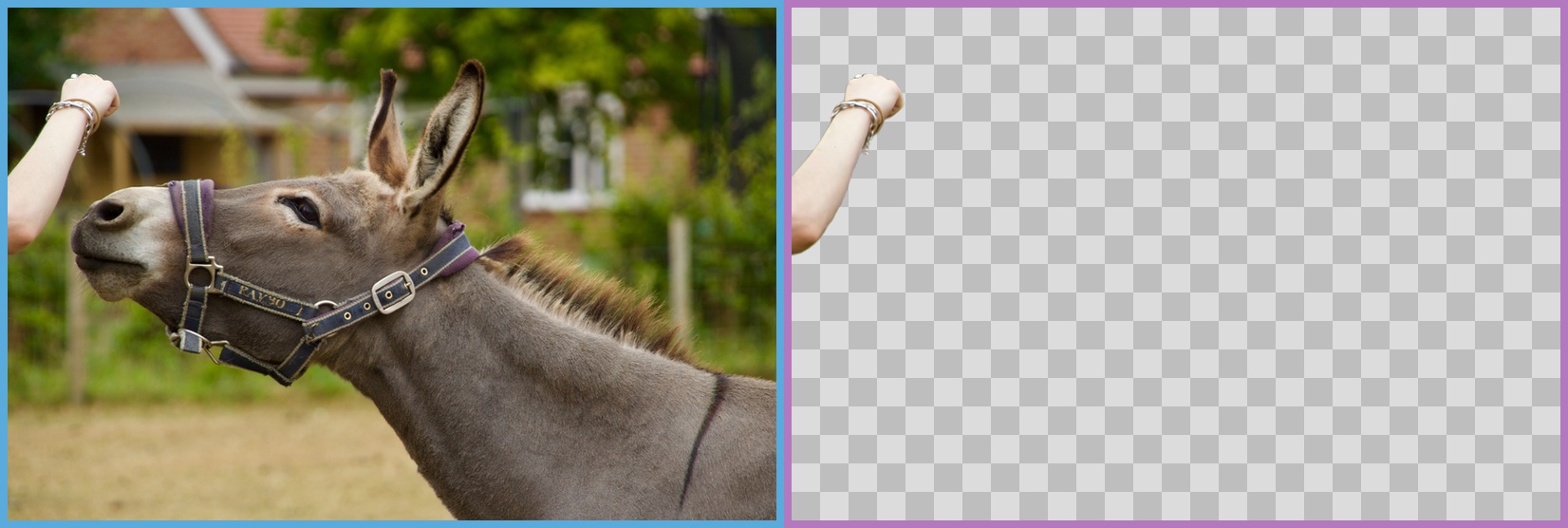}
    \caption{Extract the object described by the text, preserving fine details and transparency: the human arm on the left edge of the image}
    \label{fig:app:refmatte:start}
\end{figure*}

\begin{figure*}[htbp]
    \centering
    \includegraphics[width=1\linewidth]{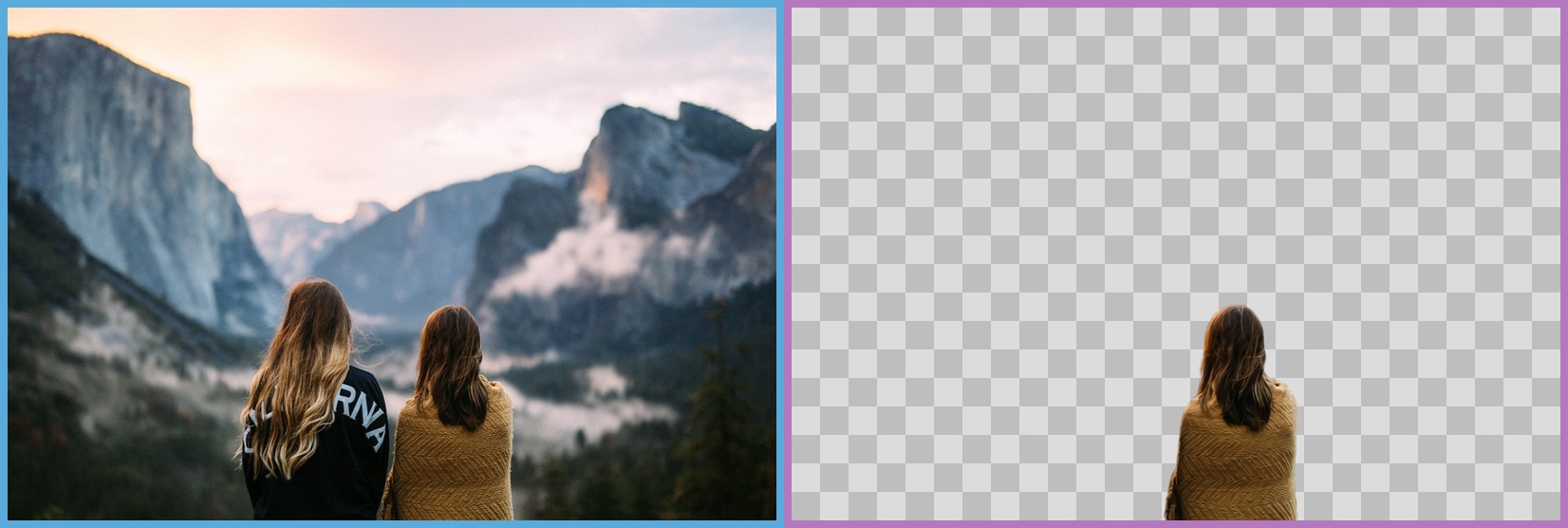}
    \caption{Extract the object described by the text, preserving fine details and transparency: a girl with medium-length brown hair in yellow located on the center right of the photo}
    \label{fig:app:refmatte:rim_rw_ab22e37f_layer_01}
\end{figure*}

\begin{figure*}[htbp]
    \centering
    \includegraphics[width=1\linewidth]{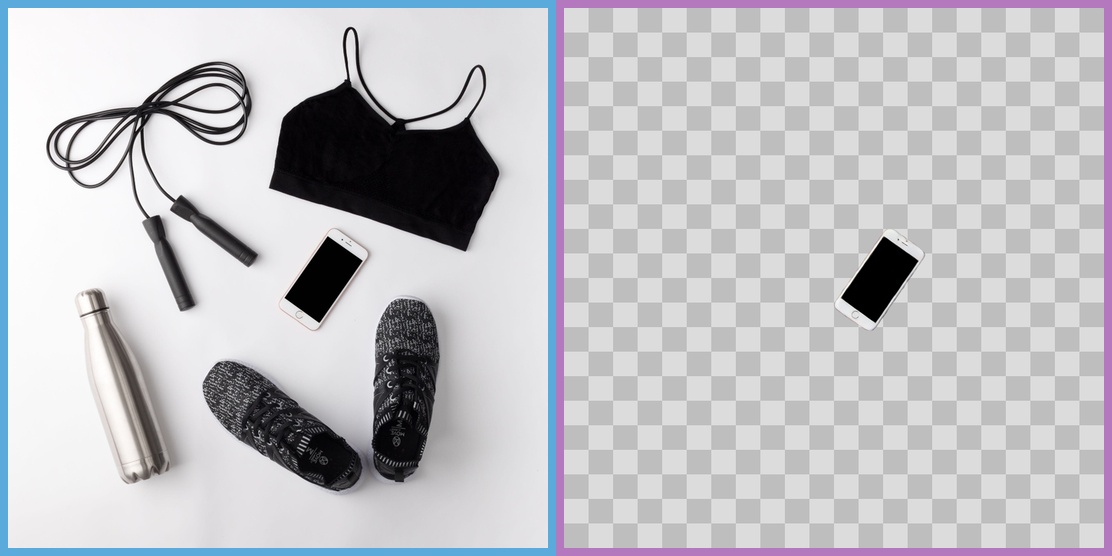}
    \caption{Extract the object described by the text, preserving fine details and transparency: the white phone that is on the middle of the image}
    \label{fig:app:refmatte:rim_rw_5dd2fa88_layer_03}
\end{figure*}

\begin{figure*}[htbp]
    \centering
    \includegraphics[width=1\linewidth]{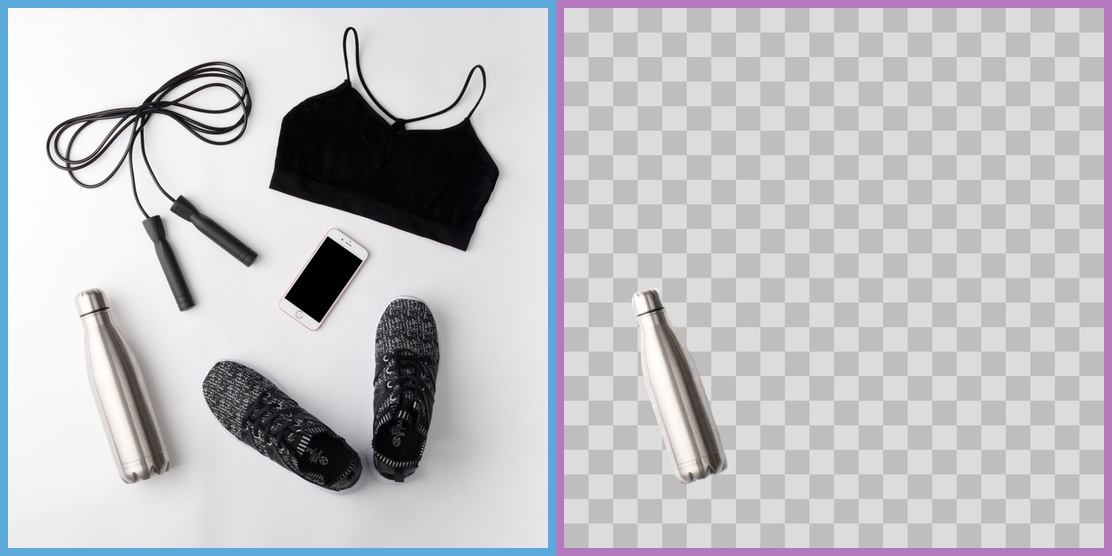}
    \caption{Extract the object described by the text, preserving fine details and transparency: a silver metal water bottle that is on the left bottom of the photo}
    \label{fig:app:refmatte:rim_rw_5dd2fa88_layer_01}
\end{figure*}

\begin{figure*}[htbp]
    \centering
    \includegraphics[width=1\linewidth]{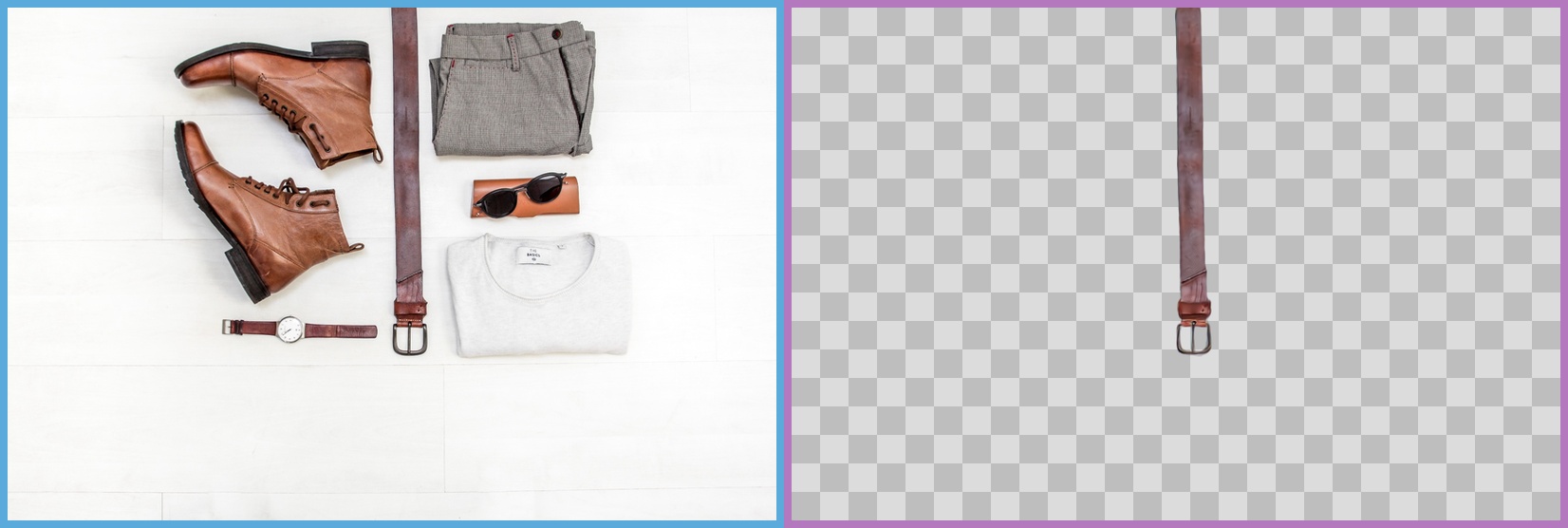}
    \caption{Extract the object described by the text, preserving fine details and transparency: the leather belt on the middle of the image}
    \label{fig:app:refmatte:rim_rw_22e45b23_layer_02}
\end{figure*}

\begin{figure*}[htbp]
    \centering
    \includegraphics[width=1\linewidth]{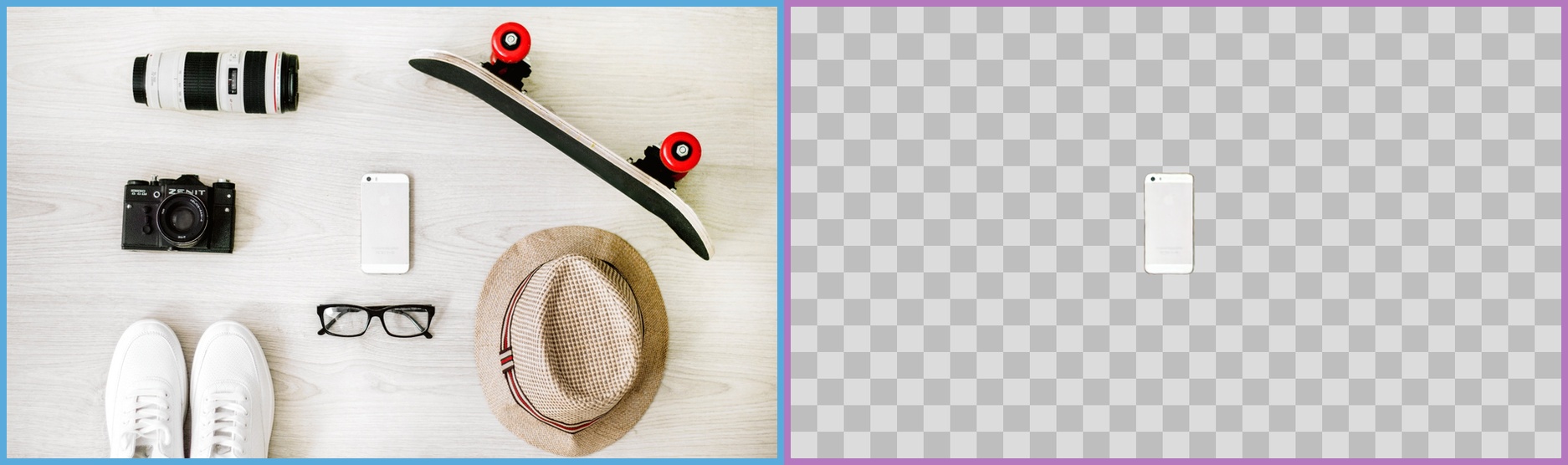}
    \caption{Extract the object described by the text, preserving fine details and transparency: the silver iphone is in the middle of the image}
    \label{fig:app:refmatte:rim_rw_9e27487c_layer_03}
\end{figure*}

\begin{figure*}[htbp]
    \centering
    \includegraphics[width=1\linewidth]{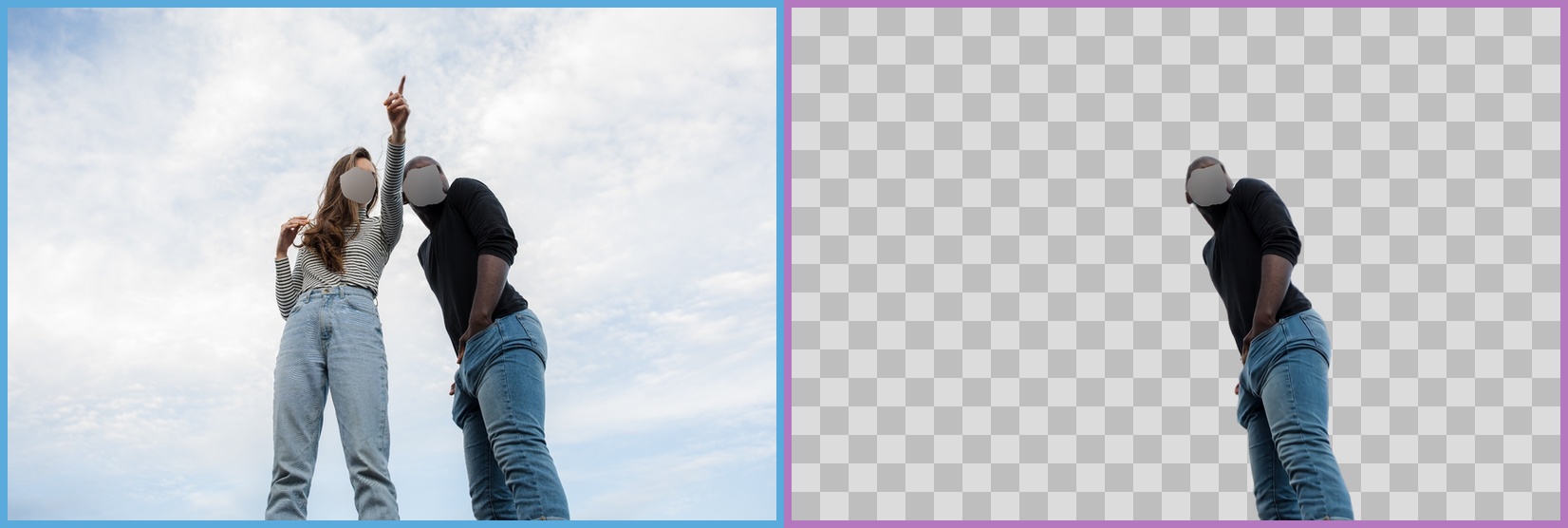}
    \caption{Extract the object described by the text, preserving fine details and transparency: A black man in a black shirt and blue pants is on the right side of the photo}
    \label{fig:app:refmatte:rim_rw_1df85524_layer_01}
\end{figure*}

\begin{figure*}[htbp]
    \centering
    \includegraphics[width=1\linewidth]{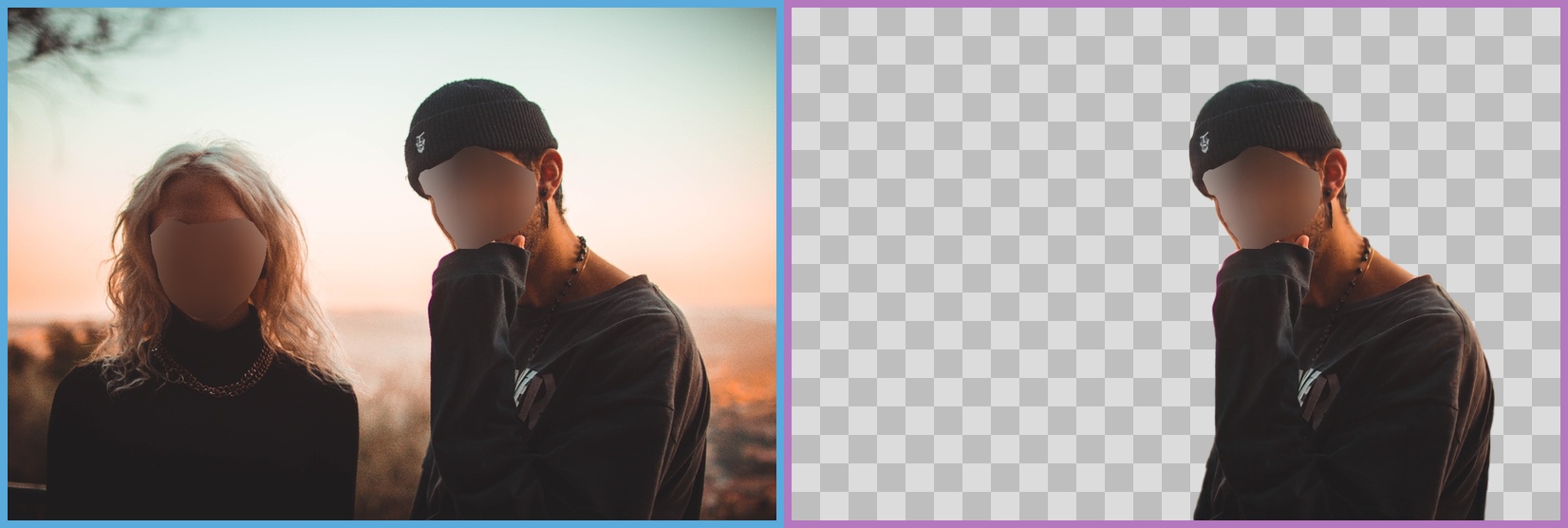}
    \caption{Extract the object described by the text, preserving fine details and transparency: the male human-being on the right side of the image}
    \label{fig:app:refmatte:rim_rw_ceeb03d9_layer_01}
\end{figure*}

\begin{figure*}[htbp]
    \centering
    \includegraphics[width=1\linewidth]{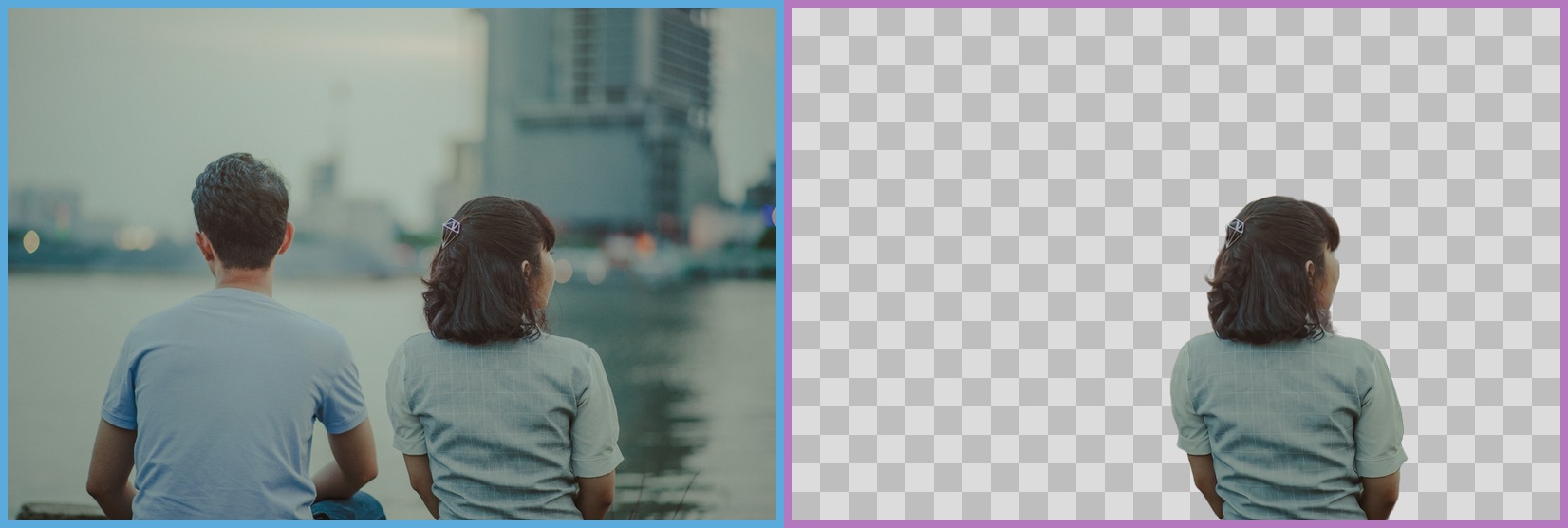}
    \caption{Extract the object described by the text, preserving fine details and transparency: the girl is sitting at the rightside of the photo}
    \label{fig:app:refmatte:rim_rw_d7560f25_layer_01}
\end{figure*}

\begin{figure*}[htbp]
    \centering
    \includegraphics[width=1\linewidth]{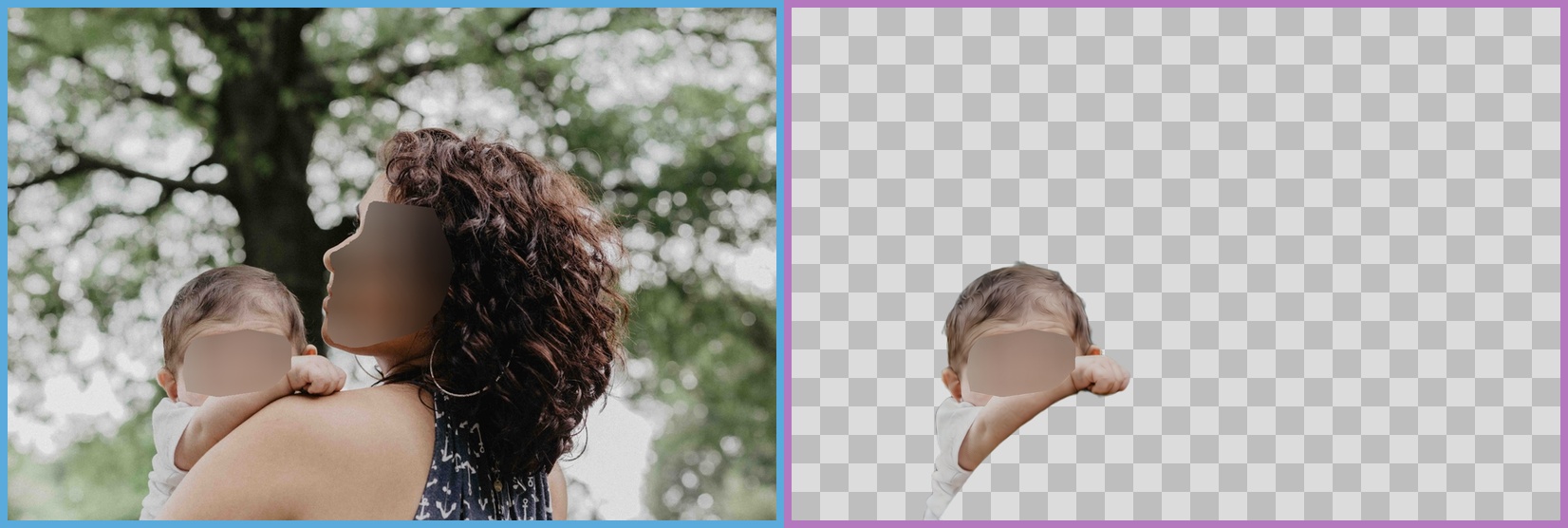}
    \caption{Extract the object described by the text, preserving fine details and transparency: the male baby in white t-shirt locates at the left of the picture}
    \label{fig:app:refmatte:end}
\end{figure*}

\begin{figure*}[htbp]
    \centering
    \includegraphics[width=1\linewidth]{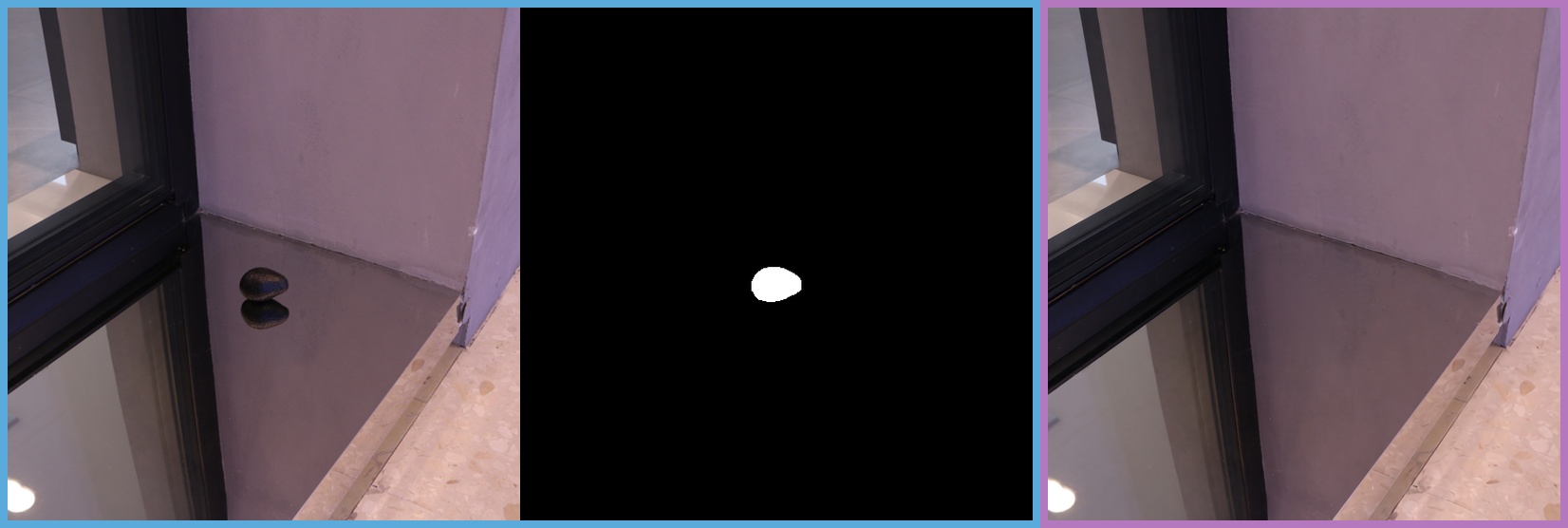}
    \caption{Use the first image as the source scene and the second image as the object mask. Remove the masked object and all of its associated effects, including shadows, reflections, highlights, contact traces, and residual artifacts, even when these effects extend beyond the mask. Reconstruct the clean base background as if the object had never been present.}
    \label{fig:app:removal:start}
\end{figure*}

\begin{figure*}[htbp]
    \centering
    \includegraphics[width=1\linewidth]{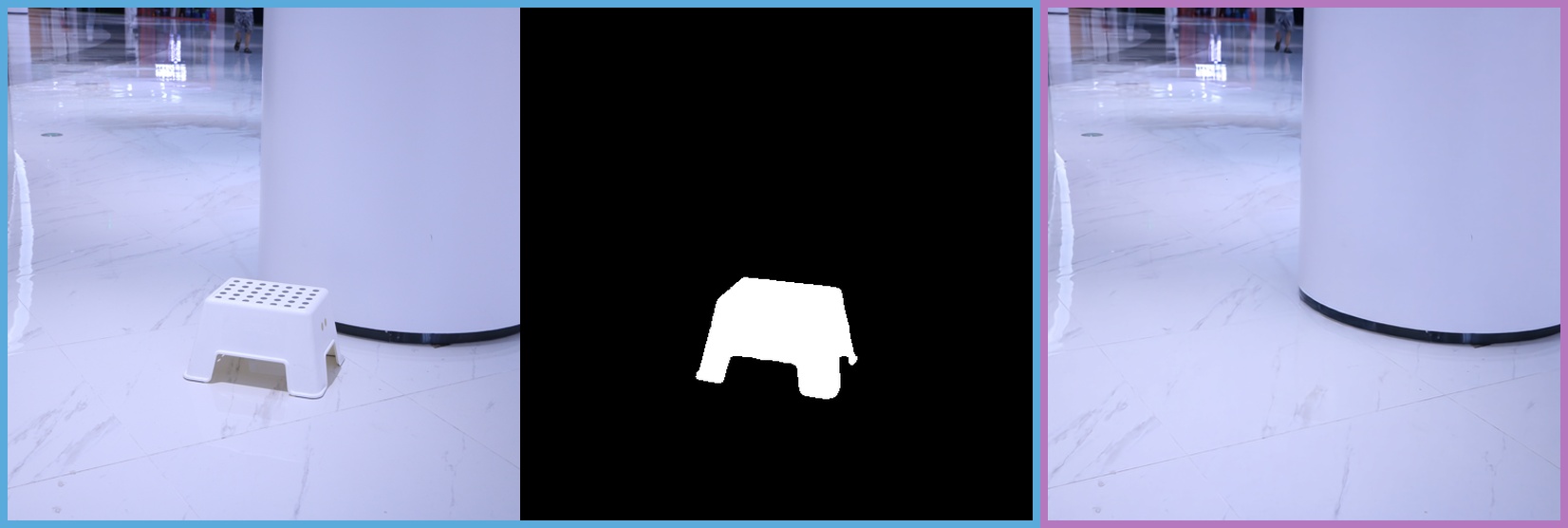}
    \caption{Use the first image as the source scene and the second image as the object mask. Remove the masked object and all of its associated effects, including shadows, reflections, highlights, contact traces, and residual artifacts, even when these effects extend beyond the mask. Reconstruct the clean base background as if the object had never been present.}
    \label{fig:app:removal:OBER-Test:091}
\end{figure*}

\begin{figure*}[htbp]
    \centering
    \includegraphics[width=1\linewidth]{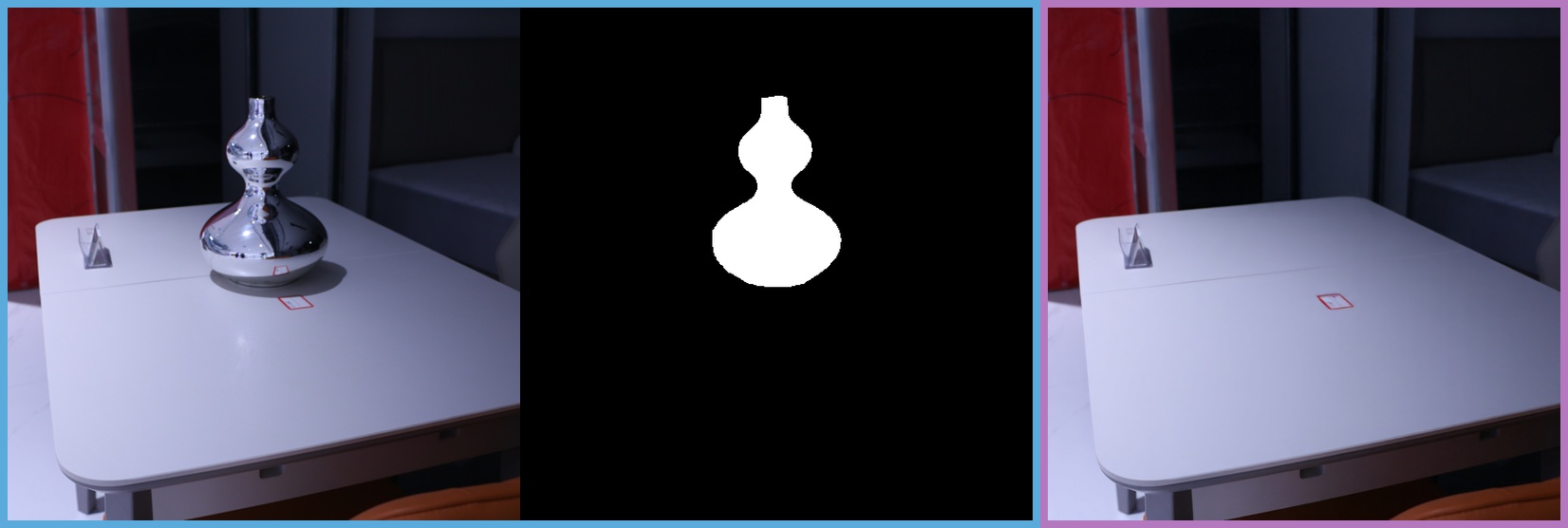}
    \caption{Use the first image as the source scene and the second image as the object mask. Remove the masked object and all of its associated effects, including shadows, reflections, highlights, contact traces, and residual artifacts, even when these effects extend beyond the mask. Reconstruct the clean base background as if the object had never been present.}
    \label{fig:app:removal:OBER-Test:111}
\end{figure*}

\begin{figure*}[htbp]
    \centering
    \includegraphics[width=1\linewidth]{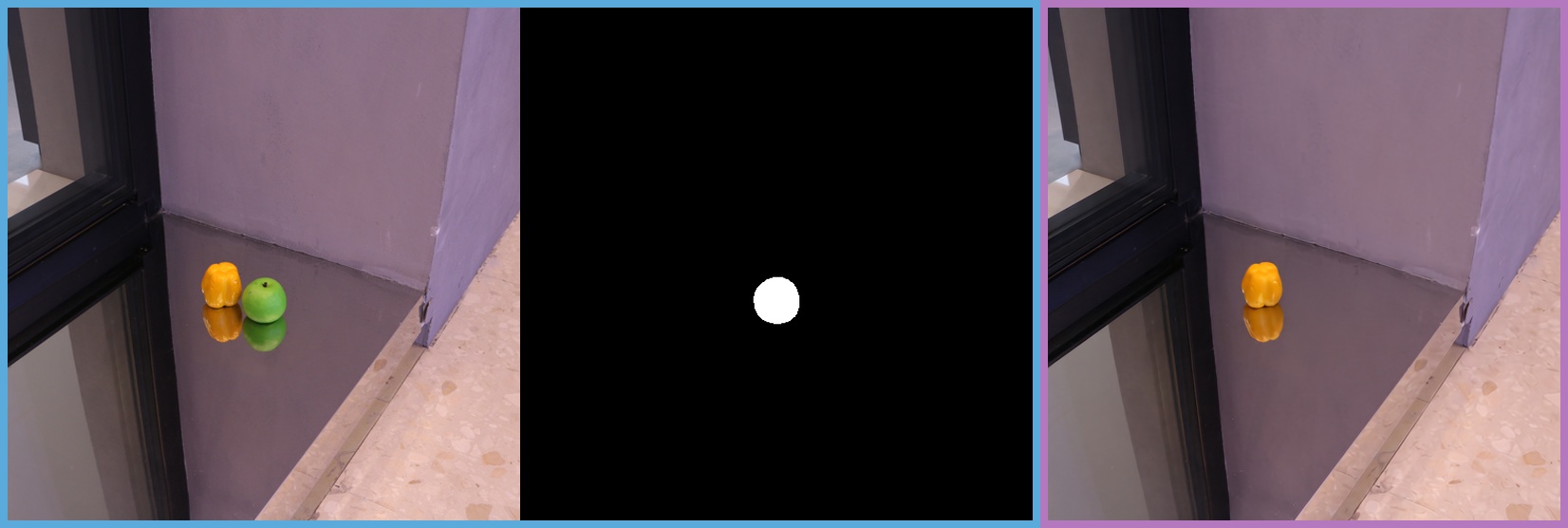}
    \caption{Use the first image as the source scene and the second image as the object mask. Remove the masked object and all of its associated effects, including shadows, reflections, highlights, contact traces, and residual artifacts, even when these effects extend beyond the mask. Reconstruct the clean base background as if the object had never been present.}
    \label{fig:app:removal:OBER-Test:052}
\end{figure*}

\begin{figure*}[htbp]
    \centering
    \includegraphics[width=1\linewidth]{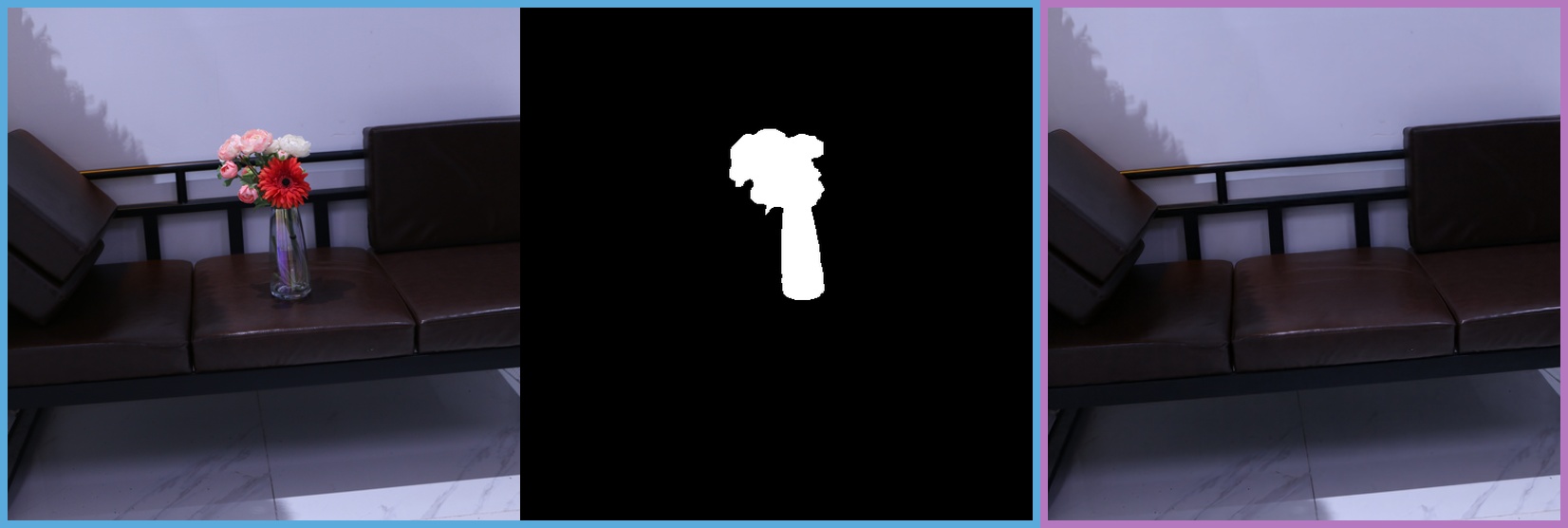}
    \caption{Use the first image as the source scene and the second image as the object mask. Remove the masked object and all of its associated effects, including shadows, reflections, highlights, contact traces, and residual artifacts, even when these effects extend beyond the mask. Reconstruct the clean base background as if the object had never been present.}
    \label{fig:app:removal:OBER-Test:102}
\end{figure*}

\begin{figure*}[htbp]
    \centering
    \includegraphics[width=1\linewidth]{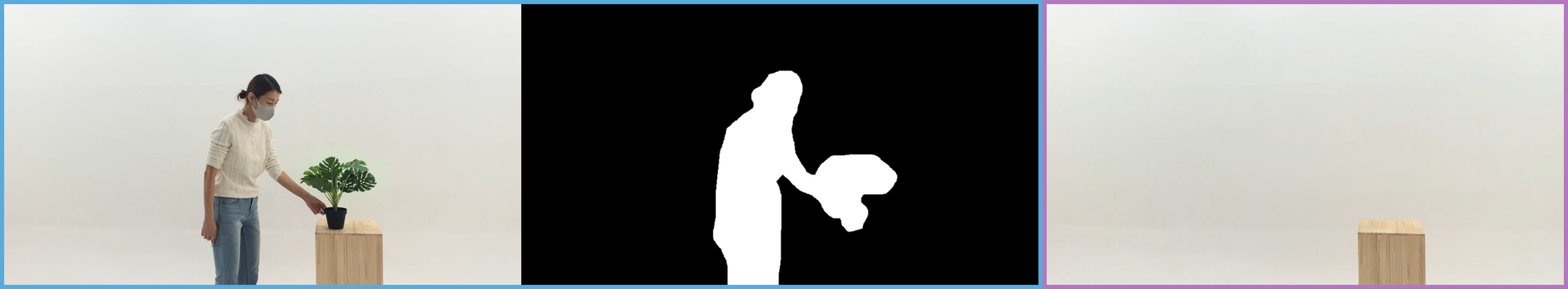}
    \caption{Use the first image as the source scene and the second image as the object mask. Remove the masked object and all of its associated effects, including shadows, reflections, highlights, contact traces, and residual artifacts, even when these effects extend beyond the mask. Reconstruct the clean base background as if the object had never been present.}
    \label{fig:app:removal:RORD-Val-343:I-211016_I09025_T03}
\end{figure*}

\begin{figure*}[htbp]
    \centering
    \includegraphics[width=1\linewidth]{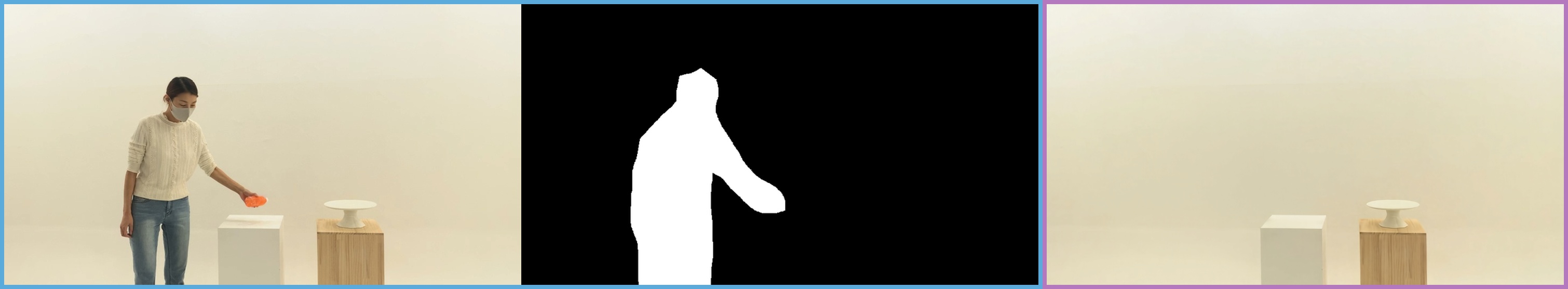}
    \caption{Use the first image as the source scene and the second image as the object mask. Remove the masked object and all of its associated effects, including shadows, reflections, highlights, contact traces, and residual artifacts, even when these effects extend beyond the mask. Reconstruct the clean base background as if the object had never been present.}
    \label{fig:app:removal:RORD-Val-343:I-211016_I09026_T02}
\end{figure*}

\begin{figure*}[htbp]
    \centering
    \includegraphics[width=1\linewidth]{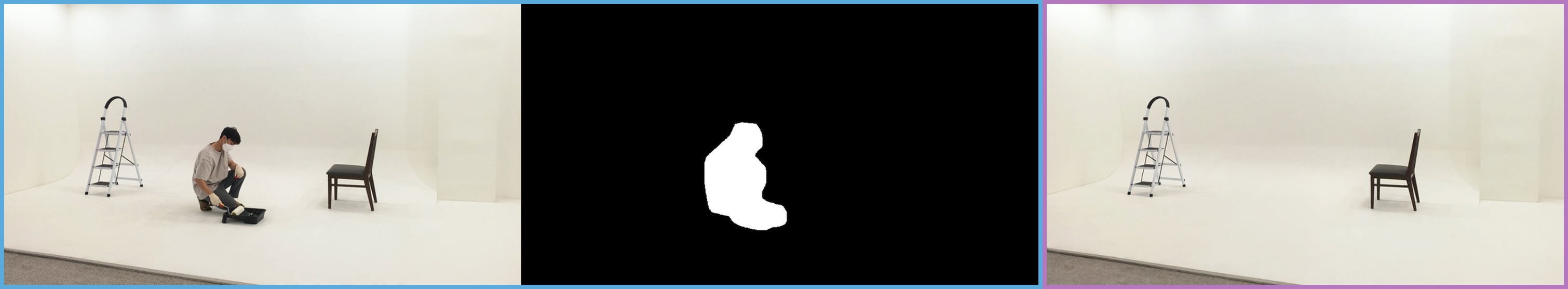}
    \caption{Use the first image as the source scene and the second image as the object mask. Remove the masked object and all of its associated effects, including shadows, reflections, highlights, contact traces, and residual artifacts, even when these effects extend beyond the mask. Reconstruct the clean base background as if the object had never been present.}
    \label{fig:app:removal:RORD-Val-343:I-211016_I02009_T07}
\end{figure*}

\begin{figure*}[htbp]
    \centering
    \includegraphics[width=1\linewidth]{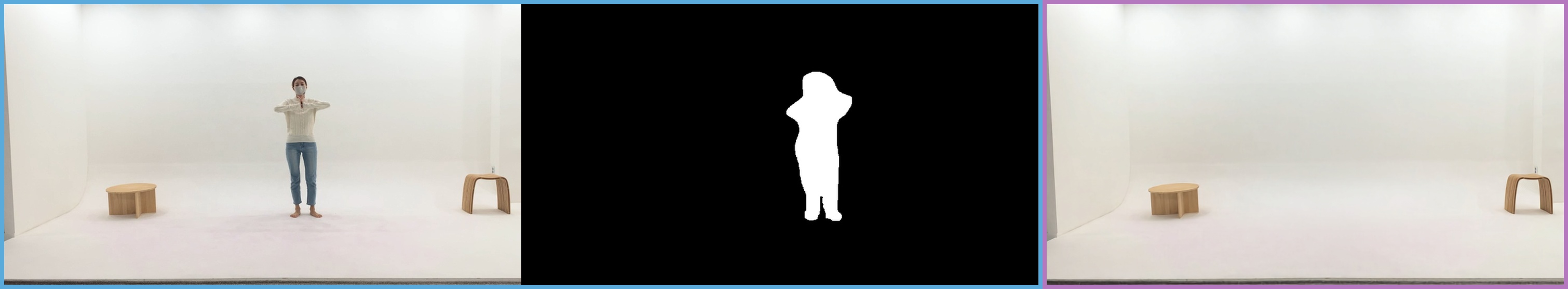}
    \caption{Use the first image as the source scene and the second image as the object mask. Remove the masked object and all of its associated effects, including shadows, reflections, highlights, contact traces, and residual artifacts, even when these effects extend beyond the mask. Reconstruct the clean base background as if the object had never been present.}
    \label{fig:app:removal:RORD-Val-343:I-211016_I04015_T07}
\end{figure*}

\begin{figure*}[htbp]
    \centering
    \includegraphics[width=1\linewidth]{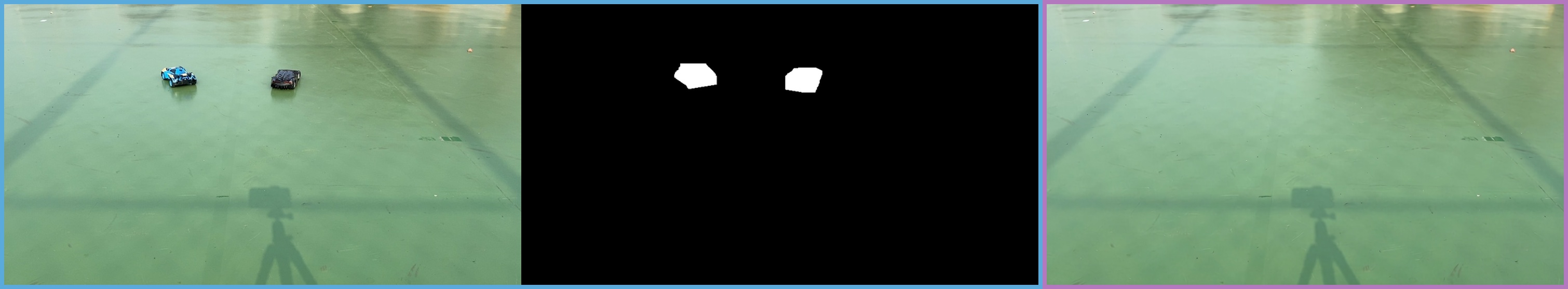}
    \caption{Use the first image as the source scene and the second image as the object mask. Remove the masked object and all of its associated effects, including shadows, reflections, highlights, contact traces, and residual artifacts, even when these effects extend beyond the mask. Reconstruct the clean base background as if the object had never been present.}
    \label{fig:app:removal:RORD-Val-343:I-211120_O07035_T24}
\end{figure*}

\begin{figure*}[htbp]
    \centering
    \includegraphics[width=1\linewidth]{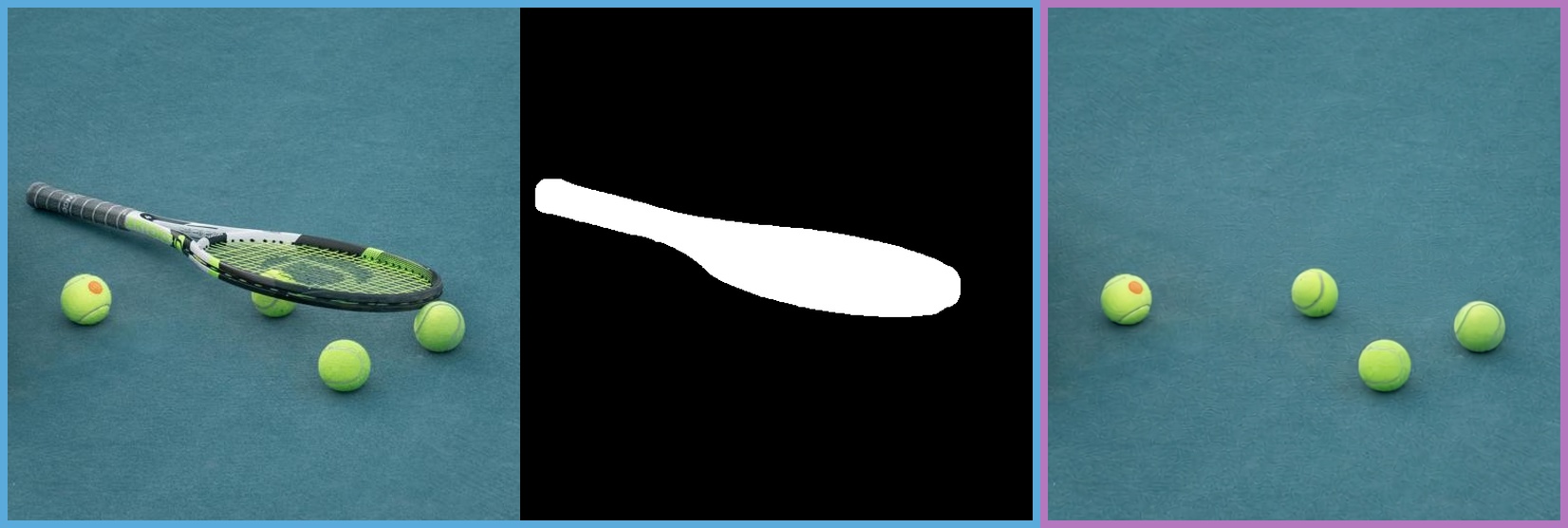}
    \caption{Use the first image as the source scene and the second image as the object mask. Remove the masked object and all of its associated effects, including shadows, reflections, highlights, contact traces, and residual artifacts, even when these effects extend beyond the mask. Reconstruct the clean base background as if the object had never been present.}
    \label{fig:app:removal:OBER-Wild:026}
\end{figure*}

\begin{figure*}[htbp]
    \centering
    \includegraphics[width=1\linewidth]{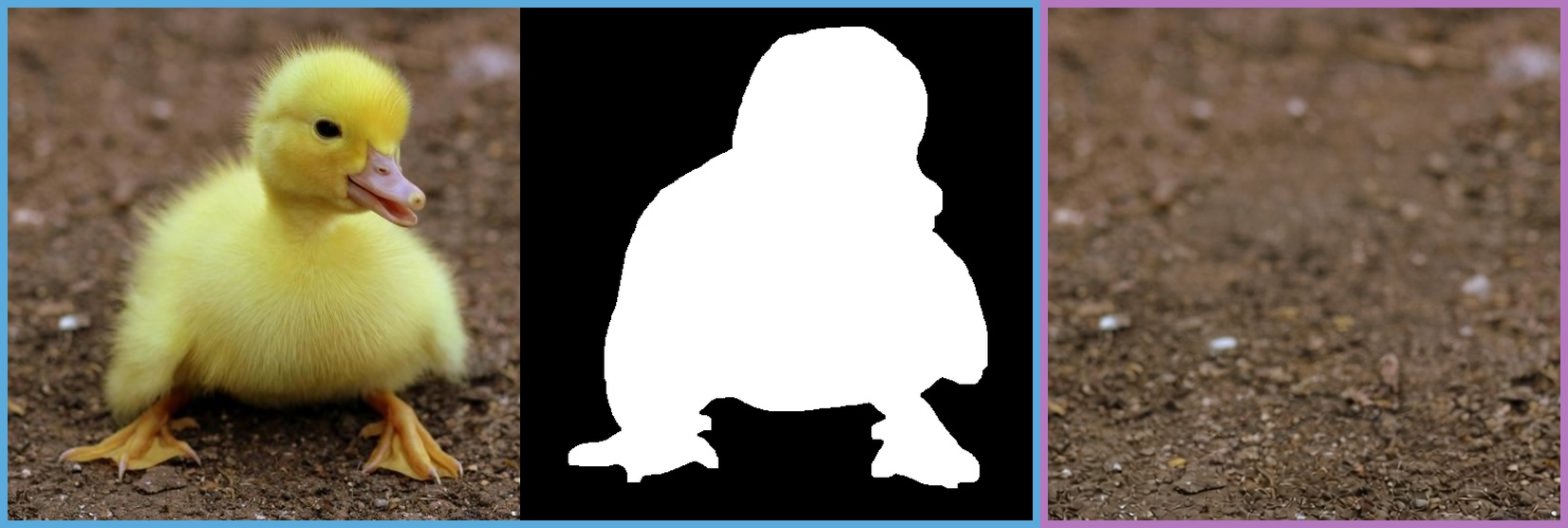}
    \caption{Use the first image as the source scene and the second image as the object mask. Remove the masked object and all of its associated effects, including shadows, reflections, highlights, contact traces, and residual artifacts, even when these effects extend beyond the mask. Reconstruct the clean base background as if the object had never been present.}
    \label{fig:app:removal:OBER-Wild:291}
\end{figure*}

\begin{figure*}[htbp]
    \centering
    \includegraphics[width=1\linewidth]{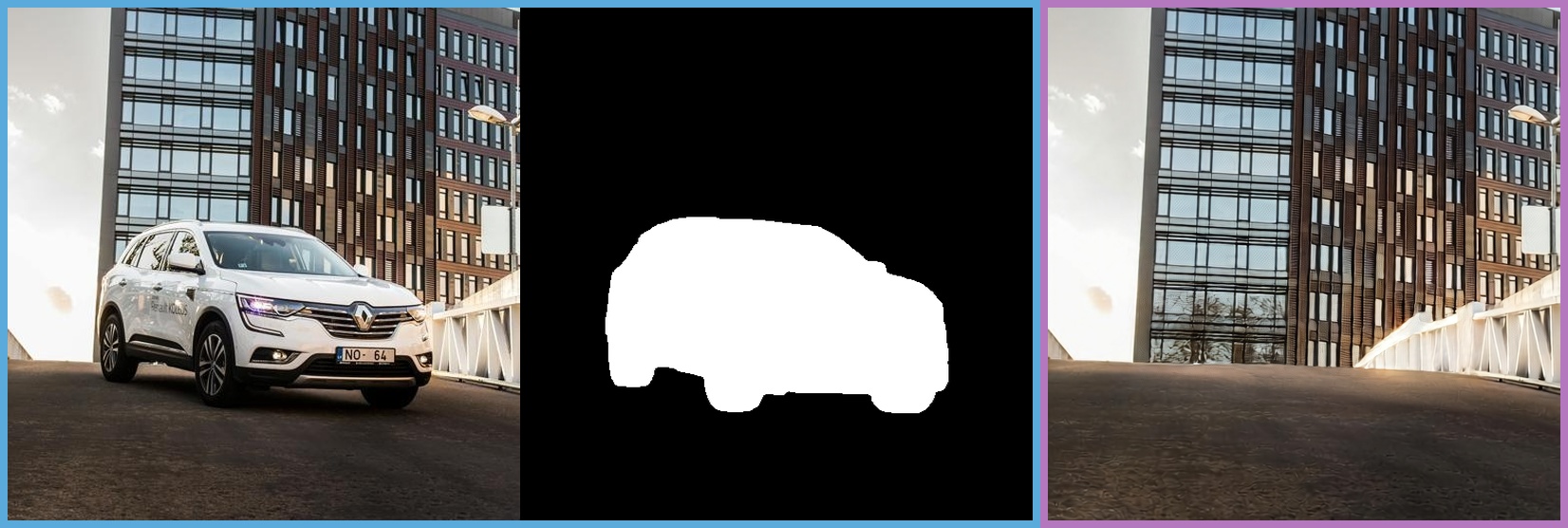}
    \caption{Use the first image as the source scene and the second image as the object mask. Remove the masked object and all of its associated effects, including shadows, reflections, highlights, contact traces, and residual artifacts, even when these effects extend beyond the mask. Reconstruct the clean base background as if the object had never been present.}
    \label{fig:app:removal:OBER-Wild:004}
\end{figure*}

\begin{figure*}[htbp]
    \centering
    \includegraphics[width=1\linewidth]{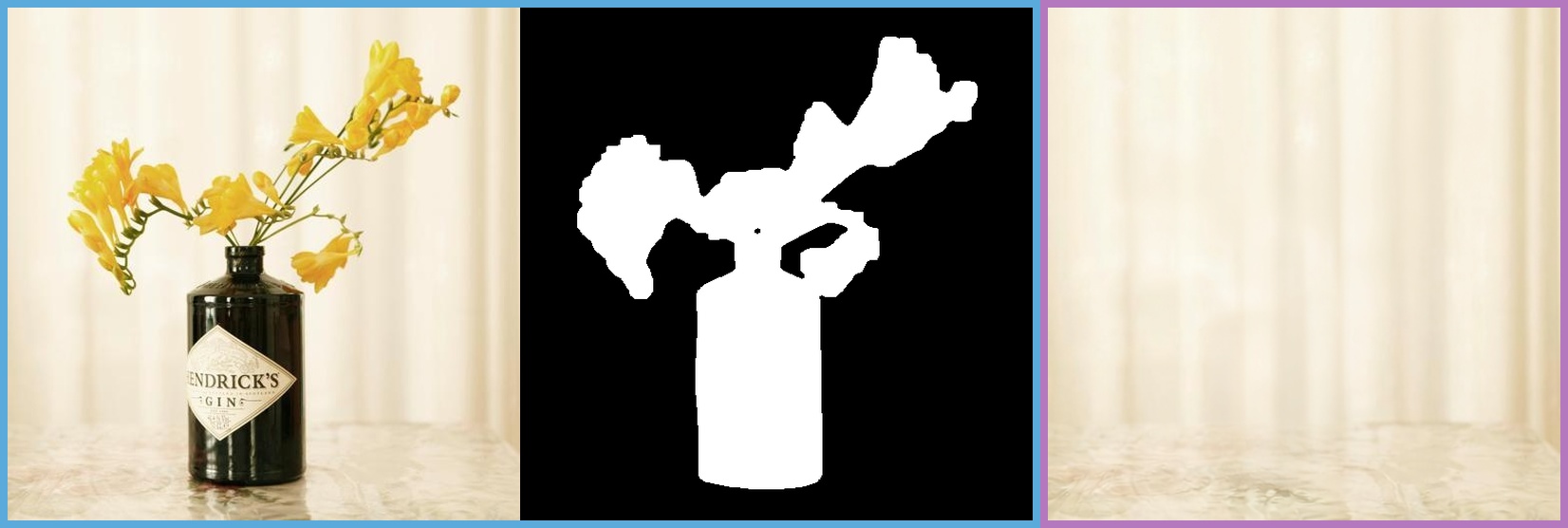}
    \caption{Use the first image as the source scene and the second image as the object mask. Remove the masked object and all of its associated effects, including shadows, reflections, highlights, contact traces, and residual artifacts, even when these effects extend beyond the mask. Reconstruct the clean base background as if the object had never been present.}
    \label{fig:app:removal:OBER-Wild:211}
\end{figure*}

\begin{figure*}[htbp]
    \centering
    \includegraphics[width=1\linewidth]{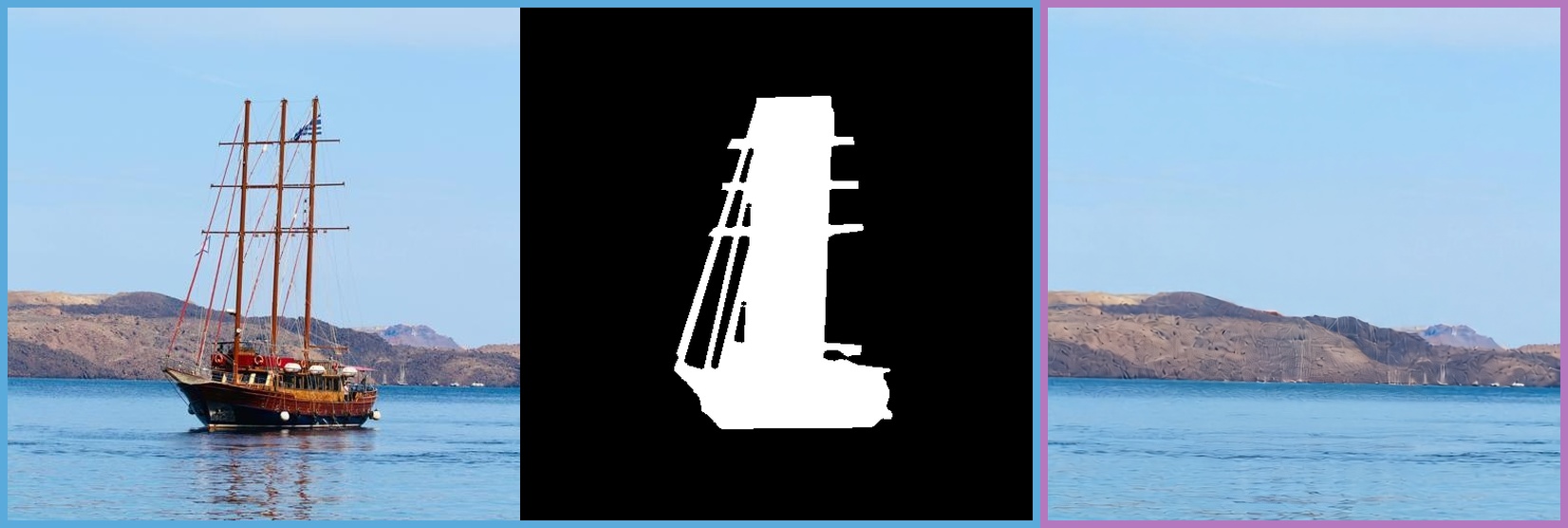}
    \caption{Use the first image as the source scene and the second image as the object mask. Remove the masked object and all of its associated effects, including shadows, reflections, highlights, contact traces, and residual artifacts, even when these effects extend beyond the mask. Reconstruct the clean base background as if the object had never been present.}
    \label{fig:app:removal:end}
\end{figure*}

\begin{figure*}[htbp]
    \centering
    \includegraphics[width=1\linewidth]{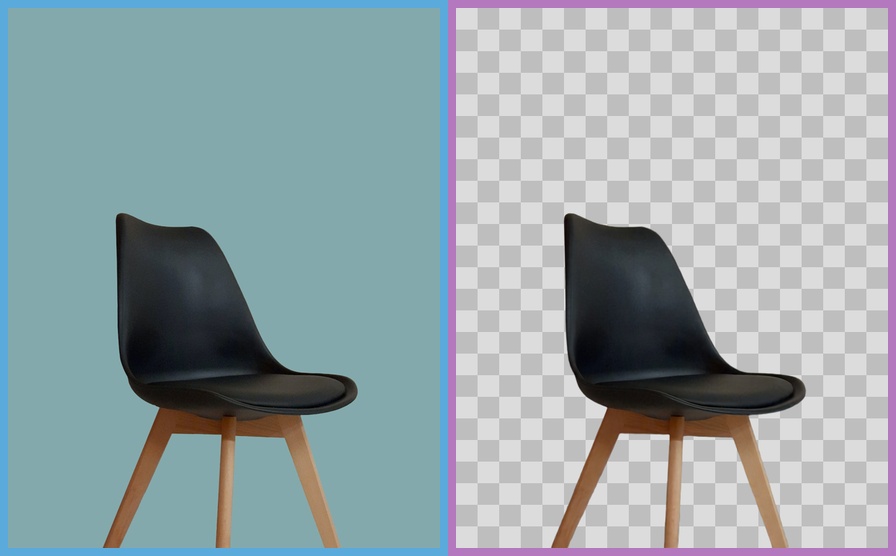}
    \caption{Automatically matte this image and extract the foreground with a physically accurate alpha channel that preserves true transparency and fine details.}
    \label{fig:app:automatting:start}
\end{figure*}

\begin{figure*}[htbp]
    \centering
    \includegraphics[width=1\linewidth]{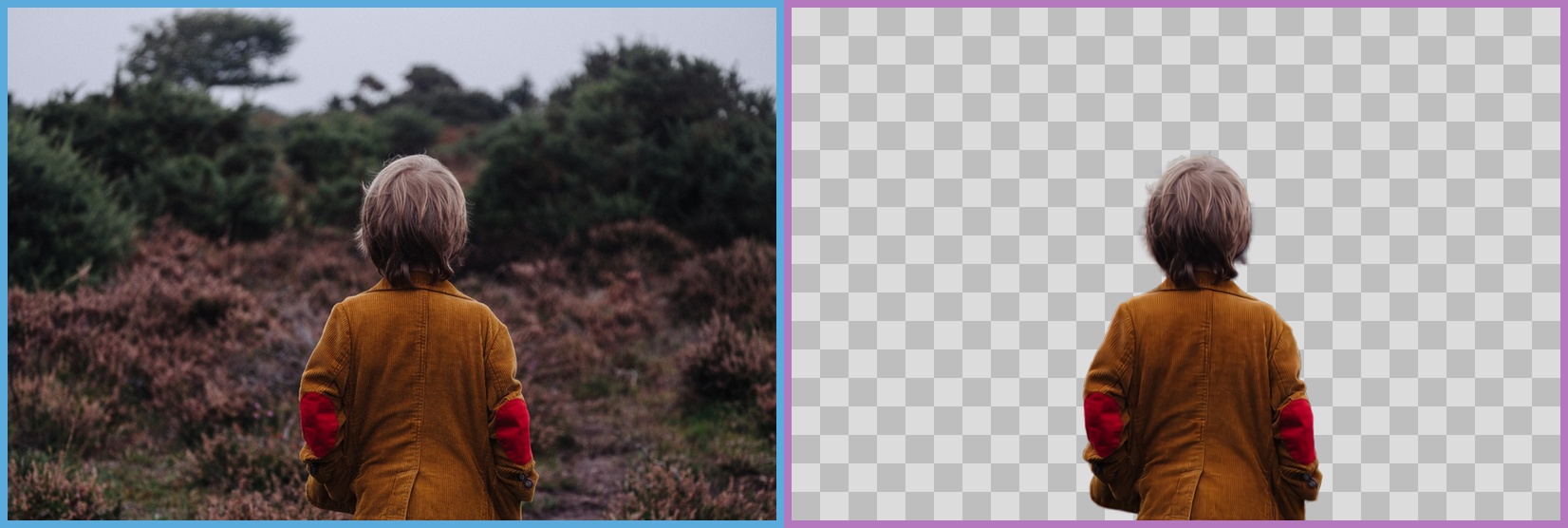}
    \caption{Automatically matte this image and extract the foreground with a physically accurate alpha channel that preserves true transparency and fine details.}
    \label{fig:app:automatting:o_eaeab68d}
\end{figure*}

\begin{figure*}[htbp]
    \centering
    \includegraphics[width=1\linewidth]{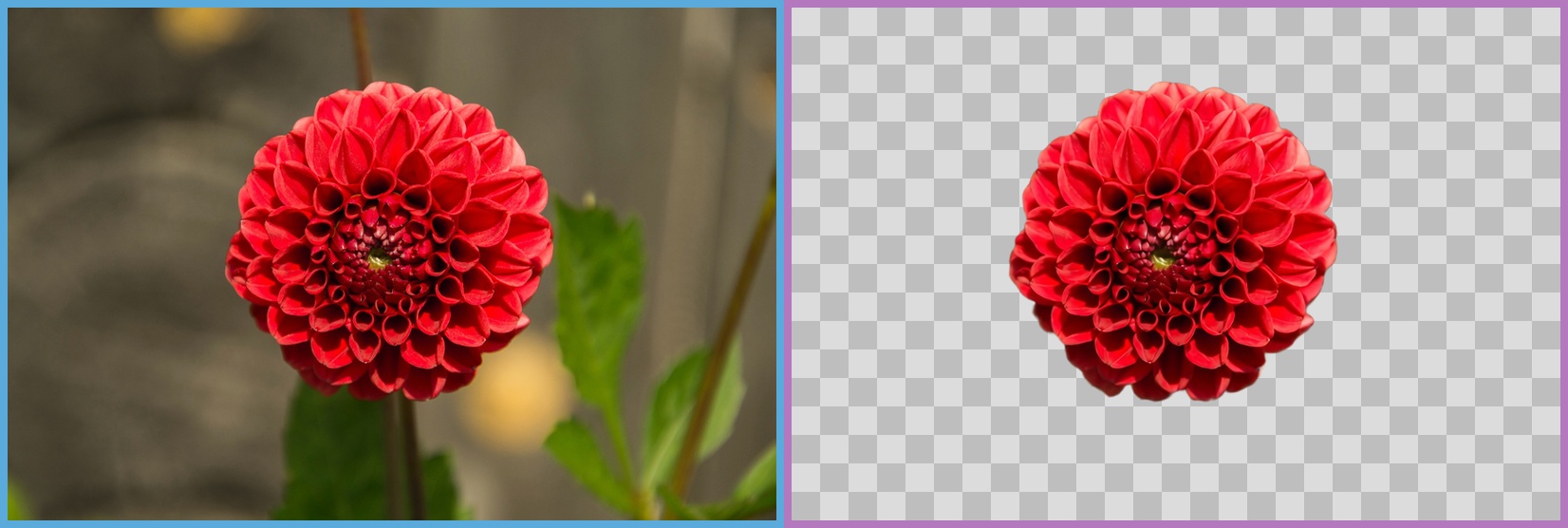}
    \caption{Automatically matte this image and extract the foreground with a physically accurate alpha channel that preserves true transparency and fine details.}
    \label{fig:app:automatting:o_a1513d78}
\end{figure*}

\begin{figure*}[htbp]
    \centering
    \includegraphics[width=1\linewidth]{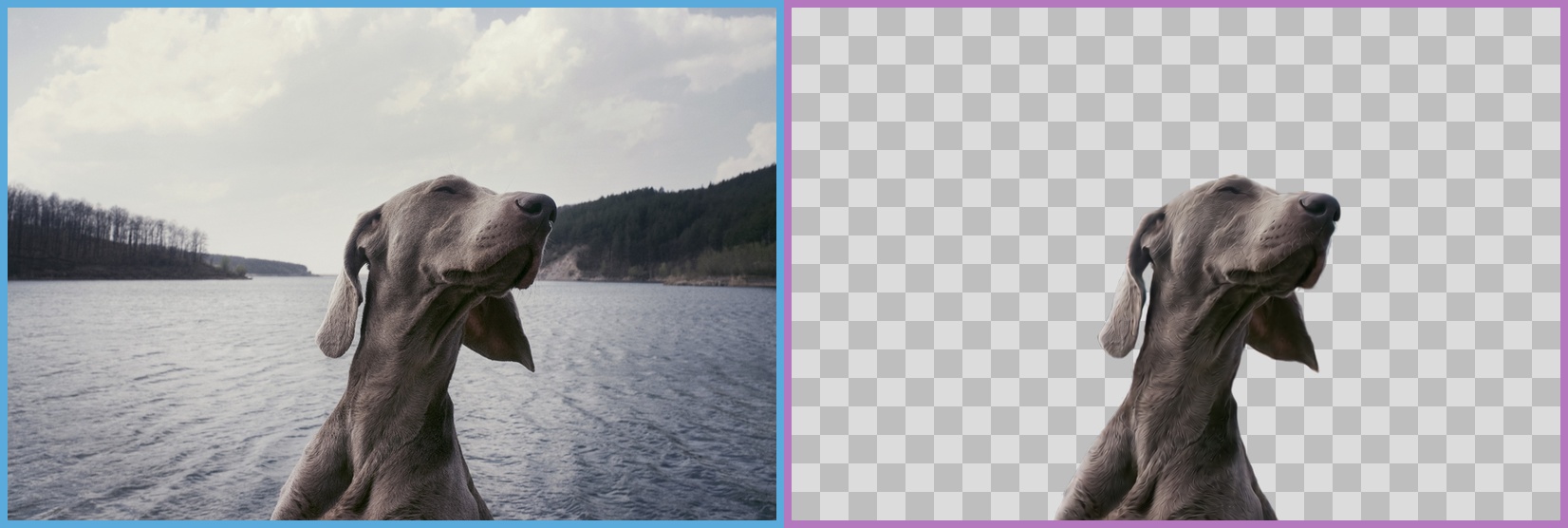}
    \caption{Automatically matte this image and extract the foreground with a physically accurate alpha channel that preserves true transparency and fine details.}
    \label{fig:app:automatting:o_706c33eb}
\end{figure*}

\begin{figure*}[htbp]
    \centering
    \includegraphics[width=1\linewidth]{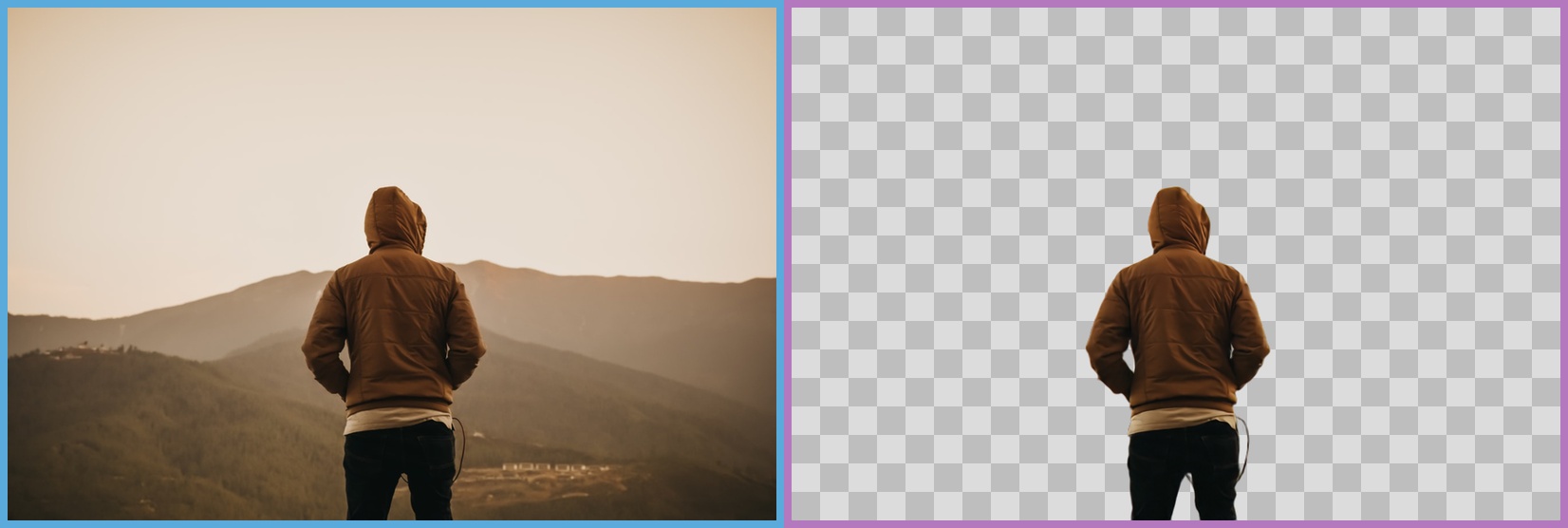}
    \caption{Automatically matte this image and extract the foreground with a physically accurate alpha channel that preserves true transparency and fine details.}
    \label{fig:app:automatting:o_73696a44}
\end{figure*}

\begin{figure*}[htbp]
    \centering
    \includegraphics[width=1\linewidth]{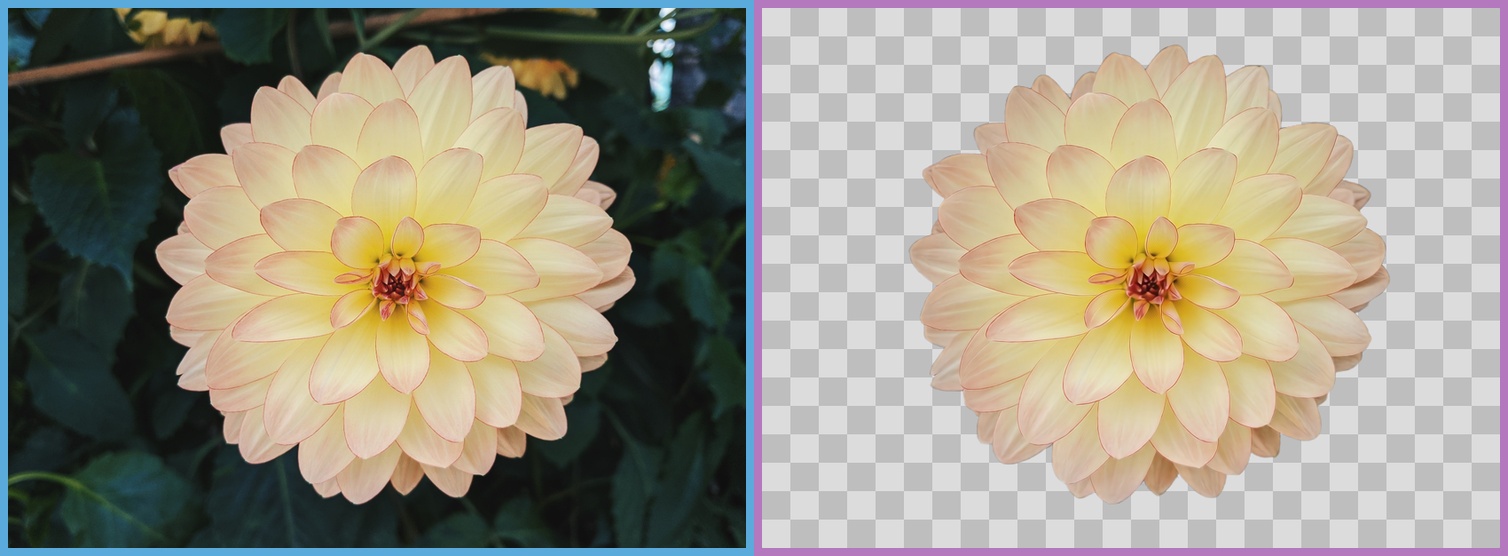}
    \caption{Automatically matte this image and extract the foreground with a physically accurate alpha channel that preserves true transparency and fine details.}
    \label{fig:app:automatting:o_dbef692f}
\end{figure*}

\begin{figure*}[htbp]
    \centering
    \includegraphics[width=1\linewidth]{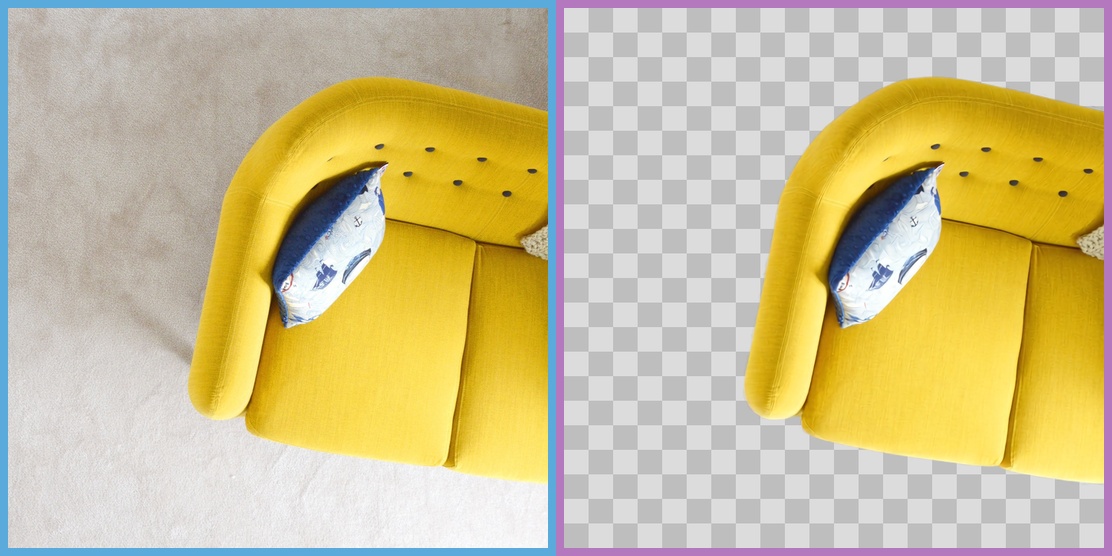}
    \caption{Automatically matte this image and extract the foreground with a physically accurate alpha channel that preserves true transparency and fine details.}
    \label{fig:app:automatting:o_5851551d}
\end{figure*}

\begin{figure*}[htbp]
    \centering
    \includegraphics[width=1\linewidth]{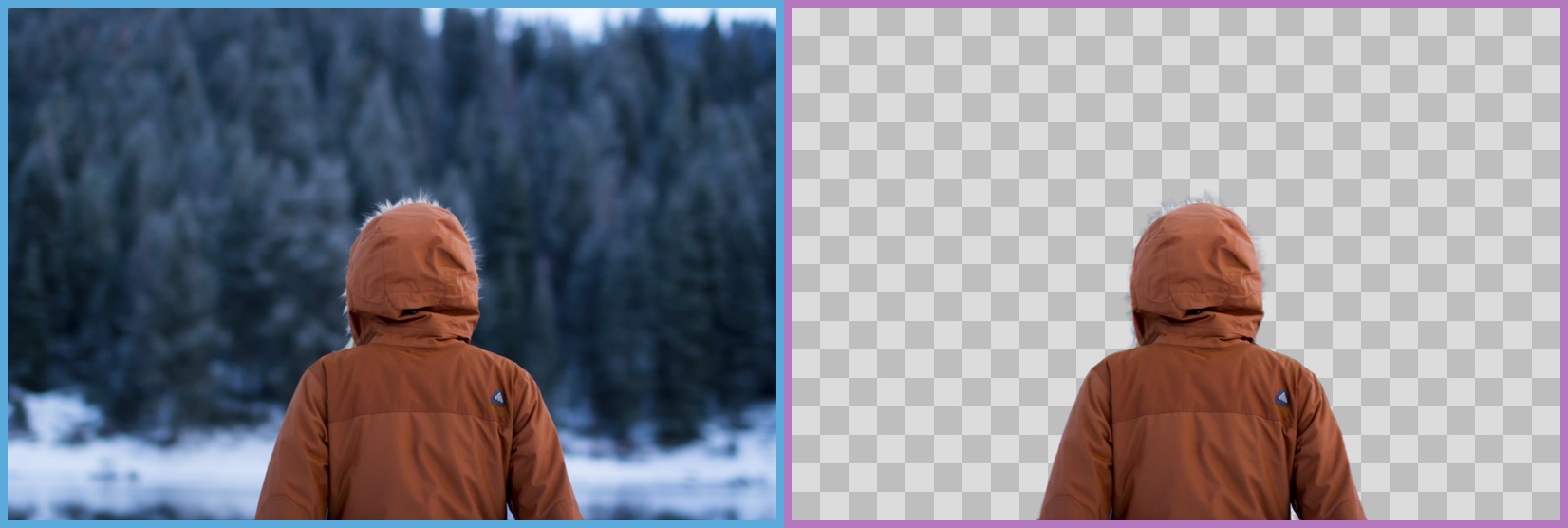}
    \caption{Automatically matte this image and extract the foreground with a physically accurate alpha channel that preserves true transparency and fine details.}
    \label{fig:app:automatting:o_dd0868e0}
\end{figure*}

\begin{figure*}[htbp]
    \centering
    \includegraphics[width=1\linewidth]{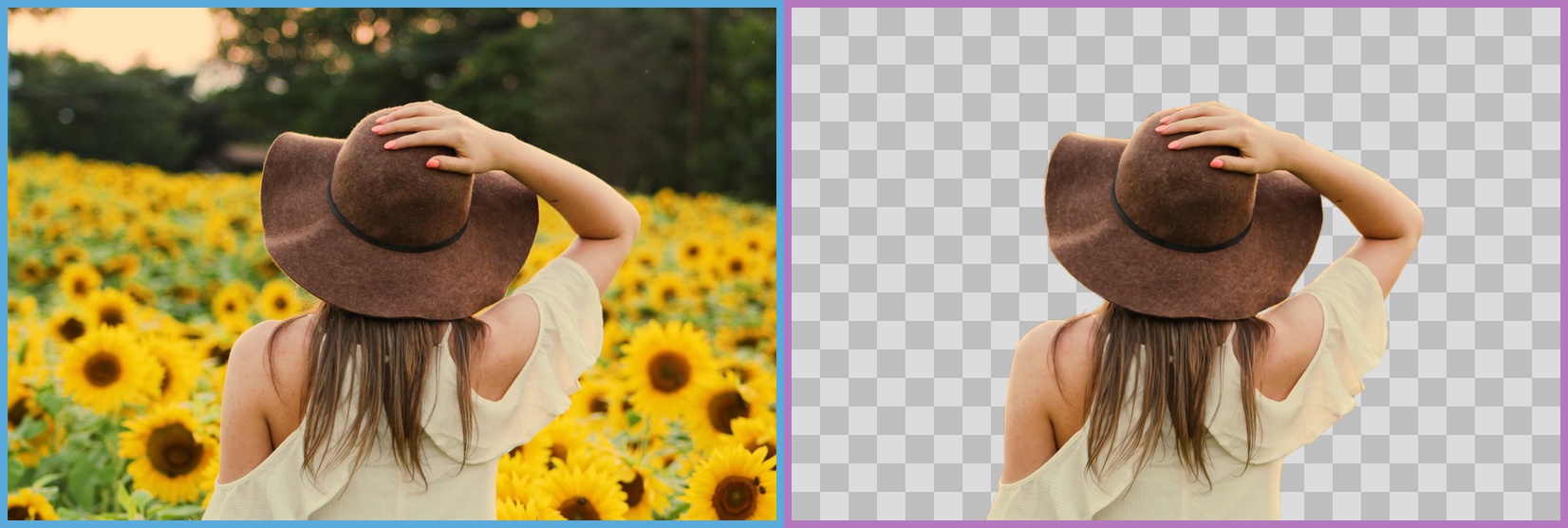}
    \caption{Automatically matte this image and extract the foreground with a physically accurate alpha channel that preserves true transparency and fine details.}
    \label{fig:app:automatting:o_9fb57e99}
\end{figure*}

\begin{figure*}[htbp]
    \centering
    \includegraphics[width=1\linewidth]{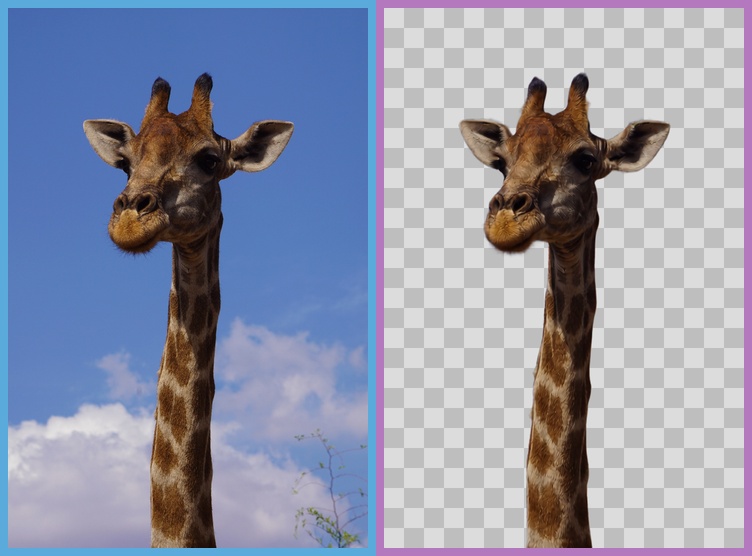}
    \caption{Automatically matte this image and extract the foreground with a physically accurate alpha channel that preserves true transparency and fine details.}
    \label{fig:app:automatting:end}
\end{figure*}

\begin{figure*}[htbp]
    \centering
    \includegraphics[width=1\linewidth]{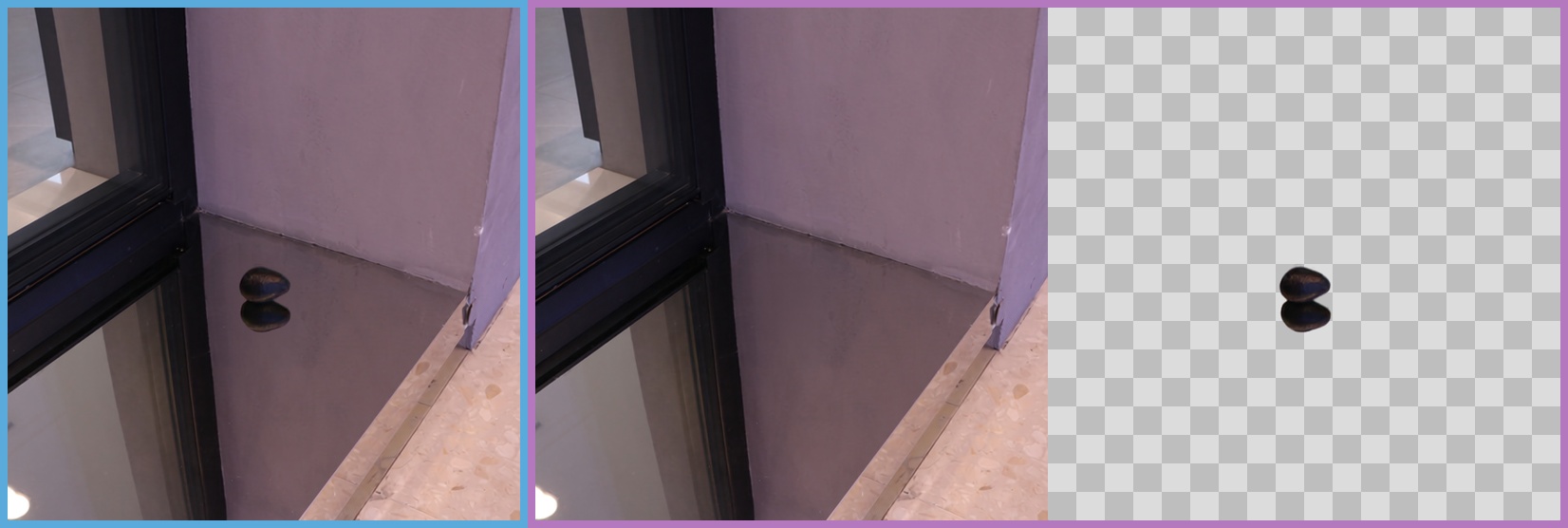}
    \caption{Qualitative results of image decomposition. From left to right: reconstructed input image, predicted base layer, and predicted object layer visualized on a checkerboard background.}
    \label{fig:app:decompose:start}
\end{figure*}

\begin{figure*}[htbp]
    \centering
    \includegraphics[width=1\linewidth]{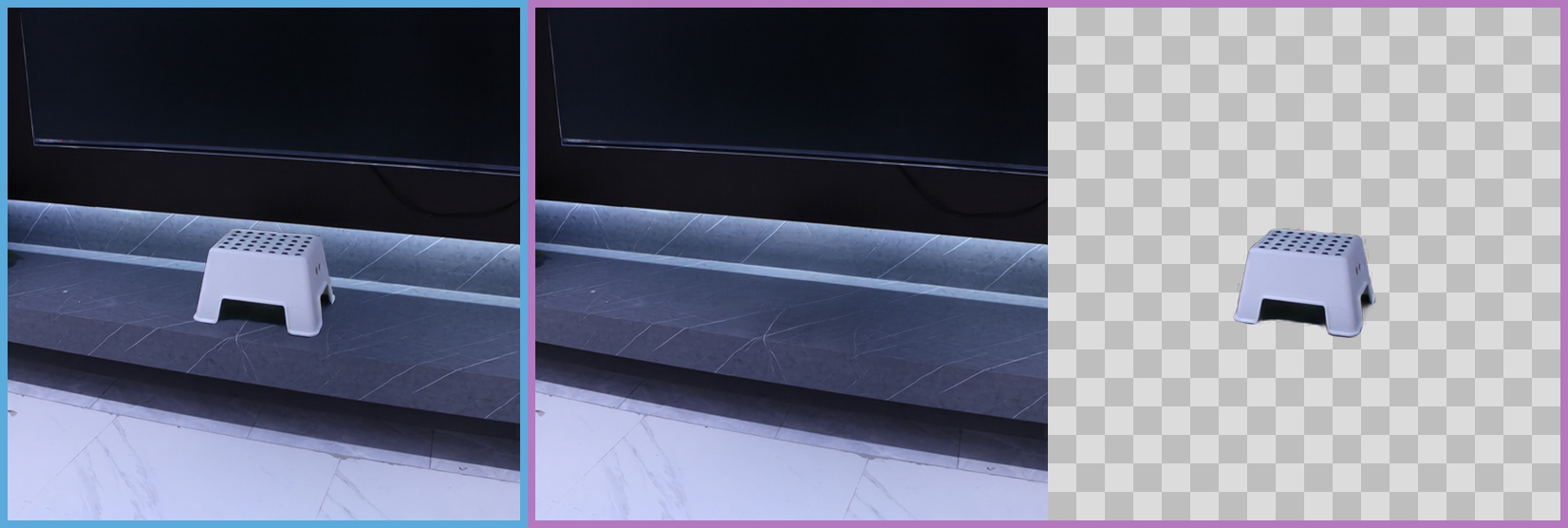}
    \caption{Qualitative results of image decomposition. From left to right: reconstructed input image, predicted base layer, and predicted object layer visualized on a checkerboard background.}
    \label{fig:app:decompose:OBER-Test_095}
\end{figure*}

\begin{figure*}[htbp]
    \centering
    \includegraphics[width=1\linewidth]{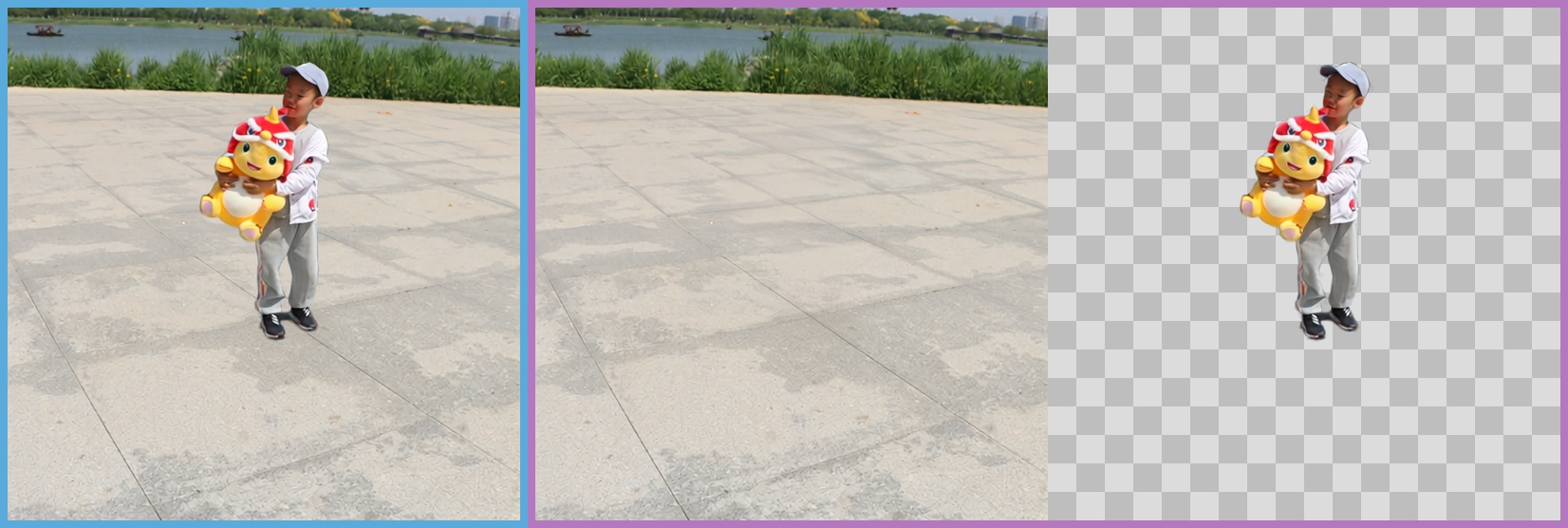}
    \caption{Qualitative results of image decomposition. From left to right: reconstructed input image, predicted base layer, and predicted object layer visualized on a checkerboard background.}
    \label{fig:app:decompose:end}
\end{figure*}

Figures \ref{fig:app:t2i:start}–\ref{fig:app:decompose:end} present a set of randomly sampled results. Each caption corresponds to the prompt provided to \method{}. Images outlined in blue denote inputs, while those outlined in purple represent predictions. 
% \todo{}

\begin{figure*}[htbp]
    \centering
    \includegraphics[width=1\linewidth]{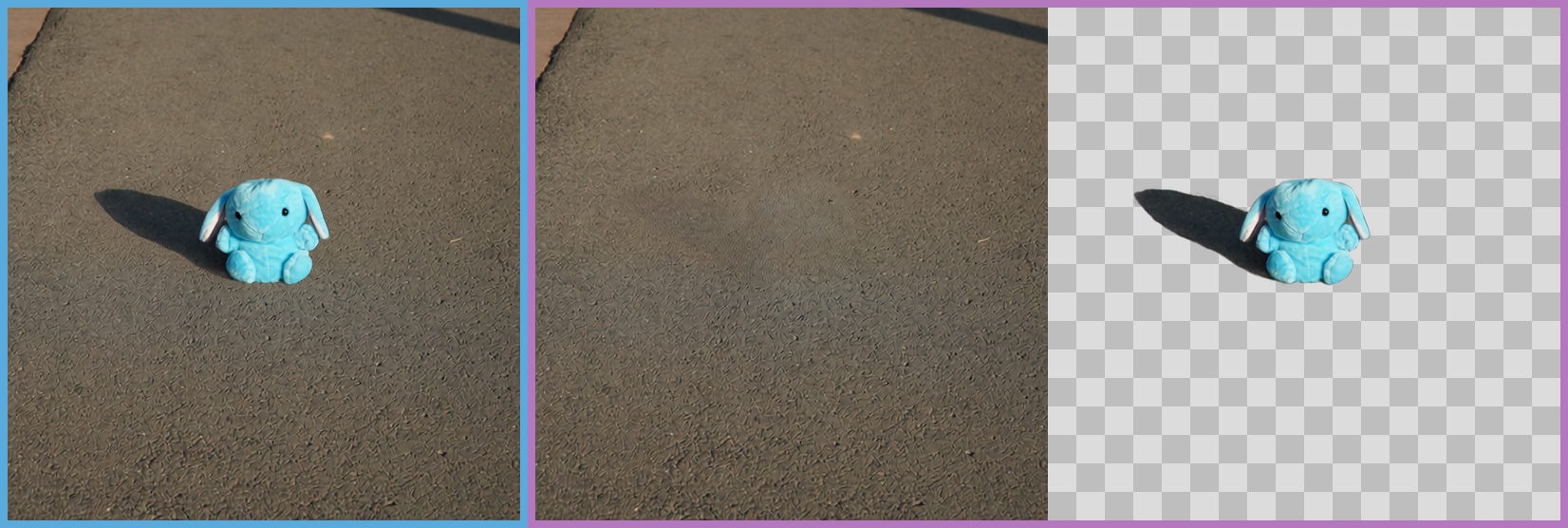}
    \caption{Qualitative results of image decomposition. From left to right: reconstructed input image, predicted base layer, and predicted object layer visualized on a checkerboard background.}
    \label{fig:app:decompose:OBER-Test_147}
\end{figure*}

\begin{figure*}[htbp]
    \centering
    \includegraphics[width=1\linewidth]{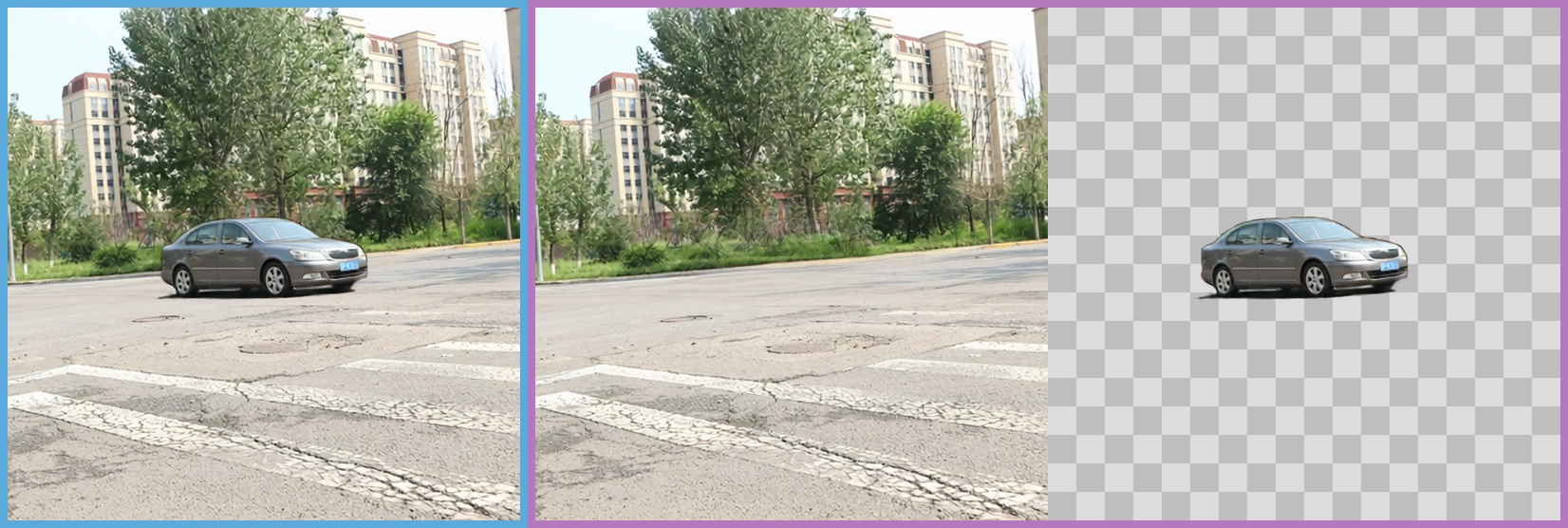}
    \caption{Qualitative results of image decomposition. From left to right: reconstructed input image, predicted base layer, and predicted object layer visualized on a checkerboard background.}
    \label{fig:app:decompose:OBER-Test_160}
\end{figure*}

\begin{figure*}[htbp]
    \centering
    \includegraphics[width=1\linewidth]{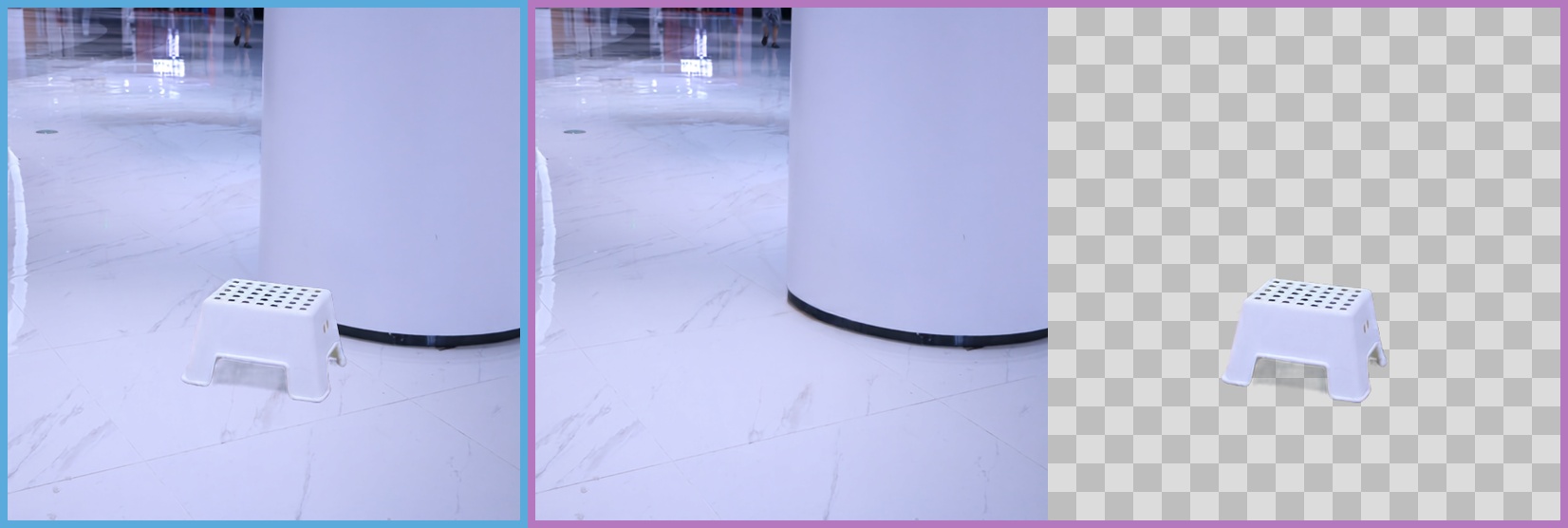}
    \caption{Qualitative results of image decomposition. From left to right: reconstructed input image, predicted base layer, and predicted object layer visualized on a checkerboard background.}
    \label{fig:app:decompose:OBER-Test_091}
\end{figure*}

\begin{figure*}[htbp]
    \centering
    \includegraphics[width=1\linewidth]{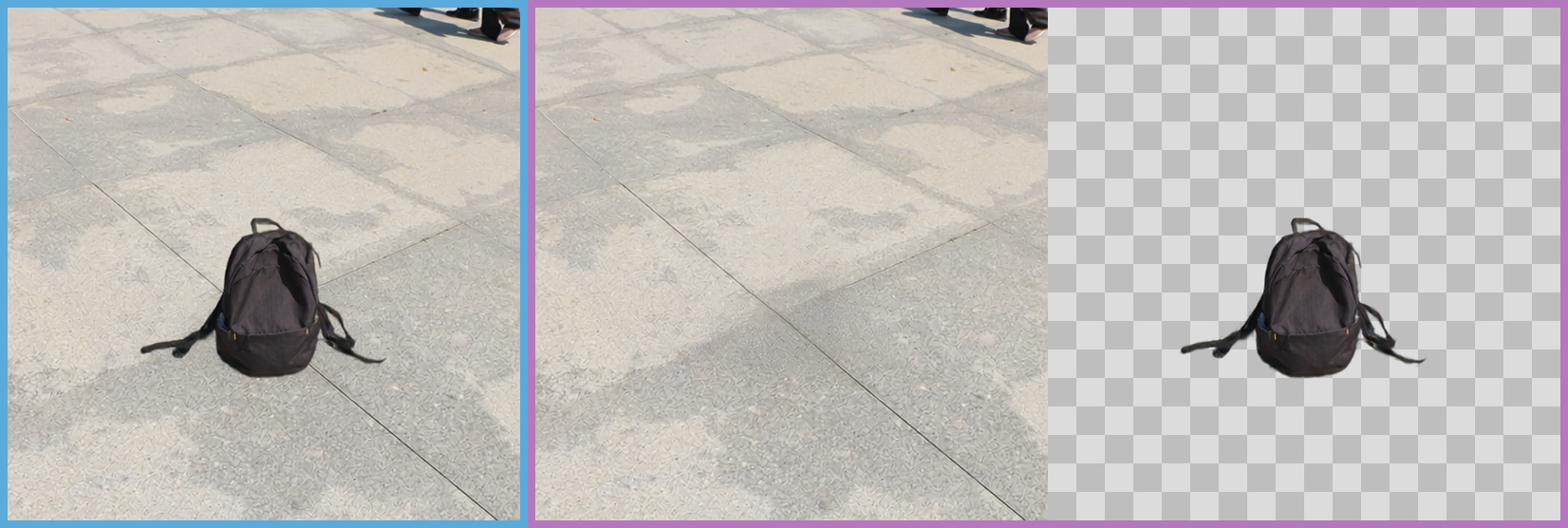}
    \caption{Qualitative results of image decomposition. From left to right: reconstructed input image, predicted base layer, and predicted object layer visualized on a checkerboard background.}
    \label{fig:app:decompose:OBER-Test_020}
\end{figure*}

\begin{figure*}[htbp]
    \centering
    \includegraphics[width=1\linewidth]{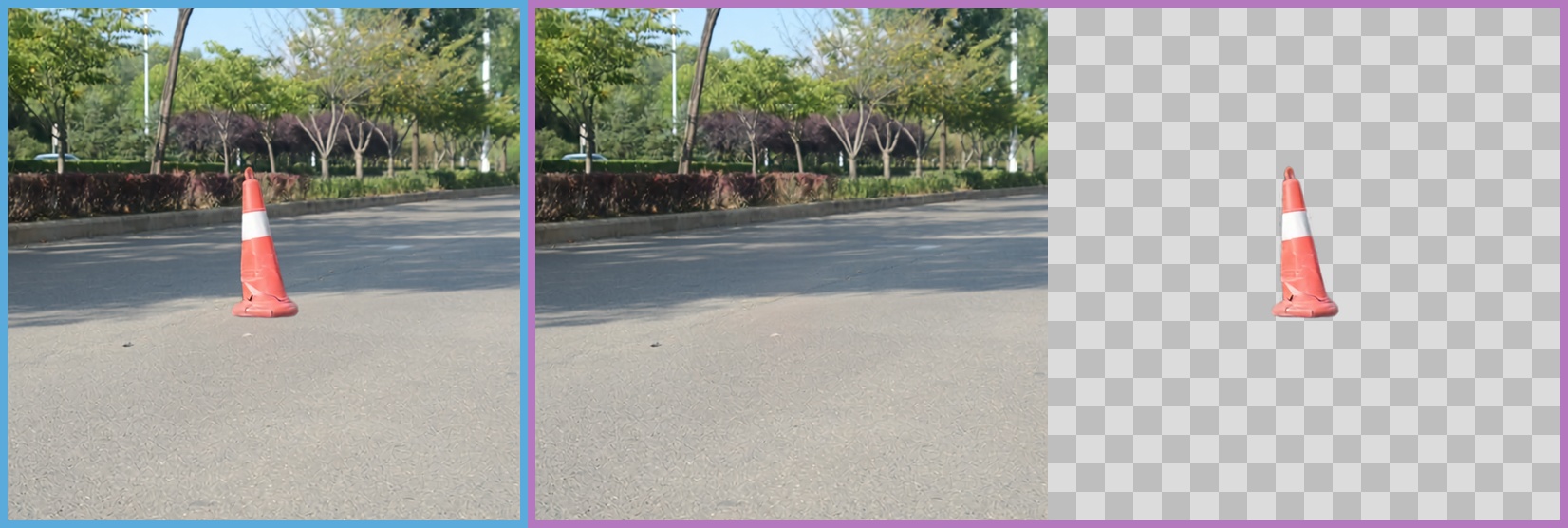}
    \caption{Qualitative results of image decomposition. From left to right: reconstructed input image, predicted base layer, and predicted object layer visualized on a checkerboard background.}
    \label{fig:app:decompose:OBER-Test_149}
\end{figure*}

\begin{figure*}[htbp]
    \centering
    \includegraphics[width=1\linewidth]{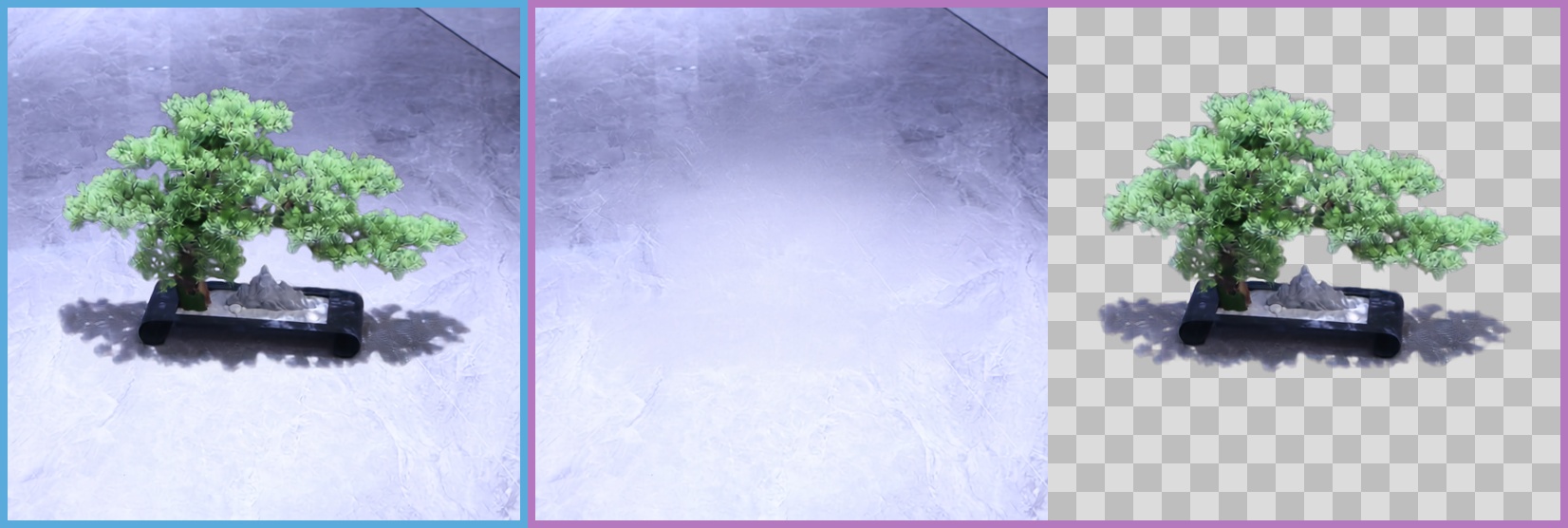}
    \caption{Qualitative results of image decomposition. From left to right: reconstructed input image, predicted base layer, and predicted object layer visualized on a checkerboard background.}
    \label{fig:app:decompose:OBER-Test_051}
\end{figure*}

\begin{figure*}[htbp]
    \centering
    \includegraphics[width=1\linewidth]{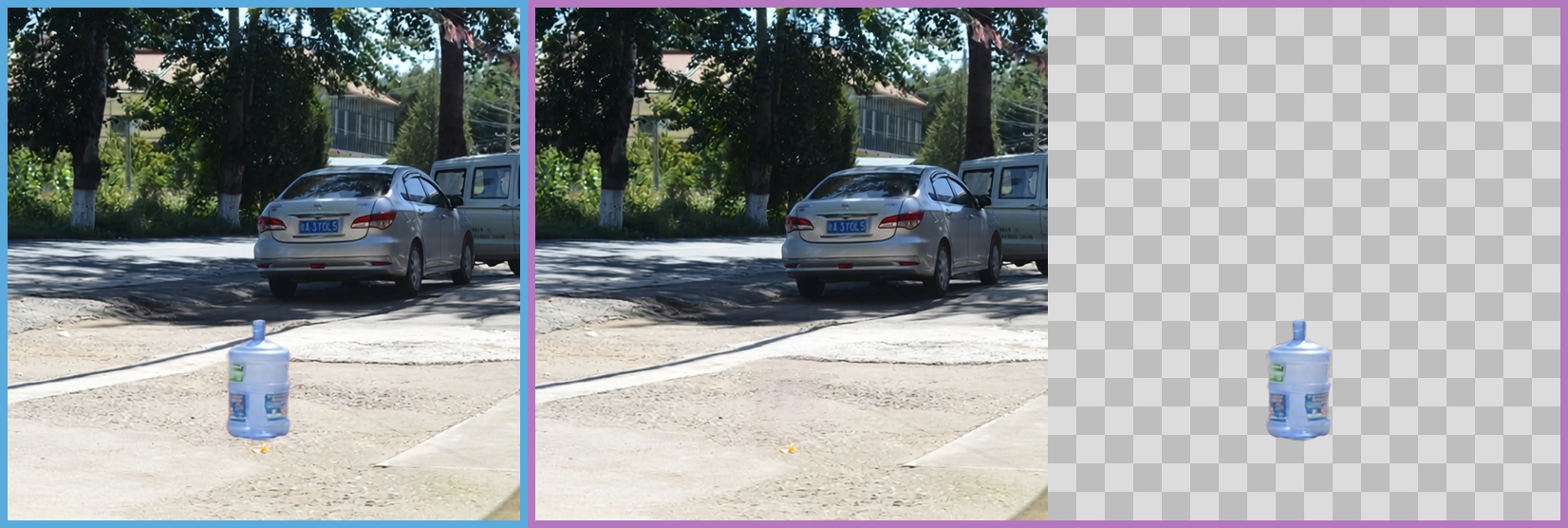}
    \caption{Qualitative results of image decomposition. From left to right: reconstructed input image, predicted base layer, and predicted object layer visualized on a checkerboard background.}
    \label{fig:app:decompose:OBER-Test_012}
\end{figure*}

\begin{itemize}
% \item Figures \ref{fig:app:matting:start}–\ref{fig:app:matting:end} show results for the image matting task category.\todo{}
% \item Figures \ref{fig:app:completion:start}–\ref{fig:app:completion:end} present the layer-conditioned completion task category.\todo{}
% \item Figures \ref{fig:app:decomposition:start}–\ref{fig:app:decomposition:end} illustrate the layer decomposition task category.\todo{}
\item Figures \ref{fig:app:t2i:start}–\ref{fig:app:t2i:end} showcase the text-to-image task category.
\item Figures \ref{fig:app:refmatte:start}–\ref{fig:app:refmatte:end} show results for the referring matting task category.
\item Figures \ref{fig:app:removal:start}–\ref{fig:app:removal:end} display results for the object removal task category.
\item Figures \ref{fig:app:automatting:start}–\ref{fig:app:automatting:end} show results for the automatic matting task category.
\item Figures \ref{fig:app:decompose:start}–\ref{fig:app:decompose:end} illustrate the layer decomposition task category.
\end{itemize}

\clearpage

\clearpage

\clearpage

\clearpage

\end{document}